*To be submitted*

# The structure of variational autoencoder latent spaces and its consequences for representation and optimization

Sergei V. Kalinin

Department of Materials Science and Engineering, University of Tennessee, Knoxville, Tennessee, USA

**Abstract**

Variational autoencoders (VAEs) are broadly used to build low-dimensional representations of high-dimensional scientific data, with the expectation that regularization produces smooth coordinates for visualization, interpolation, generation and optimization. However, applications in microscopy and related fields reveal the presence of domains, narrow transition regions and rare-state islands in the latent spaces of VAEs that are not explained by smoothness alone. Here we show that, for prototype data with squared-error reconstruction and an optimal decoder, the latent-space objective has a direct correspondence with the Flory–Huggins theory of a regular solution. We use this correspondence to analyze how the reconstruction objective, posterior distributions and latent dimensionality determine the organization of the representation. In the optimized Gaussian-prototype model, data families occupy domains separated by finite-width walls; increasing regularization broadens these walls until class separation is lost through posterior collapse. The classes centroids behavior in the latent space resembles that of particles in a harmonic trap: the prior confines them, structural dissimilarity sets a soft repulsion between them, and together these determine which classes become neighbors and how far apart they sit, with separations growing only logarithmically with dissimilarity. Rare states can remain unresolved, form small islands, accumulate within walls or develop their own intervening domains, depending on their frequency and similarity to the common states. The posterior covariance influences the orientation and arrangement of domains and can align them with the latent axes, but axis alignment does not by itself imply separation of physical factors. Rotational variability can produce rings and encoder cuts, depending on data symmetry and the representation, while insufficient dimensionality can lead to suppression, mixing or folding of additional factors. Comparisons with flow models, deep kernel learning (DKL) and hybrid VAE–DKL models show how these structures change when reconstruction and prediction are assigned different roles. We analyze the implications of this latent structure for interpreting scientific data, representing rare events, generating intermediate states and optimizing physical properties.

## 1. Introduction

Variational autoencoders provide a way to represent high-dimensional observations using a small number of continuous latent variables [1–3]. In physical science, this offers the possibility of describing images, spectra and processing histories in terms of variables related to phase, orientation, composition, defects or response. The stochastic encoder and regularization encourage a smooth representation, making the latent coordinates attractive for visualization, interpolation and optimization. However, these coordinates are learned from the reconstruction task and are not automatically physical variables. Their interpretation depends on the data, the model and the symmetries or additional information supplied during training [4–10].

This distinction has become important in statistical physics and microscopy, where latent variables have been compared with order parameters, local structural states and transformation pathways [11–24]. Related representations connect complementary measurements, characterize microstructures and processing trajectories, and provide continuous coordinates for molecules and crystals [25–31]. In all these uses, the geometry between observed states becomes part of the scientific model. The question is therefore not only whether a VAE reconstructs its training data, but what its latent distances, interpolations and unoccupied regions mean.

Multiple observations in microscopy and related applications suggest that continuity alone is not enough to characterize a learned latent space [17–24]. A continuous decoder gives nearby outputs at nearby latent points, but those outputs need not all occur in the data or represent physically accessible states. Discrete families can occupy domains separated by narrow transitions; rare states can form small islands or become absorbed into the transitions between common families; and rotations can produce rings, cores or cuts. If the latent dimension is too small, some variation must be discarded or combined with other factors. These observations illustrate how a simple dataset can acquire a complicated representation through the requirements imposed by the model.

A separate literature concerns prediction and design after a representation has been learned. Latent molecular optimization encounters poorly supported or invalid decoded states [29, 32, 33]. DKL instead learns distances useful for a measured target and can discard structural distinctions irrelevant to that target [34–37]. Hybrid objectives ask one representation to serve both reconstruction and prediction [38, 39]. Comparing these objectives helps distinguish geometric features imposed by the VAE from those associated with a more general use of continuous coordinates.

The paper is organized as follows. Section 2 summarizes the observed latent-space structure and introduces the model, its optimal decoder and the connection to Flory–Huggins theory. Section 3 examines domain walls, posterior collapse and the representation of rare classes. Section 4 considers continuous variation, symmetry, latent dimensionality and training. Section 5 compares reconstruction-based and predictive representations, and Section 6 discusses the consequences for generation, interpolation and search. Section 7 summarizes the main findings, followed by the broader conclusions in Section 8. The Supplementary Information follows the same sequence and provides the derivations, numerical calculations and illustrative benchmarks.

## 2. Model, definitions and exact identities

We first summarize the qualitative observations of latent-space structures that motivate the analysis. We then consider a simplified VAE in which each data class is represented by a single prototype. This model allows us to obtain the optimal decoder, determine how regularization constrains the class distributions and identify the connection to Flory–Huggins theory. We also distinguish the quantities used to describe a wall from those used to assess the reliability of a decoded state.

## 2.1 Observed structure of VAE latent spaces

Applications of VAEs to microscopy and other scientific data have repeatedly shown that a latent space is not uniformly organized [17–24]. Different structural families occupy separate regions; narrow transitions appear between them; uncommon states can form small islands; and strong regularization can remove distinctions through posterior collapse [40–48]. Rotational variability and limited latent dimensionality introduce additional structure. These features can be inspected directly when the latent space has two dimensions. In higher-dimensional representations, a two-dimensional projection can mix domain interiors and boundaries and conceal their separation.

Figure 1 summarizes four complementary ways to examine a two-dimensional MNIST representation. The encoded means (latent distribution) show where the observed digit classes are placed, whereas the decoded class map (latent representation) shows what the model generates across the latent plane. Classifier confidence identifies outputs that are difficult for an independent classifier to recognize. The prior-to-encoded-mean density ratio compares the generation policy with the distribution of encoded observations. These quantities answer different questions: a point can be probable under the prior, sparsely represented by the data and nevertheless receive a confident classifier label.

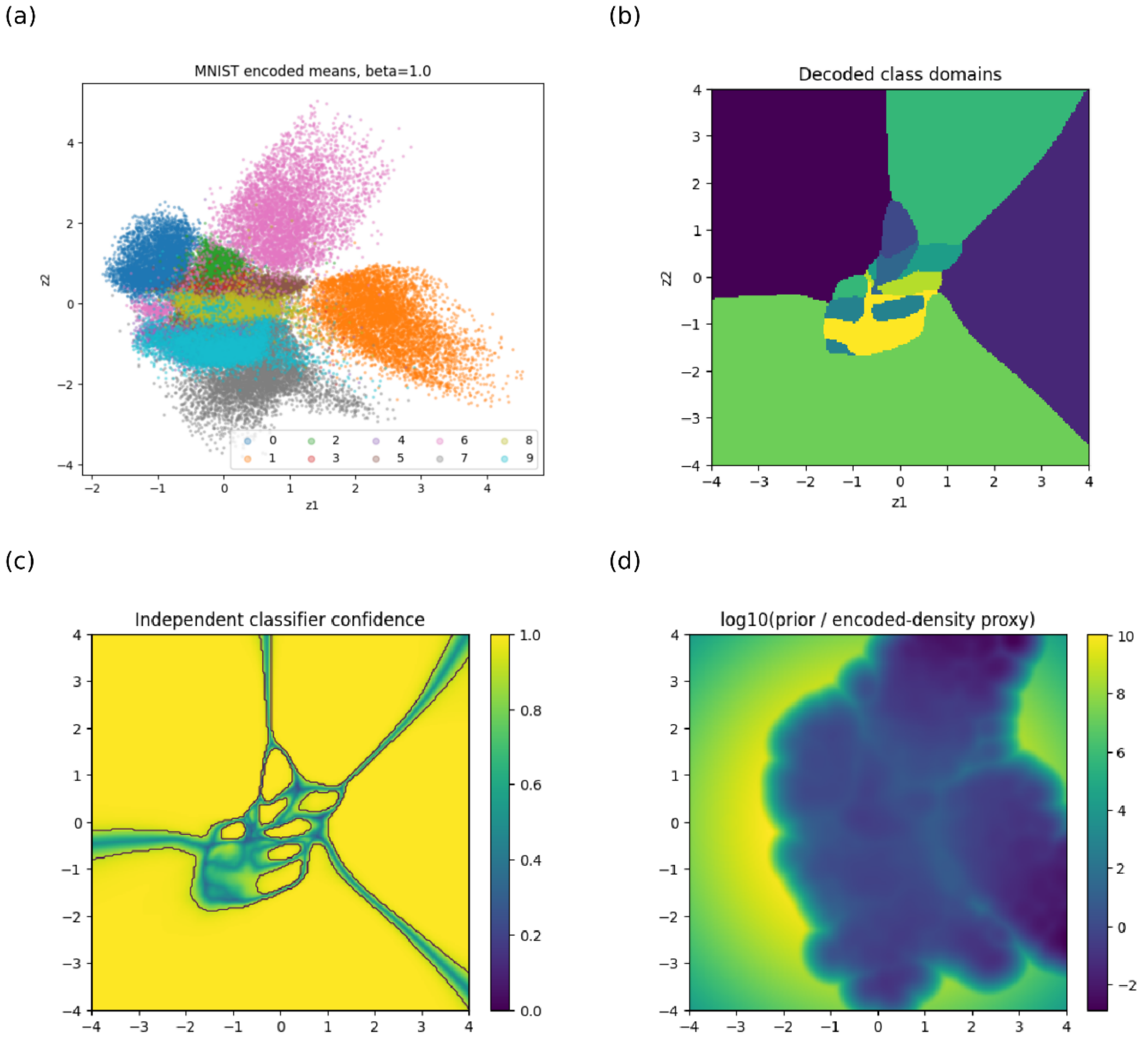

**Figure 1.** Representations of a two-dimensional MNIST VAE latent space. (a) Encoded posterior means, colored by digit identity. (b) Class labels assigned by an independent classifier to images decoded on a latent grid. (c) Maximum classifier probability for those decoded images. (d) $\log_{10}[p(z)/\hat{q}_\mu(z)]$, where $p(z)$ is the standard-Gaussian prior and $\hat{q}_\mu(z)$ is a kernel-density estimate of the encoded means. The encoded-mean density is not the full aggregate posterior, and classifier confidence is not a measure of physical validity.

Prior work addresses several parts of this problem. Results for continuous push-forward generators constrain representation of disconnected target supports [49–53], while Riemannian analyses identify metric distortion and failures of straight-line interpolation [54–57]. Aggregate-posterior mismatch motivates learned priors [40–44], and linear, dynamical and rate–distortion analyses characterize posterior collapse [45–48]. Symmetry-aware and topology-aware VAEs account for known transformations [7, 19, 24, 58], while dimensional compression also connects to folding in cortical-map and communication models [59–62]. These approaches do not by themselves determine all of the finite wall widths, rare-state regimes or covariance-dependent layouts studied here. The present analysis uses these established foundations to derive finite wall profiles, constrained packing and rare-state allocation, rather than proposing a new conditional-mean identity or a new interpretation of collapse as a transition [40, 45–47].

### 2.2 Objective and optimal decoder

As a first step, we analyze a simplified dataset in which each class is represented by a single prototype. There are K classes with frequencies $\pi_j^0$ and prototypes $a_j$. The latent prior is standard Gaussian in $d$ dimensions, the reconstruction likelihood is Gaussian with variance $\sigma_x^2$, and the KL contribution has weight β. Each prototype has one Gaussian class posterior $q_j$(z) and the aggregate distribution is their frequency-weighted mixture. The objective, divided by β and written per observation, is

$$\mathcal{L} = \frac{1}{2\beta\sigma_x^2}\mathbb{E}\|x - f(z)\|^2 + \mathrm{KL}\big(q(z \mid x) \parallel \mathcal{N}(0, I)\big) \tag{1}$$

The pairwise reconstruction coupling is

$$R_{ij} = \frac{\|a_i - a_j\|^2}{8\beta\sigma_x^2} \tag{2}$$

For two symmetric classes $\pm a$ this gives $R = a^2/(2\beta\sigma_x^2)$; small $\beta$ means large $R$. For a fixed encoder, minimizing squared reconstruction error at each latent point gives the conditional mean. This is result T1:

$$f^*(z) = \textstyle\sum_j r_j\,(z)\,a_j \tag{3a}$$

$$r_j(z) = \frac{\pi_j^0 q_j(z)}{\sum_i \pi_i^0 q_i(z)} \tag{3b}$$

The responsibilities $r_j$(z) are normalized probabilities of class membership at $z$; they are not the class densities $q_j$(z). Substituting the conditional mean into the reconstruction loss gives the conditional-variance, or Bayes-risk, expression

$$\mathcal{L}_{\mathrm{rec}} = \textstyle\sum_{i<j} 4\,R_{ij}\,\mathbb{E}_{\bar{q}}\big[r_i(z)\,r_j(z)\big] \tag{4}$$

The optimal decoder therefore blends the prototypes wherever their posterior distributions overlap. The transition between two dominant classes forms a wall separating their latent domains. For Gaussian class distributions, equality of the two weighted probabilities gives a quadratic equation in the latent coordinates. Equal covariances reduce it to a plane, whereas unequal covariances can produce curved boundaries and, when one class dominates only in a bounded region, an island. Across a locally two-class wall with linear log odds, the decoder follows a tanh profile. Its finite width is set by the overlap of the class distributions.

### 2.3 Affine stationarity and the packing constraint

An affine reparameterization $z \to Az + b$ of the latent leaves the reconstruction unchanged, because the encoder output layer and the decoder input layer absorb it and only the KL term changes. Setting its derivative with respect to $A$ and $b$ to zero at a stationary point gives the second result, T2:

$$\mathbb{E}_{\bar{q}}[z] = 0 \tag{5a}$$

$$\mathbb{E}_{\bar{q}}[zz^{\top}] = I \tag{5b}$$

The full matrix identity requires an encoder family that permits the corresponding affine transformation and a decoder that can compensate without changing reconstruction or introducing another penalty. For a diagonal posterior family, coordinatewise shifts and rescalings give zero aggregate mean and unit second moment in each coordinate, but do not force all cross-covariances to vanish. These are stationary-point conditions for an ideally trained VAE, not guarantees at a finite training checkpoint. A fixed smoothing scale in latent units breaks the rescaling freedom, as considered in Section S3.1.

For Gaussian prototype posteriors, the stationary moment condition reduces the average KL contribution to a frequency-weighted volume cost,

$$\mathrm{KL} = -\frac{1}{2}\sum_j \pi_j^0 \ln\det\Sigma_j + \mathrm{const} \tag{6}$$

The VAE therefore maximizes the frequency-weighted log volume of its class posteriors, subject to packing them into unit second moment and paying for wall overlap. For a shared isotropic width $s$, measure class positions $x_j = \mu_j / s$ in width units and define $V = \sum_j \pi_j^0 \left|x_j\right|^2$. Then

$$\mathrm{KL} = \frac{d}{2}\ln\left(1 + \frac{V}{d}\right) \tag{7}$$

Reconstruction penalizes overlap between classes, whereas regularization favors broader posteriors while keeping the overall latent spread fixed. Their competition defines a packing problem. Having zero mean and unit variance does not make the encoded distribution Gaussian, and separate clusters and sparsely populated boundaries can satisfy the same moment constraints. The derivation and the distinction between encoded means and complete posteriors are given in Section S1.

### 2.4 Connection to Flory–Huggins theory

The competition between reconstruction and regularization can now be written in the form of a mixing problem. The average class KL divergence separates into a penalty for the aggregate density and the mutual information between class identity and latent position [40]. Combining this relation with the reconstruction loss gives a local mixing term of the Flory–Huggins form:

$$\mathcal{L} = H(\pi) + KL(\bar{q} \parallel p) + \mathbb{E}_{\bar{q}}[f(r)] \tag{8a}$$

$$f(r) = \sum_{i<j} \chi_{ij}\, r_i r_j + \sum_j r_j \ln r_j \tag{8b}$$

$$\chi_{ij} = 4R_{ij} \tag{8c}$$

The local function $f(r)$ has the Flory–Huggins regular-solution form for monomeric mixing units [63, 64]. At a latent point $z$, the responsibility $r_j(z)$ is the probability that the point belongs to class $j$; it therefore plays the role of a local composition. Averaging this probability over all encoded points recovers the class frequency, $\int q(z) r_j(z)\, dz = \pi_j^0$, because $q(z) r_j(z) = \pi_j^0 q_j(z)$ and each class density integrates to one. This is the meaning of composition conservation here; it does not describe the transport of material through latent space. The couplings $\chi_{ij}$ measure the reconstruction cost of mixing classes, and increasing $\beta\sigma_x^2$ reduces these couplings, much as increasing temperature favors mixing. The decomposition itself does not require the stationary moment condition.

The equality in equations (8a)–(8c) is an exact rewriting of the stated VAE objective. It does not mean that the VAE can realize every composition pattern allowed in a classical mixture. With one Gaussian posterior per class, changing a mean or a covariance changes the responsibilities over the whole latent space; the composition cannot be chosen independently at every point. The total encoded density also varies with position and is penalized when it differs from the prior. These additional constraints explain why the coupling at which a separated state first becomes favorable can differ from the classical regular-solution value. For unequal class frequencies, separation can become favorable before the mixed state loses stability, producing the metastability discussed in Section 3.3.

Note that there is also no separately imposed Cahn–Hilliard gradient penalty in this objective [65]. The wall width follows from overlapping posteriors rather than from an independently specified interfacial energy. A small-noise expansion resembles a square-gradient term only when the decoder changes slowly over the width of a posterior and the residual–curvature contribution is negligible; it fails for sharp walls. Likewise, training moves encoder parameters, not locally conserved material. Familiar ideas such as curvature-driven coarsening or thermally activated nucleation therefore require additional assumptions. Sections S1.4 and S6 state which parts of the physical analogy follow from the model.

### 2.5 Identifying latent domains and walls

For the calculations below, we identify a wall as a region in which no class has responsibility greater than 0.9. In a two-class region, this is equivalent to an absolute log-odds value below ln9. This threshold defines class ambiguity, not physical impossibility. The aggregate density $q(z)$ measures how frequently latent points are sampled by encoding the observed data, including the encoder noise. The prior $p(z)$ instead determines where unconditional generation samples. Comparing the two reveals regions that generation visits often even though the observations provide little support there.

An independent classifier provides a separate test of the decoded output: how readily does it recognize the generated image? Low confidence can reflect a mixture of classes, blur or another decoding error, so it is not the same as overlap between encoded class distributions. Scientific usefulness requires the relevant physical constraints or measurements. Property prediction introduces another distinction. A dataset may contain many images from a latent region while only a few of those images have measured property values. Good coverage by images therefore does not guarantee good coverage for a property predictor.

In Figure 3, class densities are approximated by Gaussian fits to encoded means. These fits describe how the centers of the per-image posteriors are distributed; they omit the uncertainty around each center. Adding encoder variance can change the inferred overlap substantially, especially near collapse. We therefore distinguish the encoded-mean proxy used in the figure from the variance-inclusive control in Section S2.2. Section S7.3 gives the estimator definitions.

## 3. Salient features of VAE latent spaces

We use the model to examine the formation and width of domain walls, the loss of class separation through posterior collapse, and the arrangement of common and rare classes in latent space. We begin with two equally frequent classes, then vary class frequency and structural dissimilarity before considering several classes together. The same analysis explains why a rare state can remain unresolved, form an isolated domain or become represented within the wall between two common states.

### 3.1 The two-class solution

We begin with the minimal non-trivial case of two equally probable prototypes in one latent dimension. It is exactly solvable and separates the two quantities that control the rest of the paper: the domain separation set by the encoder and the finite transition width imposed on the decoder. Take prototypes $\pm a$, a one-dimensional latent and $q(z \mid \pm a) = \mathcal{N}(\pm m, s^2)$. T1 gives the decoder exactly:

$$f^*(z) = a\tanh\left(\frac{z}{\ell}\right) \quad (9a)$$

$$\ell = \frac{s^2}{m} = \frac{1}{\sqrt{k(1+k)}} \quad (9b)$$

$$k = \frac{m^2}{s^2} \quad (9c)$$

Minimizing the KL at fixed $k$ gives $s^2 = 1/(1+k)$ and $m^2 = k/(1+k)$, so that $m^2 + s^2 = 1$ as T2 requires. The loss reduces to one variable,

$$\mathcal{L}(k) = R\, g(k) + \frac{1}{2}\ln(1+k) \quad (10a)$$

$$g(k) = \mathbb{E}_{\xi}\left[\mathrm{sech}^2\left(k + \sqrt{k}\,\xi\right)\right] \quad (10b)$$

For large $k$, $g(k) \approx \sqrt{\pi/2}\, k^{-1/2} e^{-k/2}$; the prefactor follows from $\int e^x \mathrm{sech}^2 x\, dx = \pi$. The optimum satisfies

$$k^* = 2\ln R + \ln k^* + \ln\frac{\pi}{2} \quad (11a)$$

$$\ell \approx \frac{1}{2\ln R} \quad (11b)$$

The squared separation measured in posterior-width units increases only logarithmically with reconstruction coupling. Consequently, lowering β sharpens the wall but does not eliminate it at finite coupling. The domain half-width approaches one, while the transition width decreases. Figure 2(a) shows the decoder profiles, Figure 2(b) compares the numerical optimum with the strong-separation asymptotes, and Figure 2(c) resolves the approach to collapse. Section S2.1 gives the derivation; the dynamics in Figure 2(d) are discussed in Section 4.5.

For small $k$, $g(k) = 1 - k + k^2 + O(k^3)$. Eliminating $s$ gives a Landau form in the domain separation $m$:

$$\mathcal{L} \approx R - \left(R - \frac{1}{2}\right) m^2 + (R - R^2) m^4 \quad (12)$$

For equal Gaussian prototypes the separated state emerges continuously at $R_c = 1/2$, corresponding to $\chi_c = 2$. The domain separation has a square-root onset and the wall width has an inverse-square-root divergence. These are mean-field exponents of the restricted variational problem, in which the order parameter is a global parameter rather than a fluctuating spatial field. In the multicomponent model the first unstable direction is selected by the largest eigenvalue of the frequency-weighted prototype covariance. This connects the calculation to linear-VAE collapse and deterministic-annealing clustering [45–48, 66]. The Bernoulli two-prototype calculation retains the asymptotic wall mechanism but shifts the collapse threshold in its own normalization.

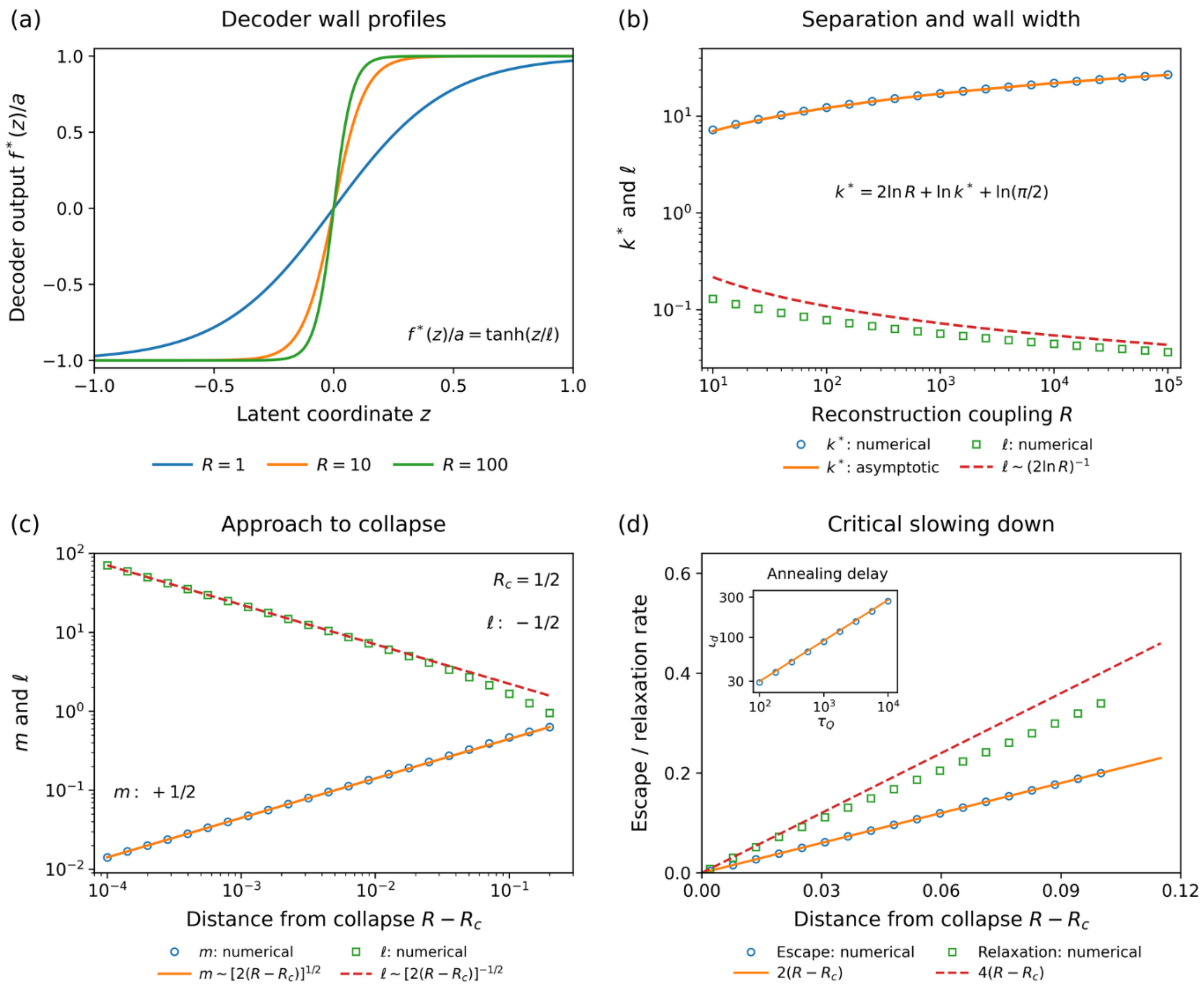

**Figure 2.** Two-class Gaussian-prototype solution. (a) Normalized optimal decoder profiles for $R = 1$, 10 and 100. (b) Optimal squared separation in posterior-width units, $k^* = m^2/s^2$, and intrinsic wall width $\ell$ versus $R$; symbols show numerical optima and lines show the strong-separation approximations in equations (11a) and (11b). (c) Domain half-separation $m$ and wall width near $R_c = 1/2$, plotted against $R - R_c$, with square-root and inverse-square-root guides. (d) Growth rate of a small separation away from the collapsed state and relaxation rate toward the separated optimum. The inset shows the delayed onset under the specified ramp of $R$. These rates describe the reduced gradient-flow model of Section 4.5, not neural-network training time.

As an illustration, we next examine a trained VAE using handwritten digits 0 and 1 from the scikit-learn digits dataset (Figure 3). Each image is encoded by a two-dimensional Gaussian, and the model is trained at several regularization strengths from three independent initializations. Gaussian fits to the encoded means define the displayed domains and their ambiguous boundary. Decoding along the line between the class means shows how the output changes from one digit to the other. Architecture, training settings and the full measurements are given in Section S2.2.

Increasing regularization brings the two class means closer together, until their separation is very small. The width measured from the encoded means, however, also decreases rather than following the broadening predicted for complete prototype posteriors. Figure 3 therefore illustrates the loss of class separation in a trained model, but it is not a quantitative confirmation of the analytical wall-width law. The distinction between the two measurements is important near collapse. Contraction of class means alone does not establish that every per-image posterior has collapsed to the prior.

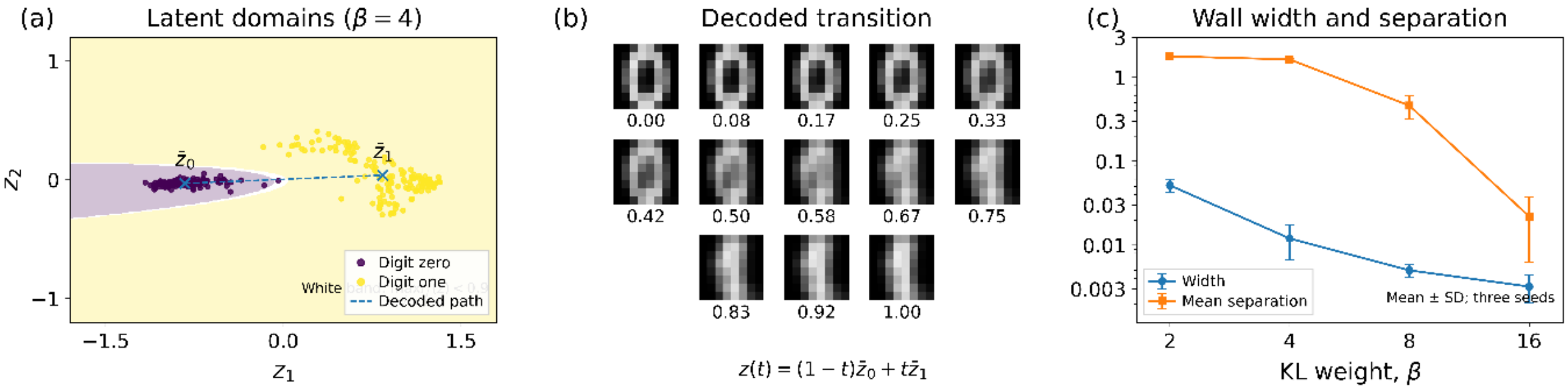

**Figure 3.** Trained two-class digit benchmark. (a) Encoded means for digits 0 and 1 at $\beta = 4$ for seed 17, with domains obtained from regularized Gaussian fits. The white band has maximum fitted responsibility below 0.9; the dashed segment joins the class means. (b) Actual decoded images at 13 equally spaced positions on that segment, read left to right and then top to bottom; $t$ labels the interpolation position. (c) Median local wall width of the encoded-mean proxy and separation of the class means versus $\beta$, on a logarithmic distance axis. Points and error bars give means and sample standard deviations over three seeds. Both quantities decrease as β increases. The fitted width does not include encoder variance and should not be interpreted as the width of the full aggregate posterior.

The discrepancy in wall width is very informative. Near collapse, the encoder means can contract into a small region while the uncertainty of each encoded observation remains broad. Fitting only the means can then produce a narrow apparent transition even though the full posteriors strongly overlap. Finite training and within-class variability also distinguish this experiment from the optimized prototype model.

Classifier confidence tests a different aspect of the result, i.e. whether a decoded image resembles a recognizable digit. A poorly recognized output need not lie in the responsibility-defined wall, and neither diagnostic alone establishes physical invalidity. Aggregate moment tests likewise must include encoder variance, rather than only the spread of encoded means. The estimator definitions are given in Section S7.3.

For two classes the optimum does not depend on the latent dimension. Additional coordinates carry no information about the class distinction, remain matched to the prior and extend the wall into a line in two dimensions and a plane in three (Section S3.1). The separation, wall width and collapse threshold are therefore those of the one-dimensional calculation.

### 3.2 Frequency, dissimilarity and multiclass packing

We next relax equal class frequency while keeping two prototypes and ask how the moment constraint redistributes latent position, posterior width and prior volume when one state is rare. With $\pi_1^0 \neq \pi_2^0$, T2 fixes the centroid at zero, so $\mu_1 = \pi_2^0 \Delta$ and $\mu_2 = -\pi_1^0 \Delta$: the rare class sits in the tail of the prior. With free widths each class must sit about $\sqrt{k^*}$ of its own widths from the wall, and the width problem becomes

$$\max \textstyle\sum_j \pi_j^0 \ln s_j \tag{13a}$$

$$\text{subject to } \pi_1^0 \pi_2^0 t^2 (s_1 + s_2)^2 + \textstyle\sum_j \pi_j^0 s_j^2 = 1 \tag{13b}$$

Here $t^2 \approx k^*$. In the separated two-class asymptotic, the posterior-width ratio approaches the class-frequency ratio. The rare class becomes narrower and moves outward to maintain the weighted centroid. With unequal widths its dominant-responsibility region can close on the far side, producing a tail island. The prior mass of that island need not equal its frequency in the observations: at $R = 10^3$, a class with 5% of the data receives only 0.2% of the prior (Section S2.5). Rare states that resemble blends of two frequent classes are analyzed in Section 3.3.

Structural dissimilarity determines which classes are likely to be neighbors as well as how sharply they are separated. Confusing two different prototypes incurs a larger reconstruction cost, which favors a larger separation between their posteriors. Figure 4(a) shows this effect as a soft repulsion between class centers. The increase in preferred separation is much smaller than the increase in dissimilarity, so latent distance is not a calibrated measure of image difference. When only one coordinate is available, the model can also reduce the loss by placing similar classes next to one another and avoiding a direct boundary between the most dissimilar pair.

In the two-class model, unequal frequencies compress the less common class and move it toward the prior tail, Figure 4(b). Panels (c–e) extend this picture to handwritten-digit prototypes. For several digit classes in two dimensions, similar mean images tend to share a boundary (c); in one dimension, their ordering determines which pairs must meet (d). The separation–dissimilarity plot in (e) compares neighboring pairs with the two-class prediction. These are optimized class-aggregate representations of mean images, not latent maps learned from every handwritten example. Their shared isotropic posterior width is an explicit restriction; the effects of anisotropic posteriors are discussed in Section 4.2.

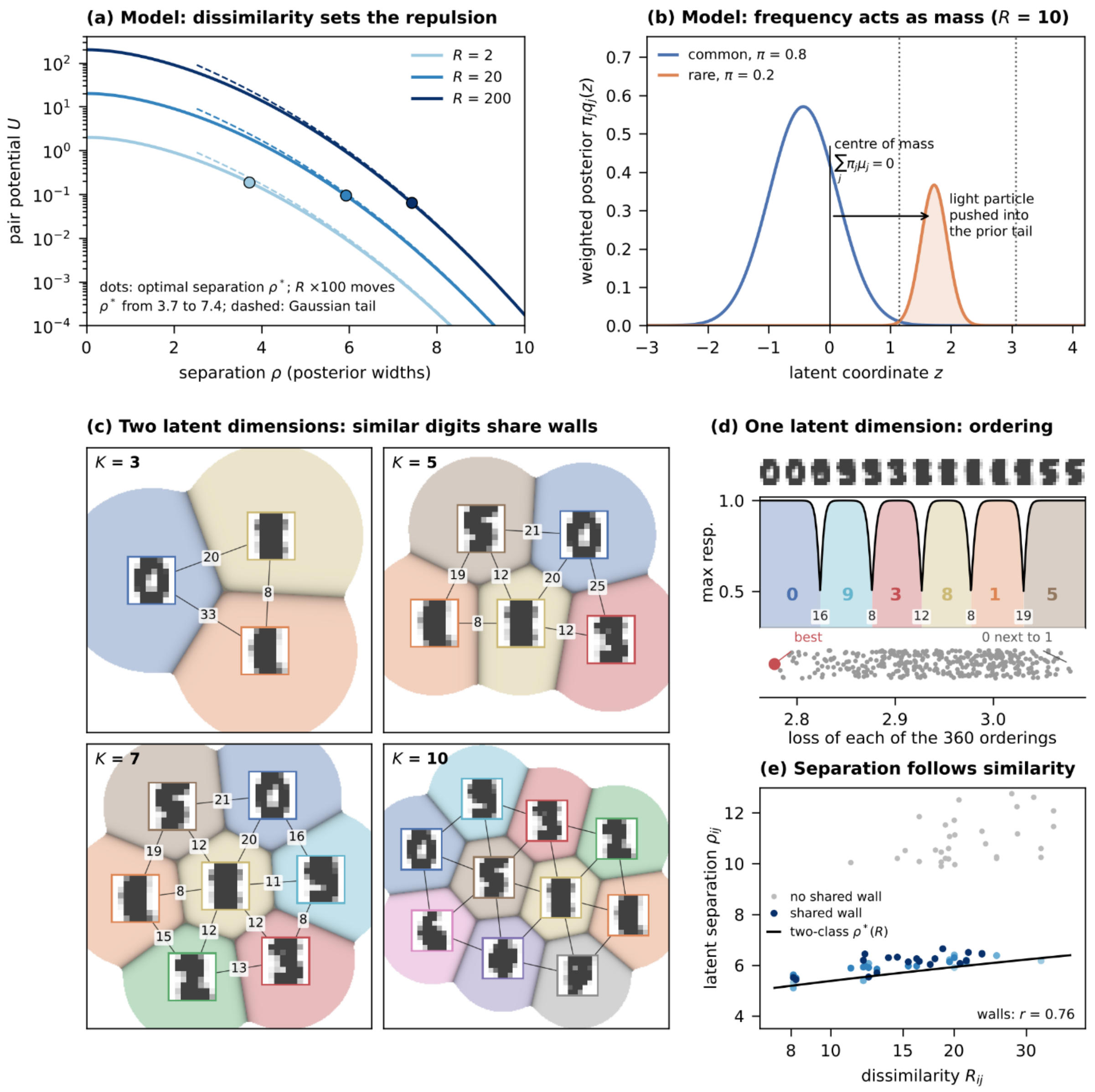

**Figure 4.** Class frequency and structural dissimilarity organize the latent layout. (a) Pair cost $U = R\,g(\rho^2/4)$ versus separation $\rho$ in posterior-width units for $R = 2$, 20 and 200. Solid curves are the calculated pair costs, dashed curves show their Gaussian-tail approximations, and points mark the separation minimizing pair overlap plus prior confinement. (b) Weighted posteriors for classes of frequencies 0.8 and 0.2 at $R = 10$, with separately optimized widths. The less frequent class forms a narrow tail island bounded by two responsibility crossings. (c–e) Class-aggregate calculations in which each digit is represented by its mean image, with shared isotropic width and median pair coupling set to 20. (c) Lowest-loss two-dimensional layouts found among the tested starts for three, five, seven and ten digits. Shaded domains show the dominant class; lines join classes sharing a wall, with thicker lines for more similar pairs. Insets show the decoded class-center images. (d) The best one-dimensional ordering of six digits, with decoded images and maximum responsibility above the losses of all 360 distinct orderings. (e) Pair separation versus prototype dissimilarity. Neighboring pairs follow the two-class trend more closely than pairs that do not share a wall. These calculations test the reduced model, not a trained image encoder.

The connection to particles confined in a trap provides a useful way to understand these layouts [67]. Reconstruction favors separation, while the prior limits the overall extent of the class centers. Note that the Gaussian multi-bubble result [68] concerns a different perimeter objective and does not certify these layouts. The analogy does not turn the boundaries into freely moving material interfaces: their locations follow from the chosen posteriors. Section S2.4 gives the pair-potential derivation, junction corrections and full numerical comparisons.

The following reduction of the objective is exact for shared isotropic widths; pairwise additivity is a separate strong-separation approximation. Write the class means in width units, $x_j = \mu_j/s$. The reconstruction cost then depends on the $x_j$ alone, and stationarity in $s$, the trace of T2, gives $s^2 = \left(1 + d^{-1}\sum_j \pi_j^0 \left|x_j\right|^2\right)^{-1}$. The loss becomes

$$\mathcal{L} = \Phi(x) + \frac{d}{2}\ln\left(1 + \frac{1}{d}\sum_j \pi_j^0 \left|x_j\right|^2\right) \tag{14}$$

where $\Phi$ is the reconstruction cost evaluated with unit-width posteriors $q_j = N(x_j, I)$. For two equal classes at $x = \pm k^{1/2}$ in one dimension, (14) is the two-class loss $R\, g(k) + 1/2\ln(1+k)$. At fixed $s$ the KL term is a harmonic trap of stiffness $\pi_j^0 s^2$ on class $j$, so class frequency acts as mass; eliminating $s$ makes the confinement logarithmic in the frequency-weighted moment of inertia. Where at most two posteriors overlap, $\Phi$ is a sum of pair potentials of the separations $\rho_{ij} = \left|x_i - x_j\right| = 2k_{ij}^{1/2}$. Each equals $4R_{ij}$ times the overlap $\int \pi_i^0\pi_j^0 q_i q_j/\left(\pi_i^0 q_i + \pi_j^0 q_j\right) \mathrm{d}u$, and for well-separated classes

$$U_{ij}(\rho) \approx 2\sqrt{2\pi}\, R_{ij}\, \frac{\sqrt{\pi_i^0\pi_j^0}}{\rho}\, \exp\left[-\frac{\rho^2}{8} - \frac{\ln^2\left(\pi_i^0/\pi_j^0\right)}{2\rho^2}\right] \tag{15}$$

with $R_{ij} = \left|a_i - a_j\right|^2/8\beta\sigma_x^2$; for $K$ equal classes the pair integral gives $U_{ij} = \left(2R_{ij}/K\right) g(\rho^2/4)$ exactly. Structural dissimilarity sets the strength of the repulsion, the posterior width sets its range, and unequal frequencies weaken it.

The force $-U_{ij}{}' \approx (\rho/4)\, U_{ij}$ is a soft repulsion with a Gaussian tail, and three properties of the layouts follow. First, at any stationary point (14) obeys the virial theorem

$$\sum_j x_j \cdot \nabla_j \Phi = -\sum_j \pi_j^0 \left|\mu_j\right|^2 = -d\,(1 - s^2) \tag{16}$$

so the total work of the forces is bounded by the latent dimension. With Gaussian tails a bounded virial fixes contact separations $k_{ij} \approx 2\ln\left(R_{ij}/R_0\right)$, with $R_0$ set by the trap: this is the logarithmic compression of dissimilarity, and for two classes it is the law for $k^*$. Second, classes two contacts apart interact with a weight of order $R^{-3}$ relative to neighbors, so a one-dimensional chain is a nearest-neighbor system and its ordering is the seriation problem above. Third, departures from pair additivity are confined to junctions and lower the loss. Relative to a wall they carry weights $e^{-k/6}$ at a triple junction and $e^{-k/2}$ at a fourfold junction, so a fourfold junction is controlled almost entirely by pair forces, consistent with its continuous splitting when their symmetry is broken. The repulsion is steep but not hard: its effective inverse-power exponent, $1 + k$, is 8–13 for $R = 10$–$100$, between Coulomb particles and hard disks. Pair potentials alone reproduce the shell sequence of the full optimization for three to ten equal classes at $R = 20$ (Section S2.4).

Near collapse the picture changes. All posteriors overlap, the forces become long-ranged and linear, and the second variation of Eq. (14) at $x = 0$ gives the instability condition $\lambda_{\max}(C_{\pi^0}) > \beta\sigma_x^2$ of linear VAEs, with $C_{\pi^0}$ the frequency-weighted covariance of the prototypes; the unstable mode places the classes along its leading principal axis. Dissimilarity therefore enters the layout linearly at onset, as in classical multidimensional scaling, and logarithmically once the classes are separated. Two effects lie outside the pair picture. An intermediate rare class lowers the conditional variance inside a wall, a three-body attraction that produces the adsorbed and layer states of rare classes. Anisotropic posteriors make $U_{ij}$ depend on the orientation of $x_i - x_j$ relative to the latent axes, which favors stack-like structures in latent space (Section 4.2).

The layouts above are computed in one and two latent dimensions. The pair potential, the logarithmic contact separations and the onset condition hold in any dimension, but the arrangement changes. In $d$ dimensions $s$ classes generically meet on a set of dimension $d - s + 1$ (Section S3.4): in two dimensions walls meet at triple points and equal classes form planar shells, whereas in three dimensions walls meet along triple lines, four classes can meet at a point, and larger sets are expected to form three-dimensional shells. In the cell-filling limit, up to $d + 1$ equal classes take the simplicial arrangement in which every pair shares a wall [68], a tetrahedron in three dimensions. Additional coordinates therefore relax the ordering constraints of one and two dimensions and allow more pairs to share walls, so the layout follows the logarithmic dissimilarities more closely.

### 3.3 Rare classes: coexistence, metastability and behavior at walls

Two questions must be separated for a rare class: when would a distinct domain give a lower loss, and can training reach that domain from a mixed representation? These need not have the same answer. A small displacement can increase the loss around a mixed state even after a distant separated state has become preferable. The model is then metastable: local optimization can remain in the mixed state unless it finds a sufficiently different configuration. For a rare class of frequency $\pi^0$, the binodal marks the change in the lowest-loss state, while the spinodal marks the loss of local stability of the mixed state:

$$\chi_b \approx 1.6\ln(1/\pi^0) \tag{17a}$$

$$\chi_s = \frac{1}{2\pi^0(1-\pi^0)} \tag{17b}$$

The two thresholds leave an interval in which the preferred representation is separated but the mixed state is still locally stable. In the Gaussian-encoder calculation, separation becomes favorable at a larger coupling than in an unconstrained regular solution because precise posteriors and their remaining overlap also contribute to the loss. The interval grows as the class becomes rarer. This explains why initialization can affect whether a rare state is resolved; it does not determine how often a noisy neural optimizer escapes the mixed state. The prefactor in equation (17a) is the computed ratio $\chi_b/\ln(1/\pi^0)$, which is 1.62 at $\pi^0 = 0.01$ and 1.69 at $\pi^0 = 0.03$; Table S1 and Section S2.5 give the binodal and spinodal values.

In the nonparametric calculations, already separated majority classes help a rare class find a distinct position. An additional latent coordinate can also let it move away without crossing the majority-class walls. Merely appending a coordinate that carries no information leaves the loss unchanged, whereas allowing that coordinate to become informative enlarges the set of possible representations. Section S2.5 gives examples of both.

A different situation arises when the rare class resembles the blends already decoded at a wall. For frequent prototypes A and B, the optimal decoder traces the segment joining their images. We locate a rare prototype C by its projection $t$ onto this segment and by the remaining perpendicular mismatch $h$. If $0 < t < 1$, a suitable blend can approximate C better than either pure endpoint. The improvement relative to the nearer endpoint, in the loss units of the model, is

$$\Delta_{\text{wall}} = \min(t, 1-t)^2\, \chi_{AB} \tag{18}$$

Figure 5 illustrates the resulting possibilities using a circle and a square as frequent states and a rounded square or an elongated rhombus as the rare state. The local mixing curves in (a) describe the tendency to mix or separate, not the full constrained phase diagram, while the prototype geometry in (b) shows why the rounded square benefits more from an A/B blend. The decoded paths and difference images in (c) distinguish dissolution, a tail island, adsorption within the wall and an intervening layer. A state adsorbed at the wall changes its decoded content without acquiring a separate domain; a layer instead replaces the A/B transition by A/C and C/B transitions. The denser regime map in (d) shows that frequency and coupling can select different outcomes even for the same rounded-square prototype. Projection onto the segment alone is therefore not a complete classification rule.

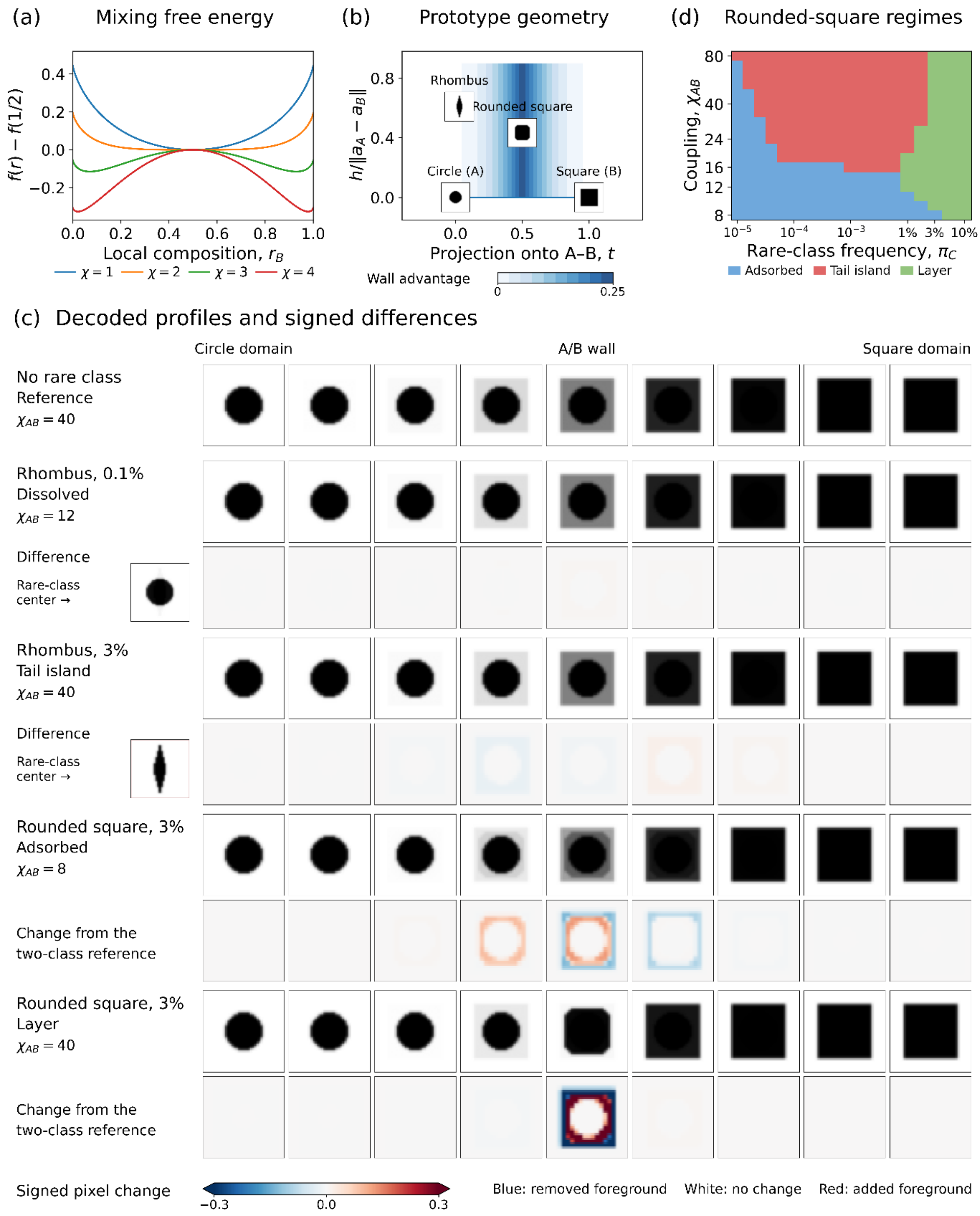

**Figure 5.** Rare-class allocation near a latent wall. (a) Regular-solution mixing term $f(r) = \chi r(1-r) + r\ln r + (1-r)\ln(1-r)$, shifted by $f(1/2)$, for several interaction strengths. The symmetric mixing threshold is $\chi_c = 2$. (b) Rare-prototype projection $t$ and perpendicular mismatch relative to the circle–square segment; shading denotes the geometric wall advantage. (c) Decoded paths from the circle domain through the wall to the square domain. Each difference row compares the rare-class model with the two-class reference at the same coupling, at corresponding wall-relative positions. Red denotes added foreground, blue removed foreground and white little change; decoded and difference images use square display fields. The side thumbnails show the decoded rare-class center for the dissolved and island cases. (d) Rounded-square regimes on an 18 × 18 frequency–coupling grid, each point classified from the lowest-loss result among the tested optimization starts. The labels describe the representation under the stated reconstruction metric, not physical phase transitions.

To test the role of the reconstruction metric, we repeat the rare-prototype calculation after smoothing all prototype images with a Gaussian of four pixels. This makes the rounded square much closer to a circle–square blend: its squared perpendicular mismatch decreases from about 0.19 to 0.017 of the squared circle–square distance. At the same coupling, several frequencies that produce a tail island or a layer under pixelwise comparison instead produce adsorption under the smoothed comparison. The rare rhombus can likewise become dissolved rather than isolated. Section S2.6 reports the individual cases. Thus, the metric can change whether a rare feature is kept distinct or represented as part of an existing transition.

Prior probability is also regime-dependent. A distinct rare class can receive much less prior mass than its observation frequency, whereas an intervening layer at frequency 0.03 occupies 5.6% of the prior at $\chi_{\mathrm{AB}} = 40$. Near collapse, a broad wall can accommodate intermediate content by adsorption without a separate layer. The physical terms adsorption and wetting describe these representation-level patterns [69]; they do not show that decoded mixtures are actual transition configurations. This qualification matters for discovery: excluding every ambiguous wall can exclude useful intermediate states as well as poorly supported blends.

## 4. The role of latent dimensionality

The prototype model separates class identity from the continuous variations that occur within a class. We now ask how those variations are represented when only a limited number of latent coordinates is available. The answer depends on the posterior covariance, the symmetry of the data and the relative importance of the factors to be reconstructed. We consider continuous factors and rotations first, then the folding of an additional factor into an occupied coordinate and the way these structures develop during training.

### 4.1 Continuous variation within a domain

The previous cases treat each class as a prototype. To distinguish discrete class separation from genuine continuous variation inside a class, we add one linear internal factor and solve how its variance is divided between latent signal and posterior noise while the class as a whole still participates in the global packing.

Let class $j$ vary along one linear factor with data variance $w_j^2$. For the linear factor and negligible interclass overlap assumed here, the internal variance split decouples from class packing. The class-aggregate width $S_j$ is fixed by packing, while the split of $S_j$ into signal $c_j$ and noise $s_j$ is fixed by the class alone:

$$\rho_j = \frac{c_j^2}{s_j^2} = 2A_j - 1 \tag{19a}$$

$$A_j = \frac{w_j^2}{2\beta\sigma_x^2} \tag{19b}$$

The internal factor collapses on its own when $A_j < 1/2$. Inside a domain the decoder is linear with slope $(w_j/S_j)\sqrt{1 - 1/(2A_j)}$; at a wall it is the tanh profile as derived in Section 3 with width set by $S_j$. The ratio of decoder Jacobians between wall and domain is about $(\|\Delta a\|/2w_j)\sqrt{2\ln R}$. Continuous variability inside a class changes the local decoder slope, but by itself it does not create an isolated latent island. Islands arise when the class frequency or its contacts with neighboring classes make a closed region favorable (section 3).

The internal factor and the class boundary therefore have different roles. One governs variation within a family; the other governs ambiguity between families. A factor can be suppressed by regularization even while class identity remains resolved. Conversely, learning a continuous factor does not guarantee that it receives an axis independent of class identity. Posterior-family comparisons below make that distinction explicit.

### 4.2 Anisotropy and axis alignment are not semantic disentanglement

Within the idealized isotropic-prior model, a full-covariance posterior can rotate together with the latent map and a compensating decoder. A diagonal covariance restriction removes this freedom. For a locally linear decoder with pulled-back metric G and effective likelihood scale τ, the posterior precision is $P = I + G/\tau^2$. The difference between the diagonal and full-covariance precision costs is

$$\Delta = \frac{1}{2}\left[\sum_a \ln P_{aa} - \ln\det P\right] \geq 0 \tag{20a}$$

$$P = I + \frac{G}{\tau^2} \tag{20b}$$

The nonnegative gap is the mean-field variational penalty. It vanishes when the precision is diagonal and favors local principal directions of the decoder metric [70]. At a sharp wall, the metric is dominated by the wall-normal direction. Reoptimizing the two-class wall at each orientation gives the distinct orientation-dependent excess cost

$$\Delta\gamma(n) \simeq -\ln\max_a |n_a| \tag{21}$$

The extra loss of a wall at 45° in two dimensions rises from 0.001 near collapse to 0.281 at R = 10 and approaches the strong-separation limit of one-half ln 2 (Figure S3(a)). This provides a quantitative axis-alignment mechanism. It does not identify an axis with a named physical factor, which would require additional information or assumptions [6, 8].

Encoding a class label along a coordinate already used for continuous variation can cost less than creating a separate active axis. Such a representation can be locally orthogonal while still mixing class identity and the continuous factor. Axis alignment therefore does not establish semantic disentanglement. Sections S3.2 and S3.4 give the covariance-family and class–factor examples in full.

These optimized model layouts need not match the mosaics observed in the ten-class digit map. Within-class variation, the encoder family and optimization history can change the balance between candidate layouts. The covariance-annealing calculations in Section S3.2 illustrate this dependence on the training path, but do not establish a general training prescription for image VAEs.

The orientation cost above is evaluated for a wall in two dimensions. In *d* dimensions the excess cost of a wall whose normal lies along a body diagonal approaches one-half ln *d* (Section S3.2), so the preference for axis-parallel walls strengthens with dimension. In three dimensions diagonal posteriors therefore still favor stacks along a single axis with passive remaining coordinates, whereas isotropic posteriors favor compact arrangements such as a tetrahedron of four classes.

### 4.3 Rotations, quotient orbits and topology tests

Discrete classes are only one source of latent geometry. A continuous physical transformation, such as rotation, traces an orbit rather than jumping between classes. If a group G acts on a state and H is the subgroup that leaves that state unchanged, then the physically distinct poses lie on the orbit $G/H$. For in-plane rotation G = SO(2); an object with exact n-fold rotational symmetry has H = $C_n$ because rotations by $2\pi/n$ are equivalent. A VAE that must represent the pose therefore inherits this orbit structure in its latent coordinates, while an invariant model aims to remove the orbit from the content representation.

Let a class vary by rotation, and expand the image in angular harmonics of the angle that the encoder can resolve: the full rotation angle for an asymmetric pattern and the quotient angle for a symmetric one (below). Let $A_m$ be the image variance carried by harmonic $m$ divided by $2\beta\sigma_x^2$, and $C = \sum_m A_m m^2$. Within the isotropic ring ansatz, the two-dimensional encoding is a ring of radius $\rho \to \sqrt{2}$ with posterior width $s^2 \to 2/C$, again as T2 requires. The posterior over angle at a latent point is a von Mises distribution with concentration $\kappa = \rho|z|/s^2$, so the decoder outputs the image smoothed by that distribution (Figure 6). Near the center $\kappa$ is small and the decoder returns the rotational average, a vortex-like core with prior mass

$$P_{\text{core}} \approx \frac{\kappa_{1/2}^2}{C^2} \propto \beta^2 \tag{22}$$

Here $\kappa_{1/2}$ is the concentration at which the leading angular harmonic is half resolved. The core is a region of unresolved orientation, not a boundary between unrelated classes. In the ring asymptotic its prior mass decreases quadratically with β, whereas discrete wall mass decreases only logarithmically. The comparison distinguishes mechanisms rather than asserting that every rotational model has the same core geometry.

For an exactly n-fold symmetric pattern the encoder cannot distinguish orientations separated by $2\pi/n$ , so the relevant variable is the quotient angle rather than the full rotation angle. Within the ring ansatz, a faithful degree-one representation of the quotient orbit is traversed n times as the physical angle completes one turn. Symmetry alone fixes periodicity: a loop avoiding its winding center has a winding that is an integer multiple of n, not necessarily n. Re-expressed in this quotient coordinate, this ring branch has the same continuous onset as the one-fold problem for every n. A first-order onset would arise only if the full angle were identifiable, which a symmetric pattern does not allow. The vortex-like core remains because the center of the ring carries little or no orientation information. Its size is controlled by how sharply the quotient angle is resolved: better angular resolution makes the ring better defined and shrinks the region in which the decoder can return only the rotational average.

An explicit angle latent changes the problem because the known pose no longer has to be encoded indirectly in the generic content coordinates. With a circular angle variable and periodic decoder, every angle corresponds to a valid pose, the aggregate angle posterior can match the circular prior, and the remaining content latent can describe only the rotation-independent structure. This is the principle behind spatial and rotationally invariant VAEs used in imaging [7, 17–24, 71]. A real-valued Gaussian angle decoded modulo $2\pi$ can still generate every orientation, but a globally faithful scalar angle coordinate requires a cut; continuous non-invertible maps to the real line are possible [72, 73]. For n-fold symmetric data, a cut-free pose variable must respect the quotient SO(2)/$C_n$ rather than treating all $2\pi$ orientations as distinct.

The situation changes when the symmetry is only approximate. Suppose the object is mainly *n*-fold symmetric but contains a weak feature that distinguishes the *n* otherwise equivalent orientations. The encoder then faces two competing descriptions: it can ignore the weak feature and keep the quotient representation, or it can resolve the full orientation. These alternatives have different winding numbers and cannot be continuously deformed into one another without passing through the center of the latent ring. Resolving the full orientation lowers the loss only when the distinguishing feature is strong enough, above a threshold that grows as about $2 \ln n$ (Section S3.3). Partial resolution by a regular plane encoding curve in the wrong winding sector can require self-intersections [73]. If the corresponding full posteriors coincide, the decoder averages the distinct orientations; coincident means alone do not imply this when covariances differ. If an explicit angle head is locked to the quotient rather than the full orientation, the unresolved distinction reappears in the content latent as n cyclic domains. Their couplings have the form of an $n$-state clock model, with a continuous onset for $n = 2$ and 4 and a weakly first-order onset for $n = 3$. Section S3.3 gives the derivations.

Figure 6 illustrates these rotational structures on rotated handwritten digits. Each digit class is augmented by random in-plane rotations and encoded by a two-dimensional VAE (Section S7.3). For digits 1, 0 and 8 the encodings form a ring around a core that decodes to a rotational average, and a base image used in training rotated through 360° winds twice about the class center ($|w| = 2$ in 97–100% of the rotation sweeps). Orientations 180° apart are encoded close together, at mean distances of 0.17–0.67 against 2.7–2.8 for orientations 90° apart, so these approximately twofold-symmetric digits are represented on the quotient ring. The weak feature that distinguishes the two halves of the orbit is resolved only in part: for digit 1 the reconstruction error against the 180°-rotated image is three times that against the input (37.3 against 12.7), whereas for digit 0 the two are nearly equal (6.8 against 6.3). Digit 3, which lacks even approximate twofold symmetry, also winds twice in the four-digit mixture, while its decoded images still distinguish the two halves of the orbit. In the mixture the classes form nested rings whose radii increase from 0 (0.40) through 3 and 8 to 1 (2.24), so orientation and class identity share the two latent coordinates. Because the sweeps reuse training-image identities, they diagnose orientation encoding on those images rather than generalization to new images. The double winding of digit 3 also shows that learned winding is not determined by object symmetry alone.

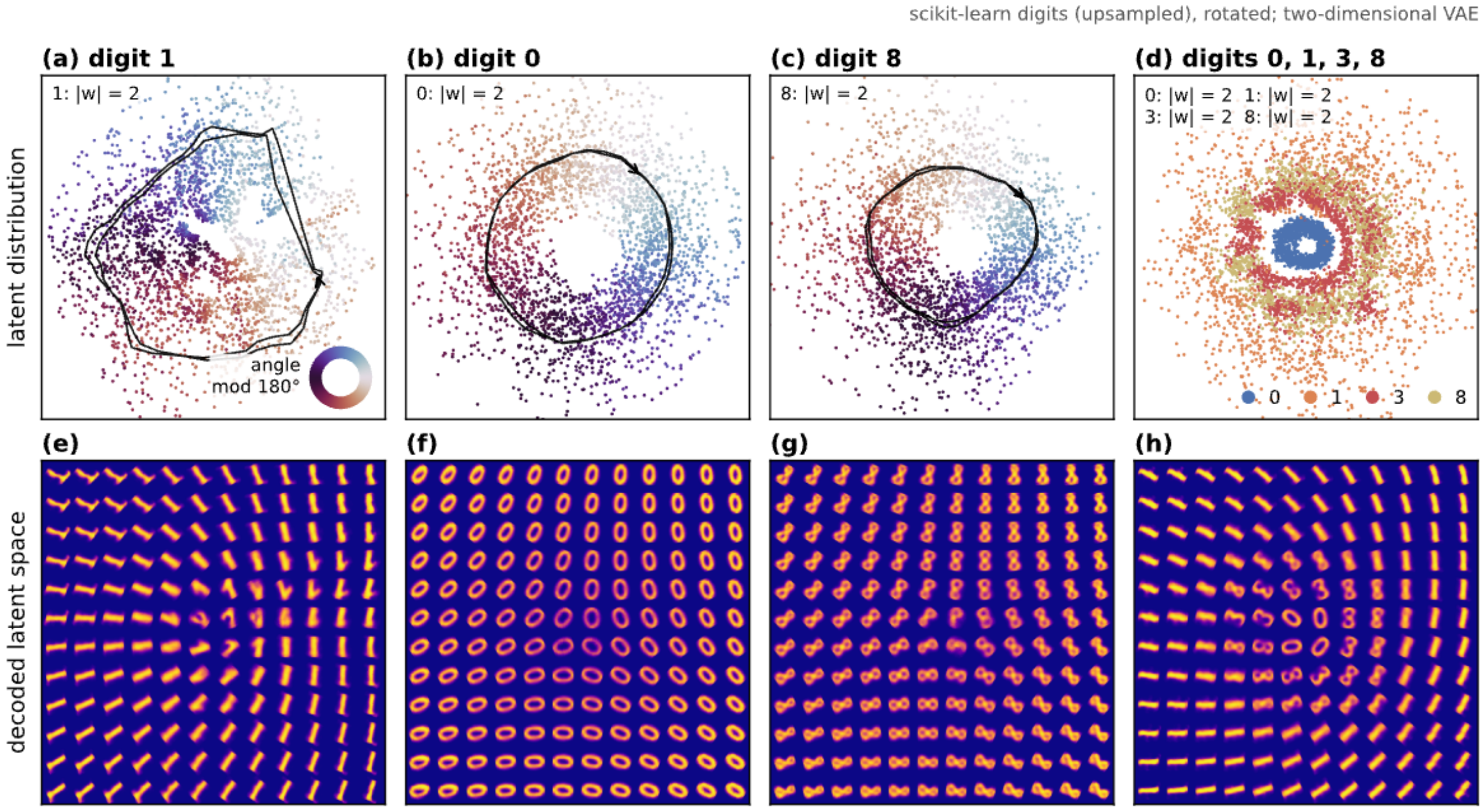

**Figure 6.** Rotated handwritten digits in a two-dimensional VAE. (a–c) Encoded means of rotated images of digits 1, 0 and 8, colored by rotation angle modulo 180°; the black curve follows one training-set base image through a full rotation and winds twice about the ring center ($|w| = 2$). (d) Joint representation of digits 0, 1, 3 and 8, colored by class, with the winding number of each class. (e–h) Images decoded on regular latent grids for the models in (a–d). The ring centers decode to rotational averages. Training and winding measurements are described in Section S7.3.

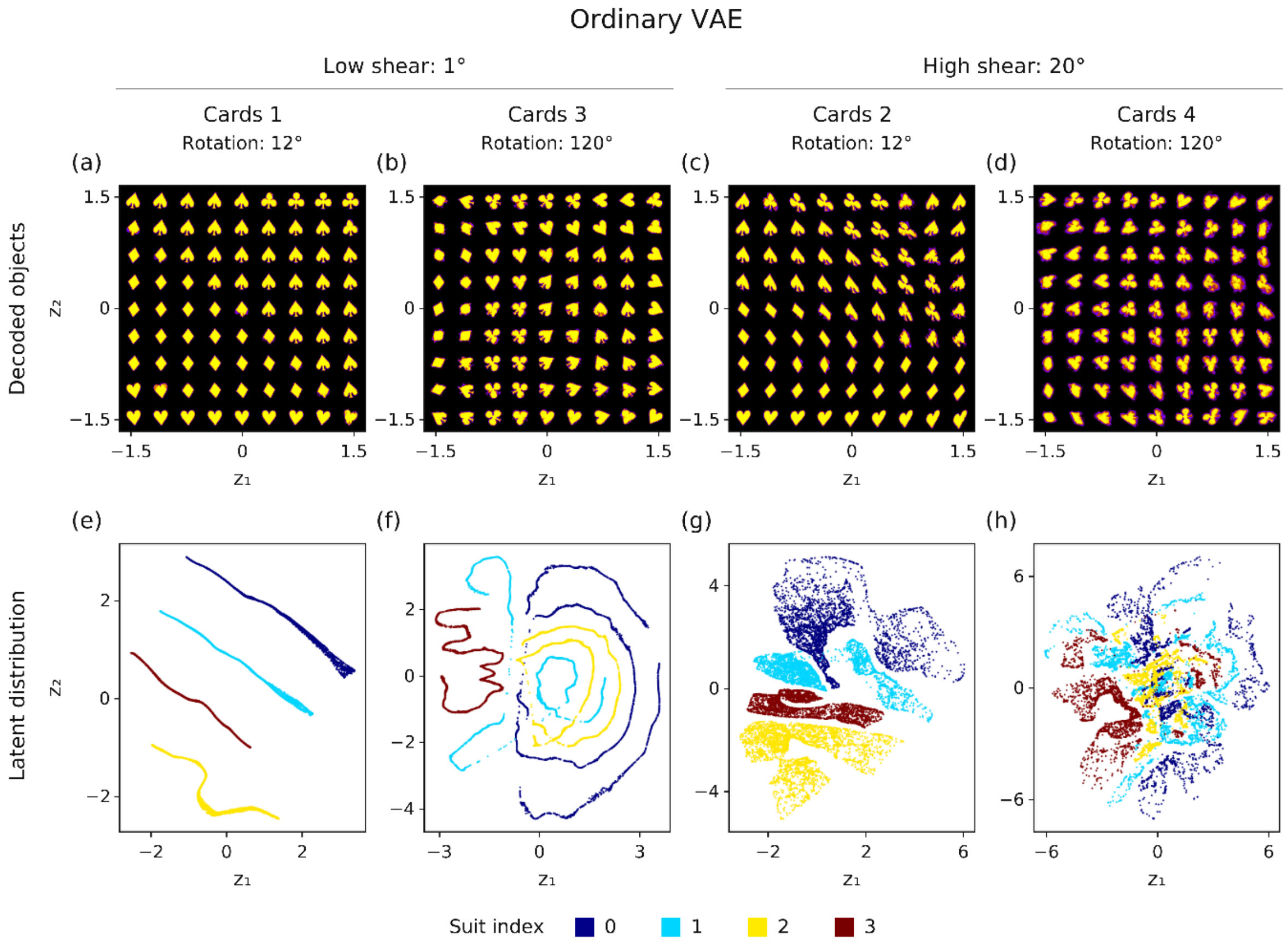

**Figure 7.** Rotation and shear in the latent representation of an ordinary VAE trained on Cards. Columns show Cards 1, 3, 2 and 4, with rotation/shear settings of 12°/1°, 120°/1°, 12°/20° and 120°/20°, respectively. Increasing rotation can therefore be compared at fixed shear within each pair of columns. (a–d) Objects decoded on regular grids of latent coordinates; these are decoder outputs, not examples drawn from the training dataset. (e–h) Corresponding distributions of encoded observations, colored by suit. Larger rotation produces curved within-class trajectories, while larger shear broadens the distributions and changes the decoded shapes. The grids and scatter plots retain their original coordinate ranges, which differ between panels. Latent axes belong to separately trained models and have not been aligned across conditions.

The Cards examples isolate rotation and shear as two sources of within-class variation (Figure 7) and illustrate folding. At low shear, each suit occupies an extended trajectory, and increasing the rotation setting from 12° to 120° makes the trajectories strongly curved. The decoded grids show that moving through these regions changes the orientation and, at class boundaries, the motif itself. Increasing shear broadens the distributions and produces stronger shape changes and mixed decodings. Much of the complexity of the latent map therefore describes transformations of the same four motifs rather than additional structural classes. The corresponding rotationally invariant VAE separates orientation into an angular variable and gives more compact suit-specific content distributions at low shear. Residual shear still produces extended content distributions. That comparison and its angular diagnostics are described in Section S3.3.

The winding arguments in this section assume that the orbit is encoded in a plane, as in the two-dimensional examples of Figures 6 and 7. In a plane, a closed curve with winding number $w$ about a point has at least $w - 1$ self-intersections, and its winding changes only when it passes through the core. With a third active coordinate neither constraint holds: the strands of a partially resolved quotient orbit can separate along that coordinate instead of crossing, and a loop can be unwound around the core rather than through it. Winding sectors are then separated by the cost of activating the additional coordinate rather than by topology. The quotient periodicity and the cut of a real-valued angle head do not depend on the latent dimension, because they follow from the symmetry of the data and the periodicity of the angle. These higher-dimensional cases have not been tested with trained networks.

### 4.4 Folding and its characteristic length scale

When the data contain more independent factors than the latent space has coordinates, the weaker factor may be discarded or combined with a stronger one (Figure 7f). A folded representation provides one way to retain it: one factor varies along each strand, while successive strands encode different levels of the additional factor. For two uniform factors, let $A_c = \frac{{w_c}^2}{2\beta {\sigma_x}^2}$ denote the dimensionless strength of the added factor, using the variance and likelihood convention of equations (19a) and (19b). After optimizing the posterior width in the narrow-posterior limit and neglecting the smaller boundary correction, the reduced loss gives

$$L_N \simeq \frac{A_c}{N^2} + \ln N + C \tag{23a}$$

$$N^* \simeq \sqrt{2A_c} \tag{23b}$$

$$\ell_{fold} \simeq \frac{L_z}{N^*} \tag{23c}$$

Here $C$ is independent of the strand count, $L_z$ is the latent span, and $\frac{L_z}{N}$ is the length available to each strand. More strands reduce the residual error in the additional factor, but their information cost increases as ln N. Equations (23b) and (23c) show that the preferred strand count grows as the square root of the factor strength, while the strand spacing decreases inversely with that count. This spacing is different from the wavelength of the first small modulation of an unfolded encoding: the stability calculation in Section S3.5 selects $k = \frac{1}{s}$, or a wavelength $2\pi s$, where $s$ is the posterior width. The modulation scale and the spacing of a fully developed fold should therefore not be treated as the same quantity. Direct optimization of the idealized model gives preferred strand counts of 3, 4 and 6 at added-factor strengths of 4, 8 and 16. For the bounded uniform factors considered here, a modulated state first appears near strength 0.7, while an ordered serpentine becomes favorable near one. The initial modulation is concentrated near the ends of the data range and keeps approximately the same wrinkle count as its amplitude increases. This sequence resembles the transition from wrinkles to folds in confined systems [74–76], although the strand law here follows from reconstruction and information costs rather than an elastic energy. The branch comparisons, stability calculation and numerical examples are given in Section S3.5.

The existence of a folded solution does not mean that ordinary network training will find it. In the tested two-dimensional VAEs with three independent synthetic factors, the third factor is usually suppressed or mixed with the first two rather than arranged into ordered strands, although individual folded trajectories can appear, as for one suit in Figure 7(f). The idealized model contains competing local minima, which offers a possible explanation but does not establish the cause of the network behavior. The corresponding trained examples and their comparison with the optimized folds are given in Section S3.5.

The fold law is derived for two factors that share one latent axis, as when three factors are encoded in two dimensions. Where $m$ continuous factors vary and $s$ classes meet, a locally disentangled representation requires $d \geq m + s - 1$ (Section S3.4), so a third coordinate removes the geometric necessity of folding three factors. It does not remove the incentive: for three classes carrying one factor, the folded layout has lower loss than the disentangled prism that becomes available in three dimensions (Figure S2(b)). In higher-dimensional latent spaces, folding is therefore decided by the relative cost of reusing an occupied axis and activating a new one rather than by dimension counting alone.

### 4.5 Training dynamics near collapse

To describe the onset of separation during training, we consider continuous-time gradient descent for the class-separation parameter $m$ of Section 3.1. Above $R_c$, a small displacement from the collapsed state grows as $m(t) \simeq m(0)e^{\lambda_{\text{esc}} t}$, with $\lambda_{\text{esc}} = 2(R - R_c)$. We call this the escape rate: it measures the growth of an infinitesimal separation away from an unstable collapsed state, not a stochastic rate for crossing a barrier. Near the separated minimum, a small deviation instead decays, with relaxation rate approaching $4(R - R_c)$ near the transition. Both rates tend to zero at collapse, so separation develops and equilibrates slowly there. Figure 2(d) shows the numerical rates and their near-transition limits.

KL annealing ramps $\beta$ so that $R$ crosses $R_c$; we model it as $R(t) = R_c + t/\tau_Q$. Starting from an encoding of size $m_0$, the separation grows as $m_0 \exp(t^2/\tau_Q)$, and symmetry breaking is delayed by

$$t_d \approx \sqrt{\frac{\tau_Q}{2} \ln \frac{2t_d}{\tau_Q m_0^2}} \tag{24}$$

The numerical gradient-flow calculation illustrates this delay and agrees with the estimate to within 3%. Near the transition, separation develops slowly because the restoring or destabilizing rate approaches zero [77]. With several classes, the covariance eigenvalues select successive directions of separation rather than requiring every class to separate at once. Section S3.6 summarizes these rates; the time scale here refers to the specified reduced dynamics.

## 5. Changing the learning objective

The preceding results follow from balancing reconstruction against regularization toward a prior. We now ask which features change when the model uses a different generator, learns its prior, omits the decoder or learns its coordinates from a measured property. These comparisons distinguish the geometry needed to reconstruct structural variation from the geometry useful for predicting a particular target.

### 5.1 Generative alternatives: flows, priors and decoder-free embeddings

A continuous generator must connect the outputs assigned to neighboring latent points. If its data consist of two well-separated families, it must therefore pass through intermediate outputs somewhere between their latent regions. How much probability falls in that transition depends on how abruptly the generator can change. A Lipschitz constant bounds that rate of change: a smaller constant permits only gentler transitions. Existing Gaussian-isoperimetric results make this connection quantitative for a Gaussian input distribution and separated target components [51, 52]. Under the assumptions in Section S4.1,

$$L \geq \frac{\Delta - 2\delta}{\Phi^{-1}(p+\eta) - \Phi^{-1}(p)} \approx \frac{(\Delta - 2\delta)\,\varphi\left(\Phi^{-1}(p)\right)}{\eta} \quad (25)$$

Here $L$ limits how rapidly the generator output can change with its input. The target families are separated by $\Delta$, and each is surrounded by an accepted neighborhood of radius $\delta$. The probability outside those neighborhoods is at most $\eta$, while $p$ is the Gaussian probability assigned to one populated input region. The functions $\Phi$ and $\phi$ are the standard-Gaussian cumulative distribution and density. The bound says that producing separated families with very little intermediate output requires a sufficiently sharp generator. It applies where the inverse cumulative probabilities are defined; the full construction is given in Section S4.1.

Figure 8 compares four ways in which a model handles separation. In the VAE reference, panel (a), the finite transition follows directly from overlap between the two Gaussian class posteriors. A flow can sharpen its transition as transport proceeds; changing a prior can instead reduce how often a retained transition is sampled. Predictive learning changes the reason for separation altogether, organizing classes by their target values rather than by reconstruction. These are analytical model comparisons, not a ranking of trained generative architectures.

In the two-prototype flow-matching example, Gaussian noise is transported toward two equally probable endpoints [78]. At intermediate times, the conditional endpoint mean changes smoothly between them. Figure 8(b) shows this transition becoming sharper as the endpoint is approached, and the inset gives the shrinking ambiguous interval. The exact limiting transport concentrates probability at the two prototypes; its endpoint is singular, so it is not a counterexample to a bound that assumes a uniformly finite Lipschitz constant. A finite-step sampler can still leave outputs between the modes. The optimal velocity for contrastive flow matching [79] is given in Section S4.1.

Changing the prior affects the sampling probability rather than removing the decoder transition. In Figure 8(c), the decoder is held fixed while a prescribed two-Gaussian prior becomes more bimodal. All priors have unit total variance, but progressively less probability lies near the wall. These curves are a parameter sweep, not a learned-prior experiment. A learned mixture can use the same mechanism, although the result also depends on how many modes it has available relative to the data classes. Section S4.1 gives the separated limit of a jointly learned two-component prior.

A decoder-free embedding has a different requirement again: it need not generate an image at every point between encoded clusters. Its gaps therefore do not produce decoded mixtures, although the representation may still omit information needed for a scientific task [80]. Figure 8(d) examines this distinction for a restricted DKL class model. Section 5.2 explains why target differences, noise and kernel scale determine the preferred separation in that calculation.

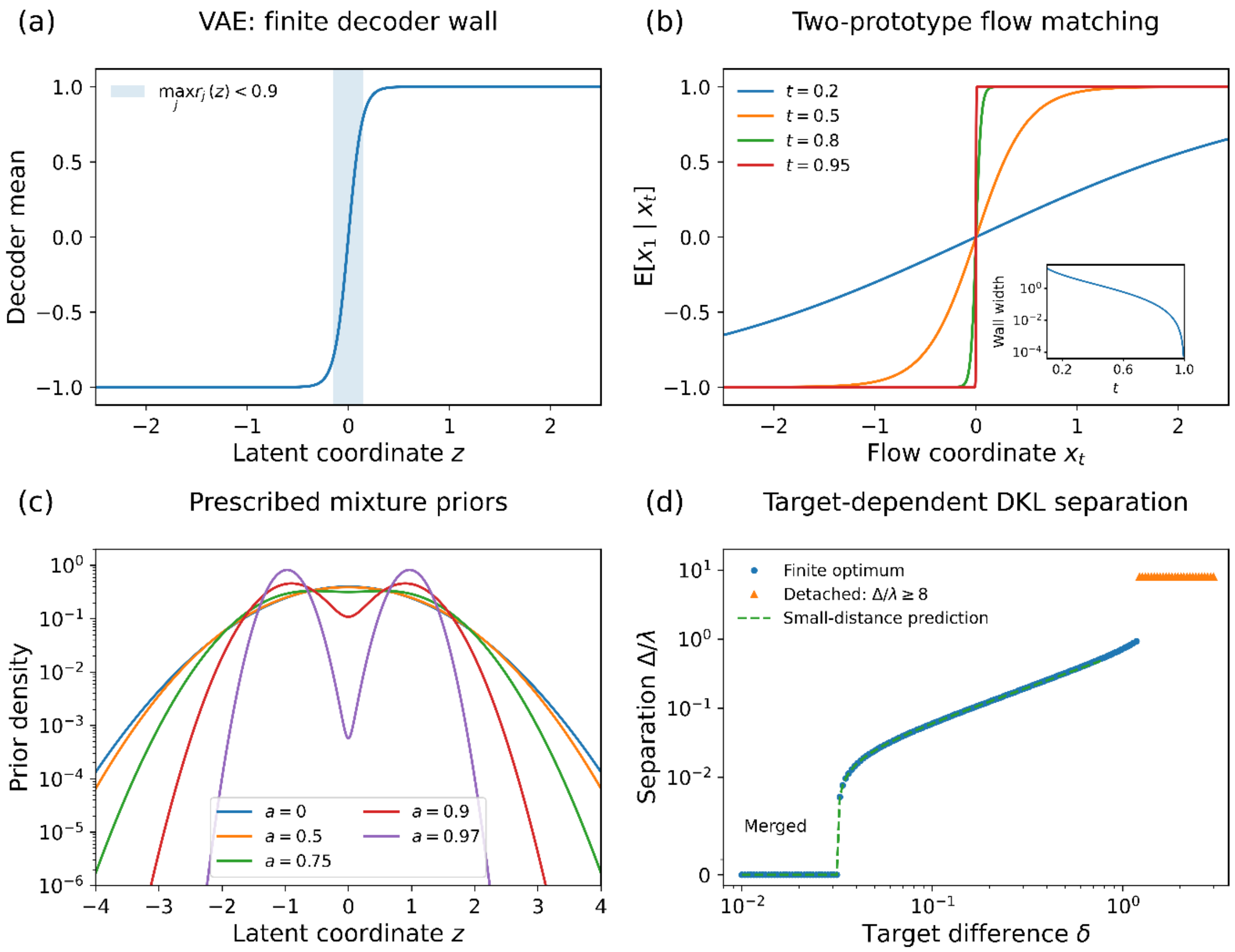

**Figure 8.** Analytical comparison of decoder transitions and target-dependent latent geometry. (a) Conditional-mean decoder for two equally probable prototypes at −1 and +1, with Gaussian class posteriors centered at ±1.5 and standard deviation 0.45. Shading marks maximum class responsibility below 0.9. (b) Conditional endpoint mean for the linear flow interpolant at four times; the inset shows the full ambiguous-band width. (c) Prescribed two-Gaussian priors with means ±a and component variance $1 - a^2$, retaining unit total variance. The decoder is unchanged. (d) Numerical minimization of the fixed-signal two-cluster GP evidence versus target difference. Circles denote finite optima, triangles mark the upper search bound $\Delta/\lambda = 8$, and the dashed curve is the small-separation approximation. A symmetric-logarithmic vertical scale retains actual zero separation for merged classes. The model uses 20 points per class, signal variance 2.5 and noise variance 0.01. A search-bound result is not an identified finite detached distance. Calculation details are given in Sections S4.1 and S4.2.

### 5.2 DKL and hybrid objectives

DKL replaces the reconstructive objective by a predictive one. A neural feature map and Gaussian-process kernel are trained to predict measured targets [34, 35]. The negative log marginal likelihood is

$$\mathcal{L}_{\mathrm{DKL}} = \frac{1}{2}\,\mathrm{y}^{\top}(K_\theta + \sigma_n^2 I)^{-1}\mathrm{y} + \frac{1}{2}\ln\det(K_\theta + \sigma_n^2 I) \tag{26}$$

Here $(K_\theta)_{ab} = \sigma_f^2 \exp(-|z_a - z_b|^2/2\lambda^2)$ is the squared-exponential kernel on learned latent features, $\sigma_f^2$ is the signal variance, $\sigma_n^2$ is the observation-noise variance, and $\lambda$ is the kernel length scale.

Unlike a VAE, DKL does not have to reconstruct an image at every latent coordinate, and it has no standard-Gaussian latent prior imposing the moment constraint T2. In the restricted class model of Figure 8(d), classes with target differences too small to resolve can merge. Resolvable differences favor larger separation in units of the GP kernel length scale, approximately in proportion to the target difference over an intermediate range. For sufficiently different targets, the optimum reaches the numerical search boundary and the classes are effectively decoupled. These results explain a mechanism for target-dependent geometry; they do not establish a globally distance-preserving map for arbitrary trained networks. Section S4.2 gives the derivation and a comparison with a three-class model.
The practical implication is an objective-dependent representation. Variation useful for predicting the trained target can be retained, while unrelated structural variation can collapse. This explains why strong target smoothness is not equivalent to a complete structural description and is consistent with known feature-collapse and marginal-likelihood concerns [36, 37]. It also means that a structural wall need not create a target jump: neighboring structural classes can have equal properties.

A semi-supervised VAE or hybrid VAE–DKL model asks the same latent coordinates to support reconstruction and prediction [29, 38, 39, 81–86]. At weak predictive weight the GP can sometimes adapt its length scale without changing the structural layout, or select among otherwise nearly degenerate arrangements. At stronger weight, target-indistinguishable classes and target-irrelevant factors can merge or shrink, degrading reconstruction. Whether a factor is visible to the GP acting on encoded means or remains decodable from posterior samples is another distinction. Section S4.3 derives the length-scale freedom, the approximate merging threshold and the relevance-dependent collapse of factors. The distinction between a fixed reconstructive representation and one learned for a target was explored in our process-optimization studies with Valleti and co-workers [28, 87]. Figure 9 examines the same question in a controlled MNIST experiment: how does the latent map change when the model is asked to reconstruct a digit, predict a scalar property, or do both? Two targets are used. Ink density is the mean pixel intensity of an image, with bright strokes and a dark background; it measures how much of the image is occupied by the digit and how intense those strokes are, not which digit is present. The second target assigns a different fixed scalar to each digit class. Both targets are standardized using the training-set statistics.

The contrast between Figures 9(a) and 9(c) is useful because the images are unchanged while the task is changed. The VAE must preserve enough information to reconstruct stroke shape, thickness and other visual details. Class-target DKL needs primarily to distinguish the assigned class values and therefore organizes much of the within-class variation into tighter target-specific regions. Ink-trained DKL in Figure 9(b) makes a different simplification: images with similar total intensity can lie close together even when they are different digits. Its nearly one-dimensional property coordinate is consequently useful for ink prediction but incomplete as a description of digit structure. The class-valued target supplies identity information explicitly; its separation is supervised, whereas ink density is a deliberately simple image statistic.

The hybrid model retains substantial digit organization at weak predictive weight, as shown in Figure 9(d). Panel (e) measures the competition directly: the horizontal axis tests how well nearby latent points identify the digit, while the vertical axis measures how similar the ink densities of latent neighbors are. Stronger predictive weighting makes ink density smoother but eventually reduces digit discrimination. Panel (f) shows the accompanying reconstruction cost: at strong weighting, improved ink smoothness comes with larger pixelwise cross-entropy. The zero-weight hybrid controls for the longer hybrid training budget. These five-seed results demonstrate a task-dependent trade-off, not a universal optimum for the predictive weight; metric definitions are given in Section S7.3. These are local diagnostic measures, not total structural information or a mathematical smoothness constant. The sparse variational GP used here tests the qualitative mechanism of the exact class-evidence model, not its numerical thresholds.

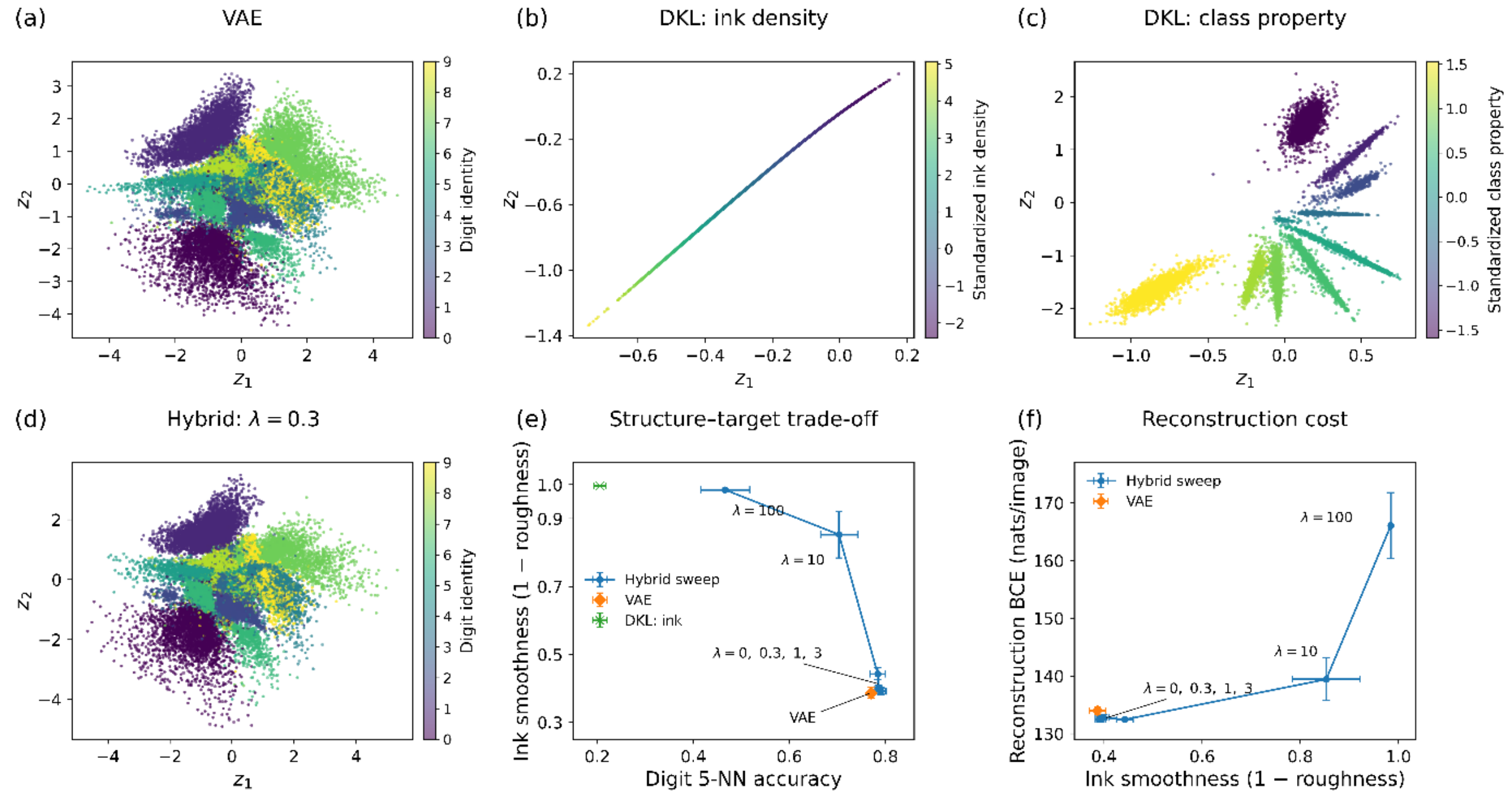

**Figure 9.** Reconstruction and prediction organize the MNIST latent space differently. (a) Ordinary VAE, colored by digit identity. (b) DKL trained to predict ink density—the mean pixel intensity—colored by its standardized value. (c) DKL trained on a fixed scalar assigned to each digit class, colored by that standardized class-valued target. (d) Hybrid VAE–DKL at predictive weight $\lambda = 0.3$, colored by digit identity. These representative maps use seed 17; their coordinate systems have not been aligned or rescaled. (e) Held-out five-nearest-neighbor digit accuracy versus ink smoothness for the hybrid sweep, with VAE and ink-trained DKL references. Smoothness is one minus the ratio of mean absolute ink differences between latent neighbors to those between random pairs. (f) Held-out pixelwise binary cross-entropy at the posterior mean versus ink smoothness; lower values indicate better reconstruction. Points and error bars in (e,f) show the mean and sample standard deviation over five seeds. Hybrid $\lambda = 0$ is a matched-training-budget control. Here $\lambda$ denotes the raw predictive weight, not the GP kernel length scale of equation (26).

## 6. Consequences for generation, interpolation and search

We now consider what the latent structure means when the model is used to generate states, interpolate between observations or search for a desired property. The relevant issue is not simply whether the coordinates are continuous, but how probability is distributed among domains, what the decoder produces between them and which distinctions matter to the objective. Rare states and folded factors introduce additional constraints on coverage and resolution.

### 6.1 Probability allocation and interpolation

If a small piece of the boundary is approximately straight, we can treat it locally as a planar wall. Such a wall with log-odds slope $2/\ell$ contributes a band of width $L_\varepsilon \ell$. Integrating over the partition gives the prior mass in walls,

$$P_{\text{wall}} \approx L_\varepsilon \, \ell \, \text{Per}_\gamma \tag{27}$$

where $\text{Per}_\gamma$ measures the total boundary size weighted by the Gaussian prior. The main implication is that a thin wall can still receive appreciable generation probability. Its prior mass decreases only slowly as reconstruction is weighted more strongly. Section S5.1 gives the quantitative comparisons and the geometric factors for the different layouts.

The encoded data and the prior can place very different probabilities in the same wall. In the separated two-class model, the prior samples the boundary much more often than the encoded observations do, and that mismatch increases with separation. Gaussian tails still provide some encoded probability there; the region is sparsely sampled, not completely absent from training. Its conditional-mean decoder nevertheless blends the neighboring prototypes.

The same wall geometry also constrains interpolation. A path that connects two encoded states can remain smooth inside a domain yet must cross a boundary whenever its endpoints belong to different domains. A straight path between encodings of classes $i$ and $j$ crosses at least one wall and spends a fraction $L_\varepsilon \ell_{ij}/|\mu_i - \mu_j|$ of its length in it, 5–15% per wall for $10 \leq R \leq 10^4$. In a one-dimensional chain of $K$ classes a random pair is separated by $\frac{K+1}{3}$ walls on average. Geodesic interpolation in the pullback metric avoids walls where a detour exists [54] but cannot avoid the wall that separates the two endpoint classes.

A continuous interpolation can therefore cross a class-ambiguous region without being discontinuous as an image path. Whether that path is a useful physical trajectory requires independent validation. Some intermediate outputs may be unrepresentative averages; others may resemble actual intermediate states. Rare-state adsorption in Section 3.3 is a concrete reason not to equate crossing a wall with leaving all scientifically interesting configurations.

**6.2 When latent geometry complicates a predictive surrogate**

Latent-space optimization seeks objects through continuous coordinates, an approach used in molecular, crystal, microstructure, process and biological design [27–29, 32, 33, 38, 87–91]. Syntax-aware or graph-aware decoders reduce some invalid outputs but do not establish empirical support or property calibration everywhere in the latent prior. In microscopy, a boundary seed can similarly be structurally ambiguous even when its coordinates look ordinary under the prior. Section S8.2 lists further applications.

Two observation distributions must be separated. Aggregate posterior density describes coverage of the data used to train the representation. GP uncertainty depends on locations with measured objective values, the kernel and noise model. High representation density can coexist with high predictive uncertainty when the property has not been measured there. Low representation density can signal extrapolation limits of the decoder, but it does not by itself diagnose the GP's training coverage. Mapping both is more informative than substituting one for the other.

A structural wall can create a short predictive length scale when the objective differs sharply across the neighboring states or when uncertain decodings receive penalties. A stationary GP may then smooth across a transition or shorten its fitted length scale. Section S5.2 gives a conditional resolution-count heuristic for this case, not an information-gain bound. It is not a universal sample-complexity penalty for every wall: equal-property classes need not create a target discontinuity, and a nonstationary or domain-aware surrogate can represent transitions differently. Similarly, observing some wall proposals does not establish acquisition-driven attraction without comparing their frequency with the candidate distribution's wall mass.

For a property predictor trained only on encoded observations, a path through a low-density blend can also separate predictor interpolation from decoder behavior. Small steps and local trust regions can reduce uncontrolled changes but do not prove that a path between structural families is meaningful. Domain-aware search and explicit cross-domain proposals are therefore candidate responses to a diagnosed failure, not replacements for validating the objective and the decoder.

**6.3 Rare-state discovery, reweighting and dimensionality**

A rare state can be reconstructed accurately yet occupy only a small fraction of the prior, so random generation or a coarse latent grid may miss it. Giving that class more weight in the entire per-example training objective can enlarge its region in the packing model, but other classes then occupy less space. The asymptotic calculation is given in Section S5.3; its scaling has not been verified in the trained image networks. For discovery, the practical check is therefore how often the states of interest are generated and retained, rather than the average reconstruction error alone.

An intermediate rare state requires a different strategy. It can lie within the wall between two common classes without ever becoming the most probable class at any latent point. Excluding every ambiguous point would then remove a state that the experiment is trying to discover. A restriction useful for optimizing within known families may therefore be unsuitable for searching for new or intermediate states. Before applying it, one should check which configurations it discards. A graded penalty or deliberate exploration of transition regions may be more appropriate than a hard exclusion.

Folding introduces a further trade-off between a compact representation and the resolution of the encoded factors. A smooth objective in the original variables can have separate local optima on different strands, while quantization of the folded factor limits the best value the representation can reach. Section S5.4 gives the corresponding scaling estimates. A two-dimensional space chosen for visualization should therefore not automatically be reused for optimization; it must resolve the factors needed for the task.

The practical choice of representation should follow the scientific objective. For structural analysis, inspect the encoded-data distribution, decoded states, rare classes and known symmetries. For prediction, test the chosen property on held-out observations. For a hybrid model, assess prediction and reconstruction together and check which structural distinctions remain. A prior threshold, a data-density threshold and a class-confidence threshold address different problems and each should be evaluated against the configurations that matter for the intended search.

**7. Summary**

The examples considered here show several ways in which a low-dimensional latent map can acquire structure that is not a direct image of the underlying physics. The common starting point is the need to preserve distinctions in the data while limiting the information carried by the latent variables. How the model meets these competing requirements depends on class frequency, similarity, symmetry, dimensionality and the training objective.

For discrete classes, posterior overlap creates finite-width walls between domains, while the KL term limits the overall extent of the representation. The Flory–Huggins correspondence connects the loss of class separation to mixing and explains why unequal frequencies allow coexistence and metastability. Rare states need not occupy proportionate volumes: they can form compressed islands, remain mixed with common states or become incorporated into the wall between them. The reconstruction metric determines which mixtures resemble a rare state and can therefore change its representation.

Continuous factors are governed by different constraints. Rotations produce periodic orbits whose allowed windings are constrained by exact symmetry, while the realized winding depends on the learned map. A faithful scalar angle representation requires a cut. The posterior covariance can favor particular orientations and arrangements without assigning a separate axis to each physical factor. When several factors must share a coordinate, the idealized objective can favor folding, with a spacing set by the reconstruction–information balance. The trained networks show that suppression or mixing can occur instead. The supplementary calculations compare the optimized model with trained networks.

Changing the learning objective changes what the latent coordinates preserve. A VAE must retain enough information to reconstruct the input; DKL retains distinctions useful for a measured property; a hybrid model balances the two. None of these representations is preferable for every purpose. Structural analysis requires coverage of relevant configurations, whereas prediction and optimization require an adequate description of the chosen property. The choice should follow the scientific question rather than the visual smoothness of the latent map.

## 8. Conclusions

Years of applying VAEs and invariant latent-variable models to microscopy and other physical data have revealed recurring features of low-dimensional latent spaces. Distinct structural families occupy domains separated by narrow transition regions, rare configurations can be confined to small islands, and increasing regularization can remove useful distinctions through posterior collapse. Rotational factors produce rings, cores or cuts depending on how symmetry is represented. When the available coordinates are insufficient, additional variation must be suppressed or combined with other factors, and folded representations become possible. These structures are particularly clear in two dimensions, where the full latent map can be inspected, but the underlying questions of overlap and data coverage also apply in higher dimensions.

In this work we analyzed these observations using a simplified geometric model and numerical examples. For Gaussian class distributions, the optimal decoder and the stationary moment constraint reduce the VAE objective to a packing problem. Its mixing term has the Flory–Huggins regular-solution form, providing a connection between latent domains, wall broadening, collapse and the behavior of rare classes. Extending the analysis to continuous factors and rotations identifies the roles of posterior covariance, symmetry and latent dimensionality, including the characteristic scales of folded representations. Comparisons with flow models, learned priors, DKL and hybrid VAE–DKL models show how the resulting geometry depends on the objective. The trained examples illustrate several of these mechanisms while also showing that neural networks need not reach the idealized lowest-loss configurations.

The implications concern how latent coordinates are used in scientific work. Good reconstruction does not ensure proportional representation of rare events, and a smooth interpolation does not establish a physical transformation pathway. Similarly, a representation suited to predicting one property can discard structural information needed for another question. For microscopy, spectroscopy, molecular design and process optimization, the latent map should therefore be examined together with the encoded-data distribution, decoded states, known symmetries and the variables relevant to the task. Understanding which features arise from the data and which are imposed by the representation is necessary both for interpreting experiments and for using generative models to guide discovery.

## Data and code availability

The notebooks for this paper are available at https://github.com/SergeiVKalinin/Paper_Notebooks/tree/main/VAE_Walls . The card data analysis is based on the notebooks in https://github.com/saimani5/VAE-tutorials

**Acknowledgement:**

This work was supported by the National Science Foundation under Award No. 2607469, ATHENA (Advanced Testbed for High-throughput Experimentation in Nano- and Atomic Science), through the Programmable Cloud Laboratories program.

**Contributions and AI use**

Sergei V. Kalinin: conceptualization, methodology, interpretation of results, and writing—original draft, review and editing. AI assistants were used during the development and revision of the work. AI assistance included discussion and checking of analytical arguments, preparation and debugging of numerical code, figure assembly from computational outputs, and drafting and language editing. AI outputs were treated as working material rather than as independent validation. Scientific interpretation, selection of results and responsibility for the manuscript remain with the author.

## References

[1] Kingma D P and Welling M 2014 Auto-encoding variational Bayes *Proc. 2nd Int. Conf. on Learning Representations (ICLR)* (arXiv:1312.6114)

[2] Rezende D J, Mohamed S and Wierstra D 2014 Stochastic backpropagation and approximate inference in deep generative models *Proc. 31st Int. Conf. on Machine Learning (PMLR vol 32)* pp 1278–86

[3] Kingma D P and Welling M 2019 An introduction to variational autoencoders *Found. Trends Mach. Learn.* **12** 307–92

[4] Higgins I, Matthey L, Pal A, Burgess C, Glorot X, Botvinick M, Mohamed S and Lerchner A 2017 β-VAE: learning basic visual concepts with a constrained variational framework *Proc. 5th Int. Conf. on Learning Representations (ICLR)*

[5] Burgess C P, Higgins I, Pal A, Matthey L, Watters N, Desjardins G and Lerchner A 2018 Understanding disentangling in β-VAE arXiv:1804.03599

[6] Locatello F, Bauer S, Lucic M, Rätsch G, Gelly S, Schölkopf B and Bachem O 2019 Challenging common assumptions in the unsupervised learning of disentangled representations *Proc. 36th Int. Conf. on Machine Learning (PMLR vol 97)* pp 4114–24

[7] Bepler T, Zhong E D, Kelley K, Brignole E and Berger B 2019 Explicitly disentangling image content from translation and rotation with spatial-VAE *Advances in Neural Information Processing Systems* **32**

[8] Khemakhem I, Kingma D P, Monti R P and Hyvärinen A 2020 Variational autoencoders and nonlinear ICA: a unifying framework *Proc. 23rd Int. Conf. on Artificial Intelligence and Statistics (PMLR vol 108)* pp 2207–17

[9] Kingma D P, Mohamed S, Rezende D J and Welling M 2014 Semi-supervised learning with deep generative models *Advances in Neural Information Processing Systems* **27**

[10] Sohn K, Lee H and Yan X 2015 Learning structured output representation using deep conditional generative models *Advances in Neural Information Processing Systems* **28**

[11] Wetzel S J 2017 Unsupervised learning of phase transitions: from principal component analysis to variational autoencoders *Phys. Rev. E* **96** 022140

[12] Baul A, Walker N, Moreno J and Tam K-M 2023 Application of the variational autoencoder to detect the critical points of the anisotropic Ising model *Phys. Rev. E* **107** 045301

[13] Luchnikov I A, Ryzhov A, Stas P-J, Filippov S N and Ouerdane H 2019 Variational autoencoder reconstruction of complex many-body physics *Entropy* **21** 1091

[14] Park S M *et al* 2022 Optimization of physical quantities in the autoencoder latent space *Sci. Rep.* **12** 9003

[15] Yamada K *et al* 2024 Melting phenomena of self-organized magnetic structures investigated by variational autoencoder *Comput. Phys. Commun.* **305** 109329

[16] Stein H S, Guevarra D, Newhouse P F, Soedarmadji E and Gregoire J M 2019 Machine learning of optical properties of materials – predicting spectra from images and images from spectra *Chem. Sci.* **10** 47–55

[17] Kalinin S V, Dyck O, Jesse S and Ziatdinov M 2021 Exploring order parameters and dynamic processes in disordered systems via variational autoencoders *Sci. Adv.* **7** eabd5084

[18] Kalinin S V, Steffes J J, Liu Y, Huey B D and Ziatdinov M 2022 Disentangling ferroelectric domain wall geometries and pathways in dynamic piezoresponse force microscopy via unsupervised machine learning *Nanotechnology* **33** 055707

[19] Kalinin S V, Zhang S, Valleti S M P, Pyles H, Baker D, De Yoreo J J and Ziatdinov M 2021 Disentangling rotational dynamics and ordering transitions in a system of self-organizing protein nanorods via rotationally invariant latent representations *ACS Nano* **15** 6471–80

[20] Oxley M P *et al* 2021 Probing atomic-scale symmetry breaking by rotationally invariant machine learning of multidimensional electron scattering *npj Comput. Mater.* **7** 65
[21] Liu Y *et al* 2021 Decoding the shift-invariant data: applications for band-excitation scanning probe microscopy *Mach. Learn.: Sci. Technol.* **2** 045028
[22] Valleti S M P *et al* 2022 Unsupervised learning of ferroic variants from atomically resolved STEM images *AIP Adv.* **12** 105122
[23] Biswas A, Ziatdinov M and Kalinin S V 2023 Combining variational autoencoders and physical bias for improved microscopy data analysis *Mach. Learn.: Sci. Technol.* **4** 045004
[24] Valleti S M P, Ziatdinov M, Liu Y and Kalinin S V 2024 Physics and chemistry from parsimonious representations: image analysis via invariant variational autoencoders *npj Comput. Mater.* **10** 183
[25] Yaman M Y, Kalinin S V, Guye K N, Ginger D S and Ziatdinov M 2023 Learning and predicting photonic responses of plasmonic nanoparticle assemblies via dual variational autoencoders *Small* **19** 2205893
[26] Lu S and Jayaraman A 2023 Pair-variational autoencoders for linking and cross-reconstruction of characterization data from complementary structural characterization techniques *JACS Au* **3** 2510–21
[27] Kim Y, Park H K, Jung J, Asghari-Rad P, Lee S, Kim J Y, Jung H G and Kim H S 2021 Exploration of optimal microstructure and mechanical properties in continuous microstructure space using a variational autoencoder *Mater. Des.* **202** 109544
[28] Valleti M, Vasudevan R K, Ziatdinov M A and Kalinin S V 2022 Bayesian optimization in continuous spaces via virtual process embeddings *Digit. Discov.* **1** 910
[29] Gómez-Bombarelli R *et al* 2018 Automatic chemical design using a data-driven continuous representation of molecules *ACS Cent. Sci.* **4** 268–76
[30] Xie T, Fu X, Ganea O-E, Barzilay R and Jaakkola T 2022 Crystal diffusion variational autoencoder for periodic material generation *Proc. 10th Int. Conf. on Learning Representations (ICLR)*
[31] Lyngby P and Thygesen K S 2022 Data-driven discovery of 2D materials by deep generative models *npj Comput. Mater.* **8** 232
[32] Griffiths R-R and Hernández-Lobato J M 2020 Constrained Bayesian optimization for automatic chemical design using variational autoencoders *Chem. Sci.* **11** 577
[33] Notin P, Hernández-Lobato J M and Gal Y 2021 Improving black-box optimization in VAE latent space using decoder uncertainty *Advances in Neural Information Processing Systems* **34**
[34] Wilson A G, Hu Z, Salakhutdinov R and Xing E P 2016 Deep kernel learning *Proc. 19th Int. Conf. on Artificial Intelligence and Statistics (PMLR vol 51)*
[35] Calandra R, Peters J, Rasmussen C E and Deisenroth M P 2016 Manifold Gaussian processes for regression *Proc. Int. Joint Conf. on Neural Networks (IJCNN)* (arXiv:1402.5876)
[36] Ober S W, Rasmussen C E and van der Wilk M 2021 The promises and pitfalls of deep kernel learning *Proc. 37th Conf. on Uncertainty in Artificial Intelligence (PMLR vol 161)*
[37] van Amersfoort J, Smith L, Jesson A, Key O and Gal Y 2021 On feature collapse and deep kernel learning for single forward pass uncertainty arXiv:2102.11409
[38] Maus N, Jones H T, Moore J S, Kusner M J, Bradshaw J and Gardner J R 2022 Local latent space Bayesian optimization over structured inputs *Advances in Neural Information Processing Systems* **35**
[39] Slautin B N, Pratiush U, Lupascu D C, Ziatdinov M A and Kalinin S V 2026 Integrating predictive and generative capabilities by latent space design via the DKL-VAE model npj Comput. Mater. 12 231

[40] Hoffman M D and Johnson M J 2016 ELBO surgery: yet another way to carve up the variational evidence lower bound *NIPS Workshop on Advances in Approximate Bayesian Inference*
[41] Rosca M, Lakshminarayanan B and Mohamed S 2018 Distribution matching in variational inference arXiv:1802.06847
[42] Dai B and Wipf D 2019 Diagnosing and enhancing VAE models *Proc. 7th Int. Conf. on Learning Representations (ICLR)*
[43] Tomczak J M and Welling M 2018 VAE with a VampPrior *Proc. 21st Int. Conf. on Artificial Intelligence and Statistics (PMLR vol 84)* pp 1214–23
[44] Aneja J, Schwing A, Kautz J and Vahdat A 2021 A contrastive learning approach for training variational autoencoder priors *Advances in Neural Information Processing Systems* **34**
[45] Lucas J, Tucker G, Grosse R and Norouzi M 2019 Don't blame the ELBO! A linear VAE perspective on posterior collapse *Advances in Neural Information Processing Systems* **32**
[46] Ichikawa Y and Hukushima K 2024 Learning dynamics in linear VAE: posterior collapse threshold, superfluous latent space pitfalls, and speedup with KL annealing Proc. 27th Int. Conf. on Artificial Intelligence and Statistics (PMLR vol 238) pp 1936–44
[47] Li Z, Zhang F, Zhang Z and Chen Y 2026 Posterior collapse as a phase transition in variational autoencoders Physica A 683 131228
[48] Alemi A, Poole B, Fischer I, Dillon J, Saurous R A and Murphy K 2018 Fixing a broken ELBO *Proc. 35th Int. Conf. on Machine Learning (PMLR vol 80)* pp 159–68
[49] Khayatkhoei M, Singh M K and Elgammal A 2018 Disconnected manifold learning for generative adversarial networks *Advances in Neural Information Processing Systems* **31**
[50] Tanielian U, Issenhuth T, Dohmatob E and Mary J 2020 Learning disconnected manifolds: a no GAN's land *Proc. 37th Int. Conf. on Machine Learning (PMLR vol 119)*
[51] Salmona A, De Bortoli V, Delon J and Desolneux A 2022 Can push-forward generative models fit multimodal distributions? *Advances in Neural Information Processing Systems* **35**
[52] Issenhuth T, Tanielian U, Mary J and Picard D 2023 Unveiling the latent space geometry of push-forward generative models *Proc. 40th Int. Conf. on Machine Learning (PMLR vol 202)*
[53] Aithal S K, Maini P, Lipton Z C and Kolter J Z 2024 Understanding hallucinations in diffusion models through mode interpolation *Advances in Neural Information Processing Systems* **37**
[54] Arvanitidis G, Hansen L K and Hauberg S 2018 Latent space oddity: on the curvature of deep generative models *Proc. 6th Int. Conf. on Learning Representations (ICLR)*
[55] Shao H, Kumar A and Fletcher P T 2018 The Riemannian geometry of deep generative models *Proc. IEEE Conf. on Computer Vision and Pattern Recognition Workshops* pp 315–23
[56] Chen N, Klushyn A, Kurle R, Jiang X, Bayer J and van der Smagt P 2018 Metrics for deep generative models *Proc. 21st Int. Conf. on Artificial Intelligence and Statistics (PMLR vol 84)* pp 1540–50
[57] Berthelot D, Raffel C, Roy A and Goodfellow I 2019 Understanding and improving interpolation in autoencoders via an adversarial regularizer *Proc. 7th Int. Conf. on Learning Representations (ICLR)*
[58] Falorsi L, de Haan P, Davidson T R, De Cao N, Weiler M, Forré P and Cohen T S 2018 Explorations in homeomorphic variational auto-encoding arXiv:1807.04689
[59] Ritter H and Schulten K 1988 Convergence properties of Kohonen's topology conserving maps: fluctuations, stability, and dimension selection *Biol. Cybern.* **60** 59–71

[60] Durbin R and Mitchison G 1990 A dimension reduction framework for understanding cortical maps *Nature* **343** 644–7
[61] Keil W and Wolf F 2011 Coverage, continuity and visual cortical architecture *Neural Syst. Circuits* **1** 17
[62] Floor P A and Ramstad T A 2009 Shannon–Kotel'nikov mappings for analog point-to-point communications arXiv:0904.1538
[63] Flory P J 1942 Thermodynamics of high polymer solutions *J. Chem. Phys.* **10** 51–61
[64] Huggins M L 1941 Solutions of long chain compounds *J. Chem. Phys.* **9** 440
[65] Cahn J W and Hilliard J E 1958 Free energy of a nonuniform system. I. Interfacial free energy *J. Chem. Phys.* **28** 258–67
[66] Rose K, Gurewitz E and Fox G C 1990 Statistical mechanics and phase transitions in clustering *Phys. Rev. Lett.* **65** 945–8
[67] Bedanov V M and Peeters F M 1994 Ordering and phase transitions of charged particles in a classical finite two-dimensional system *Phys. Rev. B* **49** 2667–76
[68] Milman E and Neeman J 2022 The Gaussian double-bubble and multi-bubble conjectures *Ann. Math.* **195** 89–206
[69] de Gennes P G 1985 Wetting: statics and dynamics *Rev. Mod. Phys.* **57** 827–63
[70] Rolínek M, Zietlow D and Martius G 2019 Variational autoencoders pursue PCA directions (by accident) *Proc. IEEE/CVF Conf. on Computer Vision and Pattern Recognition* pp 12406–15
[71] Ziatdinov M and Kalinin S V 2021 Robust feature disentanglement in imaging data via joint invariant variational autoencoders: from cards to atoms arXiv:2104.10180
[72] Batson J, Haaf C G, Kahn Y and Roberts D A 2021 Topological obstructions to autoencoding J. High Energy Phys. 04 280
[73] Esmaeili B, Walters R, Zimmermann H and van de Meent J-W 2023 Topological obstructions and how to avoid them Advances in Neural Information Processing Systems 36
[74] Seul M and Andelman D 1995 Domain shapes and patterns: the phenomenology of modulated phases *Science* **267** 476–83
[75] Cerda E and Mahadevan L 2003 Geometry and physics of wrinkling *Phys. Rev. Lett.* **90** 074302
[76] Pocivavsek L, Dellsy R, Kern A, Johnson S, Lin B, Lee K Y C and Cerda E 2008 Stress and fold localization in thin elastic membranes *Science* **320** 912–6
[77] Zurek W H 1985 Cosmological experiments in superfluid helium? *Nature* **317** 505–8
[78] Lipman Y, Chen R T Q, Ben-Hamu H, Nickel M and Le M 2023 Flow matching for generative modeling *Proc. 11th Int. Conf. on Learning Representations (ICLR)*
[79] Stoica G, Ramanujan V, Fan X, Farhadi A, Krishna R and Hoffman J 2025 Contrastive flow matching *Proc. IEEE/CVF Int. Conf. on Computer Vision (ICCV)*
[80] Wang T and Isola P 2020 Understanding contrastive representation learning through alignment and uniformity on the hypersphere *Proc. 37th Int. Conf. on Machine Learning (PMLR vol 119)* pp 9929–39
[81] Stanton S, Maddox W, Gruver N, Maffettone P, Delaney E, Greenside P and Wilson A G 2022 Accelerating Bayesian optimization for biological sequence design with denoising autoencoders *Proc. 39th Int. Conf. on Machine Learning (PMLR vol 162)*
[82] Grosnit A *et al* 2021 High-dimensional Bayesian optimisation with variational autoencoders and deep metric learning arXiv:2106.03609
[83] Yoo Y, Yun S, Chang H J, Demiris Y and Choi J Y 2017 Variational autoencoded regression: high dimensional regression of visual data on complex manifold *Proc. IEEE Conf. on Computer Vision and Pattern Recognition (CVPR)*
[84] Eismann S, Levy D, Shu R, Bartzsch S and Ermon S 2018 Bayesian optimization and attribute adjustment *Proc. 34th Conf. on Uncertainty in Artificial Intelligence (UAI)*

[85] Jean N, Xie S M and Ermon S 2018 Semi-supervised deep kernel learning: regression with unlabeled data by minimizing predictive variance *Advances in Neural Information Processing Systems* **31**
[86] Tu L, Talbot A, Gallagher N M and Carlson D E 2021 Supervising the decoder of variational autoencoders to improve scientific utility arXiv:2109.04561
[87] Valleti M, Vasudevan R K, Ziatdinov M A and Kalinin S V 2024 Deep kernel methods learn better: from cards to process optimization Mach. Learn.: Sci. Technol. 5 015012
[88] Kusner M J, Paige B and Hernández-Lobato J M 2017 Grammar variational autoencoder *Proc. 34th Int. Conf. on Machine Learning (PMLR vol 70)* pp 1945–54
[89] Tripp A, Daxberger E and Hernández-Lobato J M 2020 Sample-efficient optimization in the latent space of deep generative models via weighted retraining *Advances in Neural Information Processing Systems* **33**
[90] Ren Z *et al* 2020 An invertible crystallographic representation for general inverse design of inorganic crystals with targeted properties arXiv:2005.07609
[91] Castro E, Godavarthi A, Rubinfien J, Givechian K, Bhaskar D and Krishnaswamy S 2022 Guided generative protein design using regularized transformers arXiv:2201.09948

## Supplementary Information

This Supplementary Information presents the analytical derivations and numerical tests supporting the main paper. Model assumptions and validity ranges are stated with the corresponding results; implementation-specific records and extended parameter sweeps are not repeated.

### S1 Model definitions, identities and their domains of validity

### S1.1 Notation and optimal-decoder reduction

The prototype model has classes j = 1, …, K, frequencies $\pi_j^0 > 0$ summing to one, prototype vectors $a_j$, and Gaussian class posteriors $q_j$(z) = N($\mu_j$, $\Sigma_j$). The aggregate is q(z) = $\Sigma_j\ \pi_j^0\ q_j$(z), and the responsibilities are $r_j$(z) = $\pi_j^0\ q_j$(z)/q(z). The prior is p(z) = N(0, I). The observation-noise variance is $\sigma_x^2$ and the original KL weight is β. Main equation (1) uses loss units divided by β; $R_{ij}$ is defined in equation (2), and $\chi_{ij}$ = 4Rij. These conventions must not be replaced by a different reduction of a pixelwise reconstruction loss without rescaling the coupling.

Conditional-mean proof. At fixed z, the conditional squared error is minimized by the responsibility-weighted prototype mean, equations (3a) and (3b). The conditional variance can be expanded as one-half the responsibility-weighted sum of pairwise squared prototype distances, or equivalently a sum over unordered pairs. Integrating this expression against q(z), with the coupling normalization of equation (2), gives equation (4). This argument requires squared-error reconstruction and an unconstrained optimum over decoder outputs; it is not a claim that a finite trained decoder attains that optimum.

A wall between classes $i$ and $j$ is a level set of $L_{ij}(z) = \ln\left[\pi_i^0 q_i(z)/\pi_j^0 q_j(z)\right]$, which is quadratic in $z$ for Gaussian $q_j$. If $\Sigma_i = \Sigma_j$ the wall is a hyperplane. If $\Sigma_i \prec \Sigma_j$, class $i$ wins only inside a closed ellipsoid, an island. Across a planar wall the decoder follows a tanh profile in the normal coordinate with local width $\ell = 2/\left|\nabla L_{ij}\right|$, and inside a domain it is exponentially close to the prototype.

For more than two classes, a pairwise equal-responsibility surface need not be an exposed boundary of the dominant-class partition: a third class can dominate it. The tanh profile is consequently a local two-class result. A bounded responsibility domain also requires the corresponding quadratic inequality to have a nonempty bounded interior; covariance ordering alone does not guarantee that the class wins anywhere.

Operational diagnostics. The responsibility wall is W = {z : max$_j$ $r_j$(z) < 0.9}. Its prior mass integrates p over W, while its encoded-data mass integrates q over W. Decoder-confidence ambiguity is a separate threshold applied to a classifier of decoded outputs. Physical validity is not defined by either threshold. For sampling and search, the relevant measure and the fitted density estimator must always be stated.

### S1.2 Affine stationarity and moment accounting

For $z \to Az + b$ applied to all encodings, the average KL is

$$\frac{1}{2}\left[\mathrm{tr}(ACA^{\top}) + 2b^{\top}A\bar{m} + |b|^2 - d - \sum_j \pi_j^0 \ln\det\left(A\Sigma_j A^{\top}\right)\right] \qquad \text{(S1)}$$

Here C denotes the aggregate second moment before centering and m its mean. Stationarity with respect to translation gives Am + b = 0. After this centering step, the rescaling derivative acts on the centered covariance C − mm$^{\mathrm{T}}$; equivalently, C can denote the already centered second moment in the rescaling step. A fully admissible affine transformation makes the transformed covariance the identity. A diagonal covariance family allows the coordinatewise version.

With a diagonal encoder only diagonal $A$ is available, and the statement holds per dimension: each latent coordinate has aggregate mean 0 and second moment 1. This coordinatewise result applies when the encoder admits the rescaling and the decoder can compensate through an unconstrained affine input layer. Two practical qualifications are important. First, convergence along the symmetry orbit is slow, because the restoring gradient is proportional to $\beta$ times the deviation. Second, a decoder whose resolution is fixed in absolute latent units breaks the symmetry, and the second moment then exceeds 1 (section S3.1). At the optimum, T2 reduces the KL term to the volume expression in main equation (6).

For a trained Gaussian encoder, the aggregate covariance is the covariance of the means plus the average encoder covariance: $\mathrm{Cov}(z) = \mathrm{Cov}_x[\mu_\phi(x)] + \mathbb{E}_x[\Sigma_\phi(x)]$. The same decomposition applies within a class. A fit to the means alone omits the second term. Figure 3 uses that encoded-mean proxy; a variance-inclusive Gaussian control adds the second term. The full second-moment diagnostic is $\mathbb{E}_x[\mu_\phi(x)\mu_\phi(x)^\top + \Sigma_\phi(x)]$, not the covariance of the means alone. Section S2.2 reports the corresponding measurement procedure.

### S1.3 Regular-solution identity and dictionary

For class aggregates with the optimal decoder, equation (4) gives the reconstruction term as $\sum_{i<j} 4\, R_{ij}\, \mathbb{E}_{\bar{q}}[r_i r_j]$, because the variance of a discrete distribution is $\sum_j r_j\, |a_j - f^*|^2 = 1/2 \sum_{i,j} r_i\, r_j |a_i - a_j|^2$. With $q(j,z) = \pi_j^0 q_j(z)$, adding and subtracting $\ln\bar{q}$ splits the class-averaged KL term as $\sum_j \pi_j^0\, KL(q_j \parallel p) = KL(\bar{q} \parallel p) + I(z;j)$, with $I(z;j) = H(\pi^0) + \mathbb{E}_{\bar{q}}[\sum_j r_j \ln r_j]$. This separation follows the ELBO decomposition [40], and hence

$$\mathcal{L} = H(\pi) + KL(\bar{q} \parallel p) + \mathbb{E}_{\bar{q}}[f(r)] \quad \text{(S2a)}$$

$$f(r) = \textstyle\sum_{i<j} \chi_{ij}\, r_i r_j + \sum_j r_j \ln r_j \quad \text{(S2b)}$$

$$\chi_{ij} = 4R_{ij} \quad \text{(S2c)}$$

The local mixing term is the regular-solution free energy, integrated with the nonuniform encoded density. For chains of degrees of polymerization $N_A$ and $N_B$, the expression is $\phi\ln\phi/N_A + (1-\phi)\ln(1-\phi)/N_B + \chi\phi(1-\phi)$ [63, 64]. Here $N_A = N_B = 1$, so the model is the monomeric case with symmetric critical composition $\pi^0 = 1/2$. Class responsibilities play the role of local compositions, reconstruction couplings penalize mixing, and the aggregate-density KL term penalizes disagreement with the prior.

The analogy is exact for the displayed functional decomposition but restricted as a variational equivalence. Responsibilities must be realizable by the encoder family, the aggregate density is variable and penalized, and no independent spatial gradient term has been introduced. A freely chosen classical composition field need not be achievable by the permitted encoder.

The words temperature, composition and interface in the dictionary refer to this mapping, not to physical temperature or matter transported by a neural network.

### S1.4 Realizability, depletion and the square-gradient limit

In a Cahn–Hilliard description the interface width is set by the ratio of a gradient energy to the bulk free energy [65]. In the VAE the composition field is constrained to be a Bayes posterior of Gaussians, and this nonlocal constraint replaces the gradient term. A local expansion of the mean-square change in decoder output for $z = \mu + \varepsilon$, $\varepsilon \sim \mathcal{N}(0, \Sigma)$, gives

$$\mathcal{L}_{\text{noise}} \approx \frac{1}{2\tau^2}\, \mathrm{tr}(\Sigma J^\top J) \quad \text{(S3)}$$

where the noise cost is the mean-square decoder-output change divided by twice the likelihood variance, with decoder Jacobian $J$ and likelihood scale $\tau$; odd moments vanish, and the next terms involve the squared Hessian and third derivatives. For the two-class decoder $f = \bar{a} + 1/2\, \Delta a\, \phi$ this is a square-gradient energy with stiffness tensor proportional to $|\Delta a|^2 \Sigma/(8\tau^2)$, so a wall with normal $n$ would have width and tension proportional to $\kappa(n)^{1/2}$ with $\kappa(n) = n^{\mathsf{T}} \kappa\, n$, and higher-order terms would add a bending energy. The expansion requires the decoder to vary slowly over a posterior width. For two classes the wall width is $\ell = s^2/m$ and $\ell/s = k^{-1/2}$ falls below one as soon as $R \gtrsim 1$, so the expansion fails exactly where walls exist; the cost of a wall comes from posterior tails that cross it, a large-deviation effect. Figure S1(d) contrasts the exact reconstruction cost with the leading square-gradient estimate. The phase-field description is therefore at best a near-collapse limit, and even there it misses the wall cost by a factor of about three. Interpreting (S3) as the full leading correction to reconstruction error additionally requires a negligible reconstruction residual or a locally affine decoder; otherwise the residual–Hessian contraction contributes at the same order in the covariance.

Depletion. Equal-covariance posterior log odds vary linearly along the wall normal, giving a tanh decoder. The aggregate density at the wall center is depleted, $\bar{q}(0)/p(0) \approx 0.78/R$, but Gaussian tails leave nonzero encoded mass. Changing the prior can suppress sampling of the transition without making the decoder discontinuous. The fully separated learned two-mode limit approaches ln2, the class entropy $H(\pi^0)$ in equations (S2a)–(S2c). This is not a lower bound at every coupling: a sufficiently weakly coupled mixed state can have lower loss.

## S2 Domains, collapse, rare classes and quantitative checks

### S2.1 Two-class wall asymptotics and the collapse expansion

**Wall-cost asymptotics.** Under class +, the log-odds $L = 2mz/s^2$ is distributed as $\mathcal{N}(\mu, 2\mu)$ with $\mu = 2k$. Its density obeys $p(l) = p(-l)e^l$, and for large $\mu$ near $l = O(1)$, $p(l) \approx e^{-\mu/4} e^{l/2}/\sqrt{4\pi\mu}$. Then

$$\mathbb{E}[\sigma(L)\sigma(-L)] \approx \frac{e^{-\mu/4}}{\sqrt{4\pi\mu}} \int e^{l/2} \sigma(l)\sigma(-l)\, dl = \frac{\pi}{2} \frac{e^{-\mu/4}}{\sqrt{4\pi\mu}} \tag{S4}$$

so that $g(k) = 4\mathbb{E}[r(1-r)] \approx \sqrt{\pi/2}\, k^{-1/2} e^{-k/2}$. The integral uses $\int e^x \operatorname{sech}^2 x\, dx = 2B(3/2,1/2) = \pi$.

**Landau expansion.** With $s^2 = 1 + \sigma$ and $g \approx 1 - k + k^2$, stationarity in $s$ gives $\sigma \approx -2Rm^2$. Substitution yields $\mathcal{L} \approx R - \left(R - \frac{1}{2}\right) m^2 + (R - R^2) m^4$, with positive quartic coefficient $\frac{1}{4}$ at $R_c = \frac{1}{2}$.

The Gaussian-prototype model uses $R_c = 1/2$. The square-root order-parameter onset and inverse-square-root wall broadening are local statements near this symmetric transition. The large-coupling logarithmic law is a different asymptotic regime. Neither asymptotic should be fitted as a universal law over all regularization strengths of a trained heterogeneous image model.

### S2.2 Numerical wall checks and trained two-class measurements

Figure 3 uses normalized 8 × 8 scikit-learn digits (0 and 1), with a stratified 75%/25% train/test split. The VAE has two Gaussian latent dimensions, 128-unit tanh hidden layers, and Bernoulli reconstruction. Three seeds (17, 18 and 19) were trained at $\beta$ = 2, 4, 8 and 16 for 2,200 Adam updates each. Labels were reserved for diagnostics.

Class-mean separation and intrinsic wall width are measured using Gaussian fits to encoded means; the wall is defined by maximum fitted responsibility below 0.9. A variance-inclusive control adds the mean encoder covariance. Prior wall mass is estimated by standard-normal sampling, and decoder recognizability by an independent logistic classifier. The means and sample standard deviations across seeds are shown in Figure 3(c). These estimators are not interchangeable near collapse.

### S2.3 Rare-class widths and prior-volume bias

**Rare-class widths.** Each class sits $t \approx \sqrt{k^*}$ of its own widths from the wall, so $\Delta = t(s_1 + s_2)$. With $\mu_1 = \pi_2^0 \Delta$ and $\mu_2 = -\pi_1^0 \Delta$, T2 gives $\pi_1^0 \pi_2^0 t^2 (s_1 + s_2)^2 + \sum_j \pi_j^0 s_j^2 = 1$. The KL reduces to $-\sum_j \pi_j^0 \ln s_j$, and the dominant constraint $s_1 + s_2 = \text{const}$ gives $s_j \propto \pi_j^0$. These asymptotics concern a separated rare class that is distinct in the selected reconstruction metric. They do not apply unchanged to the adsorption and layer regimes of Section S2.6. A class can retain internal reconstructable variation while occupying a small prior island, so under-generation and within-class information loss should not be conflated.

### S2.4 Heterogeneous couplings, isotropic packing and junctions

We then allow different pairs of classes to have different reconstruction costs. The problem is no longer only how wide a wall is, but which classes become neighbors and how the latent geometry assigns sharp interfaces to dissimilar pairs and softer interfaces to similar pairs. Each pair of neighbors needs a separation $k_{ij} \approx 2\ln R_{ij}$ in width units. With a shared width the wall width is $\ell_{ij} = s/\sqrt{k_{ij}}$: dissimilar pairs get sharp walls and similar pairs soft ones, in which blends look partly plausible. In one dimension the KL is $\frac{1}{2}\ln(1 + V)$, with $V$ the variance of a chain whose gaps are $2\sqrt{k_{ij}}$. Minimizing $V$ is a seriation problem. In this chain model with equal frequencies, the optimum places the most dissimilar pair at the two ends for three classes and nearly always for more (all but 4 of 300 random coupling sets for four classes and all but 1 for five). Unequal frequencies can override this preference, because a rare class is pushed to the end of the chain even when the most dissimilar pair then becomes adjacent (164 of 300 random frequency and coupling sets for three classes and 73 of 300 for four). In higher dimension the layout is a distance-geometry embedding with contact distances that grow as $\sqrt{\ln\|a_i - a_j\|^2}$, so the latent space compresses data dissimilarity logarithmically. With free widths, a class with two neighbors takes the narrowest posterior.

We next generalize pairwise packing to multiple equivalent classes in more than one latent dimension. Symmetry reduces the equal-class problem to a regular simplex and makes it possible to determine how planar walls meet before addressing the full two-dimensional layout. By symmetry the means sit on a regular simplex. The KL reduces to $\frac{d}{2}\ln(1 + k/d)$ and the reconstruction to a single Gaussian–softmax integral, so the problem is again one-dimensional in $k$. Walls are planar and meet at junctions with the symmetry of the simplex.

The two-dimensional case is especially useful because the global topology is directly visible. We compare compact lattice-like packings with shell configurations and determine when the additional cost associated with junctions is small compared with the sum of the pairwise wall costs. A triple junction lies $2/\sqrt{3}$ times farther from the class centers than the walls do, so its cost is suppressed by $e^{-k/6}$ and the pairwise-wall theory is asymptotically exact. In the hard-wall limit the layout minimizes the second moment of $K$ disks of fixed contact distance, which gives compact patches of the triangular lattice with 120° junctions. At finite $R$ each extra contact costs a wall, and this selects ring-and-core shells.

A symmetric four-class layout can retain a fourfold junction because its boundaries follow the class posteriors rather than moving as free interfaces. Reducing one opposite-class coupling splits that junction into two triple junctions; the numerical continuation is shown in Figure S2(c). The example tests symmetry breaking within the specified model, not the stability rules of a free foam.

The Gaussian multi-bubble comparison concerns the cell-filling limit for at most d + 1 equal classes [68]. The finite-width VAE has additional posterior restrictions. Isotropic and diagonal posterior families can therefore favor different layouts even for the same prototypes.

*S2.4.1 Particle formulation, pair potential and junction corrections*

**Particles in a trap.** With a shared isotropic width $s$ and means $\mu_j = s\,x_j$, the substitution $z = s\,u$ maps every posterior to $N(x_j, I)$. The responsibilities $r_j$ and the measure $\bar{q}\,\mathrm{d}z$ then depend only on the $x_j$, and so does the reconstruction cost $\Phi = 1/2\sum_{ij}$ $\chi_{ij}\int \bar{q}\,r_i r_j\,\mathrm{d}u$, with $\chi_{ij} = 4R_{ij}$. The KL term is $1/2\sum_j \pi_j^0\left(s^2|x_j|^2 + ds^2 - d - d\ln s^2\right)$, and stationarity in $s$ gives

$$s^2\left(1+\frac{V}{d}\right) = 1 \tag{S5a}$$

$$V = \sum_j \pi_j^0\,|x_j|^2 \tag{S5b}$$

the trace of T2, $s^2 + d^{-1}\sum_j \pi_j^0\,|\mu_j|^2 = 1$. Substitution gives (14) for every configuration, not only at the optimum. Translation invariance of $\Phi$ and stationarity of the trap give $\sum_j \pi_j^0\,x_j = 0$: the frequency-weighted center of mass is at the origin, so a rare class sits far out, as a light particle does. Stationarity of $\mathcal{L}(\lambda x)$ at $\lambda = 1$ gives the virial theorem (16).

**Pair potential.** Where only classes $i$ and $j$ have appreciable density, $\bar{q} \approx \pi_i^0 q_i + \pi_j^0 q_j$ and $\Phi$ reduces to $\sum_{i<j} 4\,R_{ij}O_{ij}$. Projecting on the pair axis, with $u$ measured from the midpoint, $\lambda_{ij} = \ln(\pi_i^0/\pi_j^0)$ and $\phi$ the standard normal density,

$$O_{ij} = \int \frac{\pi_i^0\pi_j^0\,q_i q_j}{\pi_i^0 q_i + \pi_j^0 q_j}\,\mathrm{d}u = \sqrt{\pi_i^0\pi_j^0}\,e^{-\rho^2/8}\int \frac{\phi(u)\,\mathrm{d}u}{2\cosh[(\lambda_{ij}-\rho u)/2]} \tag{S6}$$

For $K$ equal classes $O_{ij} = g(\rho^2/4)/2K$. For large $\rho$ the hyperbolic factor confines the integral to a width $1/\rho$ around $u = \lambda_{ij}/\rho$, and $\int \mathrm{d}v/2\cosh(v/2) = \pi$ gives

$$O_{ij} \approx \sqrt{\frac{\pi}{2}}\,\frac{\sqrt{\pi_i^0\pi_j^0}}{\rho}\,\exp\left[-\frac{\rho^2}{8} - \frac{\lambda_{ij}^2}{2\rho^2}\right] \tag{S7}$$

which gives main equation (15). Quantitative calculations use the exact integral (S6), rather than treating its tail approximation as exact. Differentiating the asymptote gives the pair force $F_{ij} = U_{ij}(\rho/4 + 1/\rho - \lambda_{ij}^2/\rho^3)$ and effective inverse-power exponent $n = -\mathrm{d}\ln U/\mathrm{d}\ln\rho = 1 + \rho^2/4 - \lambda_{ij}^2/\rho^2$. At the symmetric two-class optimum, $\rho^2 = 4k^*$, relating the steepness of the repulsion to the optimal separation.

**Contact separations and ordering.** If each contact carries a comparable force $f$, $(\rho_{ij}/4)\,U_{ij}(\rho_{ij}) = f$ gives $\rho_{ij}^2 = 8\ln(R_{ij}/R_0) + O(\ln\ln R_{ij})$ with $R_0$ independent of the pair. For two classes the force balance is the stationarity condition of the two-class loss and gives $k^* = 2\ln R + \ln k^* + \ln(\pi/2)$. In one dimension classes two contacts apart interact with relative weight $e^{-3\rho^2/8} = e^{-3k/2} \approx R^{-3}$, so only neighbors interact.

**Onset.** Near the mixed state all posteriors overlap and the pair approximation fails, but $\Phi$ can be expanded exactly. Let $\tilde{a}_j = (a_j - \sum_i \pi_i^0\,a_i)/(2\beta\sigma_x^2)^{1/2}$ be the centered, scaled prototypes and $\bar{a}(u) = \sum_j r_j\,\tilde{a}_j$. Since $\int \bar{q}\,r_j\,\mathrm{d}u = \pi_j^0$ for every configuration, $\Phi = \Phi_0 - \int \bar{q}\,|\bar{a}(u)|^2\,\mathrm{d}u$. With $\sum_j \pi_j^0\,x_j = 0$, $r_j = \pi_j^0[1 + x_j\cdot u + O(x^2)]$, so

$$\mathcal{L} = \Phi_0 + \frac{1}{2}\sum_j \pi_j^0\,|x_j|^2 - \sum_\alpha \left|\sum_j \pi_j^0\,x_{j\alpha}\tilde{a}_j\right|^2 + O(x^3) \tag{S8}$$

The quadratic form becomes negative when $\lambda_{\max}(C_{\pi^0}) > \beta\sigma_x^2$, with $C_{\pi^0} = \sum_j \pi_j^0 \left(a_j - \bar{a}\right)\left(a_j - \bar{a}\right)^{\mathsf{T}}$ and $\bar{a} = \sum_j \pi_j^0\, a_j$, along $x_{j\alpha} \propto \left(a_j - \bar{a}\right) \cdot v_1$, where $v_1$ is the leading eigenvector of $C_{\pi^0}$. This is the spinodal of the Flory–Huggins correspondence and the collapse threshold of linear VAEs (equations (S9a) and (S9b)). For three classes with frequencies 0.5, 0.3 and 0.2 and non-collinear prototypes, the one-dimensional optimum separates at 1.01 times the predicted prototype scale and remains mixed at 0.99 times.

**Junctions.** Where $n$ equivalent classes have equal density, the exact integrand of $\Phi$ is $\bar{q}\sum_{i<j}\chi_{ij}/n^2$ and the pair sum is $\bar{q}\sum_{i<j}\chi_{ij}/2n$, so pair additivity overcounts the local cost by $n/2$: by 3/2 at a triple point and by 2 at a fourfold point. Junction corrections therefore lower the loss. A triple point of an equilateral cluster lies $\rho/\sqrt{3}$ from the class centers and the fourfold point of a square $\rho/\sqrt{2}$, against $\rho/2$ for the middle of a wall, so the corrections carry relative weights $e^{-\rho^2/24} = e^{-k/6}$ and $e^{-\rho^2/8} = e^{-k/2}$. For three to ten equal classes at $R = 20$ in two dimensions with a shared isotropic width, the pair model and the full objective select the same shells; the pair-model layout lies within $4 \times 10^{-4}$ of the full optimum, and the pair sum exceeds the full loss by 0.001–0.007, the junction correction.

**Frequencies and dissolution.** Unequal frequencies enter twice: the trap stiffness on class $j$ is $\pi_j^0 s^2$, whereas the pair amplitude is $\left(\pi_i^0\pi_j^0\right)^{1/2} e^{-\lambda_{ij}^2/2\rho^2}$. For a rare class $j$ the ratio of repulsion to confinement therefore grows as $\left(\pi_i^0/\pi_j^0\right)^{1/2}$, and the class is pushed into the prior tail. In the pair approximation the potential is bounded, $U_{ij}(0) = 4R_{ij}\pi_i^0\pi_j^0/\left(\pi_i^0 + \pi_j^0\right)$, so classes can overlap at finite cost; overlap is the dissolved state.

**Limits.** Pair additivity fails near collapse, where (S8) applies; at junctions, where the corrections above apply; and for an intermediate class in a wall, which lowers the conditional variance there and is a three-body effect. Per-class widths give each class its own range and curve the walls; diagonal widths make $U_{ij}$ depend on the orientation of $x_i - x_j$ relative to the latent axes; other likelihoods replace $\chi_{ij}$ by the corresponding divergence; and classes with internal factors are extended objects rather than points.

### S2.5 Critical point, coexistence and metastability

**Critical point and spinodal.** For $\sum_j \delta r_j = 0$ the second variation of $f$ at the mixed state $r_j = \pi_j^0$ is $1/2\sum_j \delta r_j^2/\pi_j^0 - \left|\sum_j \delta r_j a_j\right|^2/(2\beta\sigma_x^2)$, because $4R_{ij}$ is proportional to a matrix of squared distances. The mixed state is therefore unstable when

$$\lambda_{\max}(C_{\pi^0}) > \beta\sigma_x^2 \tag{S9a}$$

$$C_{\pi^0} = \sum_j \pi_j^0 \left(a_j - \bar{a}\right)\left(a_j - \bar{a}\right)^{\mathsf{T}} \tag{S9b}$$

which is the collapse threshold of linear VAEs [45]. For two classes with frequencies $\pi^0$ and $1 - \pi^0$ it reads $\chi > 1/[2\pi^0(1 - \pi^0)]$, which gives $R_c = 1/2$ at equal frequencies. Near $R_c$ the transition is mean-field: the fitted order-parameter exponent is 0.504, the condensation gain $\mathcal{L}_{\text{mixed}} - \mathcal{L}_{\text{sep}}$ scales as $(R - R_c)^2$ with fitted exponent 2.007 and prefactor 1.00, and $\ell[2(R - R_c)]^{1/2} = 0.997$ at $R - R_c = 10^{-3}$, so $\nu = 1/2$. Escape from the collapsed state occurs at rate $2(R - R_c)$ and relaxation of the ordered state at rate $4(R - R_c)$, the Landau values. Far from the transition the wall width crosses over to the logarithmic law $\ell \approx 1/(2\ln R)$ (figure S1(c)); in polymer blends the strong-segregation width scales as $\chi^{-1/2}$ [92]. In mean-field Cahn–Hilliard theory the interfacial tension vanishes as $(R - R_c)^{3/2}$ [65]; the VAE has no separate bulk and interface contributions near $R_c$, because the separated state is a single wall, and its condensation gain scales as $(R - R_c)^2$, the Landau value for the whole system.

**Binodal and solubility.** For two classes the separated optimum with free means and widths has lower loss than the mixed state above the binodal of table S1, close to $1.6\ln(1/\pi^0)$ for small $\pi^0$. These values exceed the regular-solution binodal $\ln[(1-\pi^0)/\pi^0]/(1-2\pi^0)$ because the separated state also pays for a wall and for precise posteriors. The mixed state remains locally stable up to the spinodal $1/[2\pi^0(1-\pi^0)]$, so the metastable window grows as $1/\pi^0$ (figure S1(b)). When the frequent classes already carry a continuous factor, the window narrows: at $\pi^0 = 0.03$ the binodal falls to $\chi \approx 4$–5 and the dissolved state loses stability at $\chi \approx 5$–7 instead of 17, because the ordered majority provides nucleation sites. An unused prior-matched latent coordinate contributes nothing to the loss, so an existing solution can be embedded at unchanged loss. This does not exclude a lower-loss solution with the new coordinate active. In these calculations, the additional direction let a dissolved rare class separate without crossing the walls of the majority: three classes with $\pi^0 = 0.02$ and $\chi = 10$ that remained dissolved in a one-dimensional latent separated spontaneously in two dimensions. Internal variation of a rare class survives even in one dimension ($R^2 = 0.97$ for $\pi^0$ between 0.01 and 0.1), because the KL cost is paid per data point and the prior tail is cheap. The prior volume of a distinct rare class falls below its frequency, to 18% of it at $\pi^0 = 0.05$ and $R = 10$ and to 4% at $R = 10^3$.

Table S1. Binodal and spinodal of a class with frequency $\pi^0$.

| $\pi$ | **binodal $\chi_b$ (VAE)** | **binodal, regular solution** | **spinodal $\chi_s$** |
|---|---|---|---|
| 0.30 | 2.28 | 2.12 | 2.4 |
| 0.10 | 3.97 | 2.75 | 5.6 |
| 0.03 | 5.94 | 3.70 | 17.2 |
| 0.01 | 7.46 | 4.69 | 50.5 |

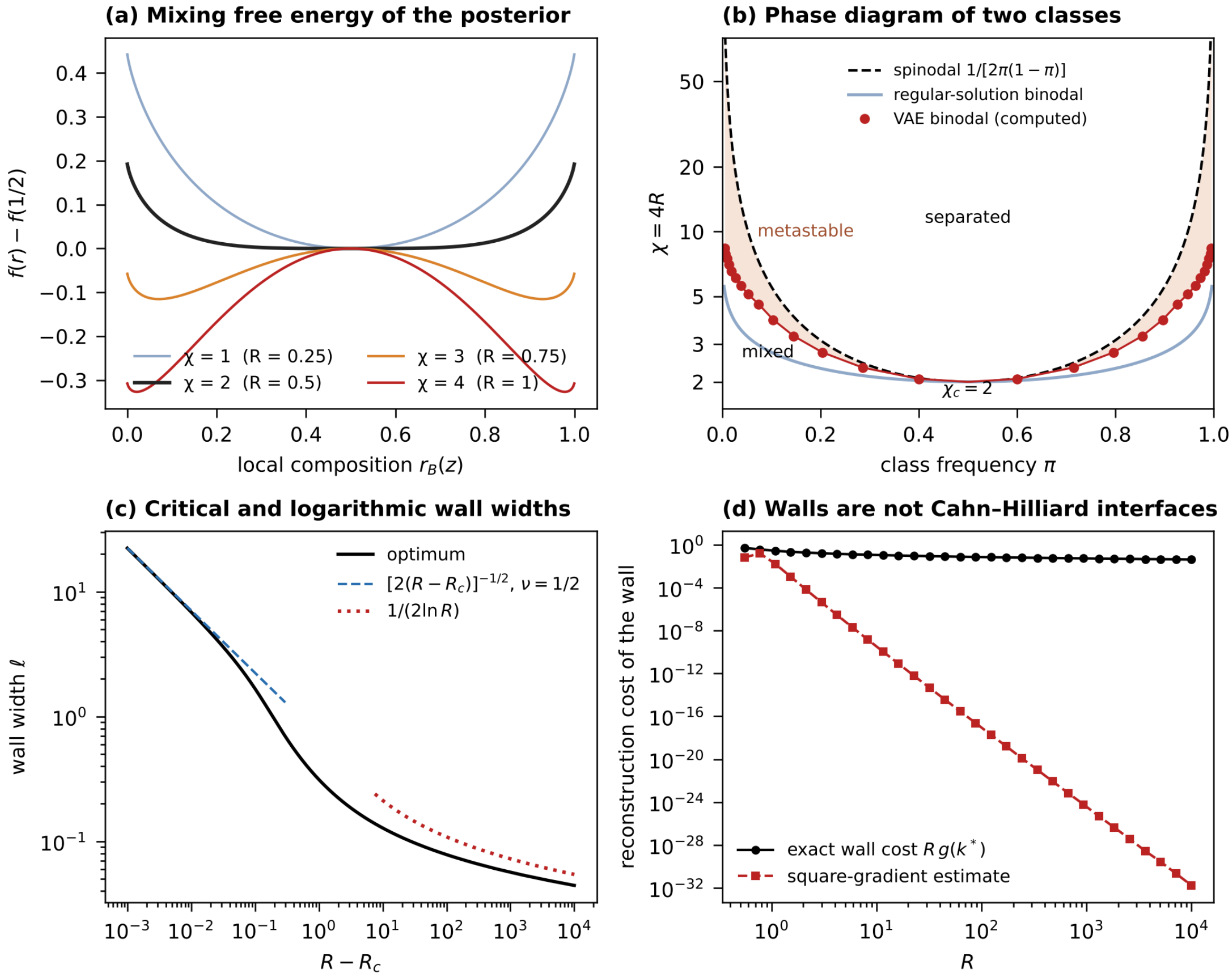

**Figure S1.** Equilibrium correspondence. (a) Local free energy $f(r)$ of equations (S2a)–(S2c) for two classes; the mixed state becomes unstable at $\chi_c = 2$, the collapse threshold $R_c = 1/2$. (b) Phase diagram of two classes: spinodal, regular-solution binodal and computed VAE binodal; the shaded region is metastable, where a rare class is supersaturated. (c) Optimal wall width, with the critical law $\ell \simeq [2(R - R_c)]^{-1/2}$ and the logarithmic law $\ell \approx 1/(2\ln R)$. (d) Exact reconstruction cost of a wall against the square-gradient estimate of equation (S3).

### S2.6 Rare prototypes at walls and reconstruction-metric dependence

**Adsorption and wetting.** The decoded content of an A/B wall lies on the segment between $a_A$ and $a_B$. A rare prototype $a_C$ with projection $t$ on this segment and perpendicular distance $h$ is reconstructed with squared error $h^2 + t^2|a_A - a_B|^2$ inside domain A but only $h^2$ at the wall point whose blend is closest to it, which gives the wall advantage of equation (18). When this advantage is small, C is dissolved into its nearest class or forms a tail island, and the wall is unchanged. A large advantage permits adsorption in the wall at low frequency and, once $\pi_C^0$ is comparable to the data mass of the wall, of order $1/R$, forms a layer that replaces the A/B wall by A/C and C/B walls; this requires $|a_C - a_A|$ and $|a_C - a_B|$ to be smaller than $|a_A - a_B|$, the analogue of complete wetting, $\sigma_{AB} > \sigma_{AC} + \sigma_{CB}$ [69]. For the $40 \times 40$ pixel prototypes of figure 5 the rounded square has $t = 0.50$ and $h^2/|a_A - a_B|^2 = 0.19$, and the rhombus $t = 0.03$ and $h^2/|a_A - a_B|^2 = 0.37$. The rhombus is dissolved into the circle class (for example at $\chi_{AB} = 12$ and $\pi_C^0 = 10^{-3}$) or forms a tail island. The rounded square is adsorbed for $\pi_C^0 \leq 10^{-4}$ when $\chi_{AB} \leq 16$ and up to $\pi_C^0 = 0.03$ at $\chi_{AB} = 8$; it forms a tail island over an intermediate range of $\pi_C^0$ that widens with $\chi_{AB}$, and a layer for $\pi_C^0 \geq 0.03$ when $\chi_{AB} \geq 16$. At $\chi_{AB} = 40$ the layer receives 5.6% of the prior at $\pi_C^0 = 0.03$.

**Reconstruction metric.** A weighted metric $M$ replaces $\chi_{ij}$ by $(a_i - a_j)^{\mathsf{T}} M (a_i - a_j)/(2\beta\sigma_x^2)$ and changes $t$ and $h$ with it. Comparing the images after smoothing with a Gaussian of 4 pixels moves the rounded square almost onto the circle–square segment ($t = 0.52$, $h^2/|a_A - a_B|^2 = 0.017$) and moves the rhombus beyond the circle end of the segment ($t = -0.23$, $h^2/|a_A - a_B|^2 = 0.14$), where the wall offers no advantage. At $\chi_{AB} = 16$ and frequencies 0.001, 0.003, 0.01 and 0.03 the rounded square forms a tail island, a tail island, a layer and a layer under the pixel metric, and is adsorbed, adsorbed, adsorbed and forms a layer under the smoothing metric; the rhombus forms a tail island at every frequency under the pixel metric and is dissolved at 0.001 under the smoothing metric (figure S2(f)).

The wall-advantage formula assumes projection inside the segment. Outside that interval, the nearest point of the segment is an endpoint; extending the same interior expression without clipping would assign a spurious wall advantage. This is why the smoothed rhombus at $t = -0.23$ is described as having no wall advantage. None of these metric calculations independently demonstrates that the prototype is a physical transition state.

## S3 Posterior family, continuous factors, symmetry and dimensionality

### S3.1 Continuous within-class variation and fixed decoder resolution

Let class $j$ vary along one linear factor with data variance $w_j^2$. The internal split decouples in the linear, well-separated-class approximation used here. The class-aggregate width $S_j$ is fixed by packing, while the split of $S_j$ into signal $c_j$ and noise $s_j$ is fixed by the class alone (main equations (19a) and (19b)). The threshold $A_j = 1/2$ applies to this linear internal factor and is not a universal threshold for nonlinear image factors.

A fixed absolute resolution limit can be modeled by a decoder receiving $z + \eta$ with $\eta \sim \mathcal{N}(0, \lambda^2)$. The class aggregates are then broadened to $S^2 + \lambda^2$. Because the smoothing scale is fixed in latent units, the affine compensation used in T2 is no longer available. This added-noise model tests decoder smoothing rather than a general finite-network capacity law.

Extra coordinates that remain prior-matched and carry no information add no loss to an existing representation. Its boundaries then extend through those coordinates, so a two-dimensional projection can hide walls without removing their probability. This passive extension does not imply that all higher-dimensional optima are identical. Whether a new coordinate becomes active depends on the reconstruction benefit, posterior family and training path. The rare-class escape examples of Section S2.5 illustrate a possible change in accessibility, not a universal invariance of the global optimum.

### S3.2 Covariance-dependent precision costs, stacks and faceting

**Origin.** Equations (20a) and (20b) give the gap between the precision costs of full-covariance and diagonal posteriors. It is the minimized reverse-KL penalty of a diagonal approximation at fixed full precision. It equals the posterior total correlation in two dimensions, but not generally in higher dimensions; for two directions it equals $-1/2\ln(1-\rho^2)$, with $\rho$ the correlation of the posterior precision. The pressure grows with information, so inactive directions feel none and can become passive. It selects local principal directions of the pulled-back data metric, as observed by Rolínek et al [70], but nothing in it ties an axis to a named physical factor [6]; auxiliary variables or non-Gaussian factors are needed for that [8, 93]. This fixed-precision penalty is not the separately reoptimized wall-orientation excess: the former can grow without bound, whereas the latter has the finite strong-separation limit in main equation (21).

**Mean-field theory.** The identity (S2a,S2b,S2c) holds for any posterior covariance; what changes between covariance families is the price of sharp walls. In strong segregation, a wall between classes $i$ and $j$ with normal $n_{ij}$ requires a Mahalanobis sharpness $k_{ij}$, so class $j$ must carry a precision tensor

$$P_j \simeq I + \sum_{i\in\partial j} k_{ij}\; n_{ij} n_{ij}^{\mathrm{T}} \tag{S10}$$

where the sum runs over the neighbors of $j$. The effective free energy is

$$\mathcal{L} \simeq \sum_j \pi_j^0 \left[1/2\,|\mu_j|^2 + \Phi(P_j)\right] + \sum_{i<j} R_{ij}\; g(k_{ij}) \tag{S11}$$

with $g$ of equations (10a) and (10b) and a precision cost $\Phi$ that depends on the posterior family (table S2).

Table S2. Precision cost and preferred layout for three posterior families; $\Phi_{\mathrm{iso}} \geq \Phi_{\mathrm{diag}} \geq \Phi_{\mathrm{full}}$.

| posterior | $\Phi(P_j)$ | preferred layout |
|---|---|---|
| isotropic | $d/2\,\ln\lambda_{\max}(P_j)$ | all directions are paid for; compact mosaics, shells, Y-junctions |
| full covariance | $1/2\,\ln\det P_j$ | each independent wall orientation costs about $1/2\ln(1+k)$ per class; one-dimensional stacks of any orientation |
| diagonal | $1/2\sum_a \ln\left(P_j\right)_{aa}$ | stacks locked to a latent axis; cubic wall anisotropy |

**Anisotropic wall tension.** A single wall at angle $\theta$ to a latent axis costs more under a diagonal posterior, because precision is cheap only along axes. The posterior responds by tilting the wall normal, $n \propto \Sigma^{-1}\Delta\mu$, back towards the axis, and asymptotically the extra cost is equation (21): $1/2\ln 2 = 0.347$ at 45° in two dimensions and $1/2\ln d$ along a body diagonal in $d$ dimensions. The anisotropy vanishes at the collapse point and grows as walls sharpen (Figure S3(a)); it is about a quarter of the total wall cost at large $R$.
For the tested prototype systems, isotropic posteriors favor compact layouts, whereas anisotropic posteriors favor stacks with a passive coordinate. Full covariance permits the stack to rotate; diagonal covariance locks its orientation (Figure S3(b)). These optimized layouts are not universal predictions for trained image networks.
The same orientation dependence can facet a rotational ring, as illustrated in Figure S3(d). This geometrical effect does not establish identification of physical factors.
A full-covariance stage followed by a diagonal restriction changes which local minimum is reached in the tested class-aggregate model (Figure S2(e)); it is not a general neural-network training prescription.

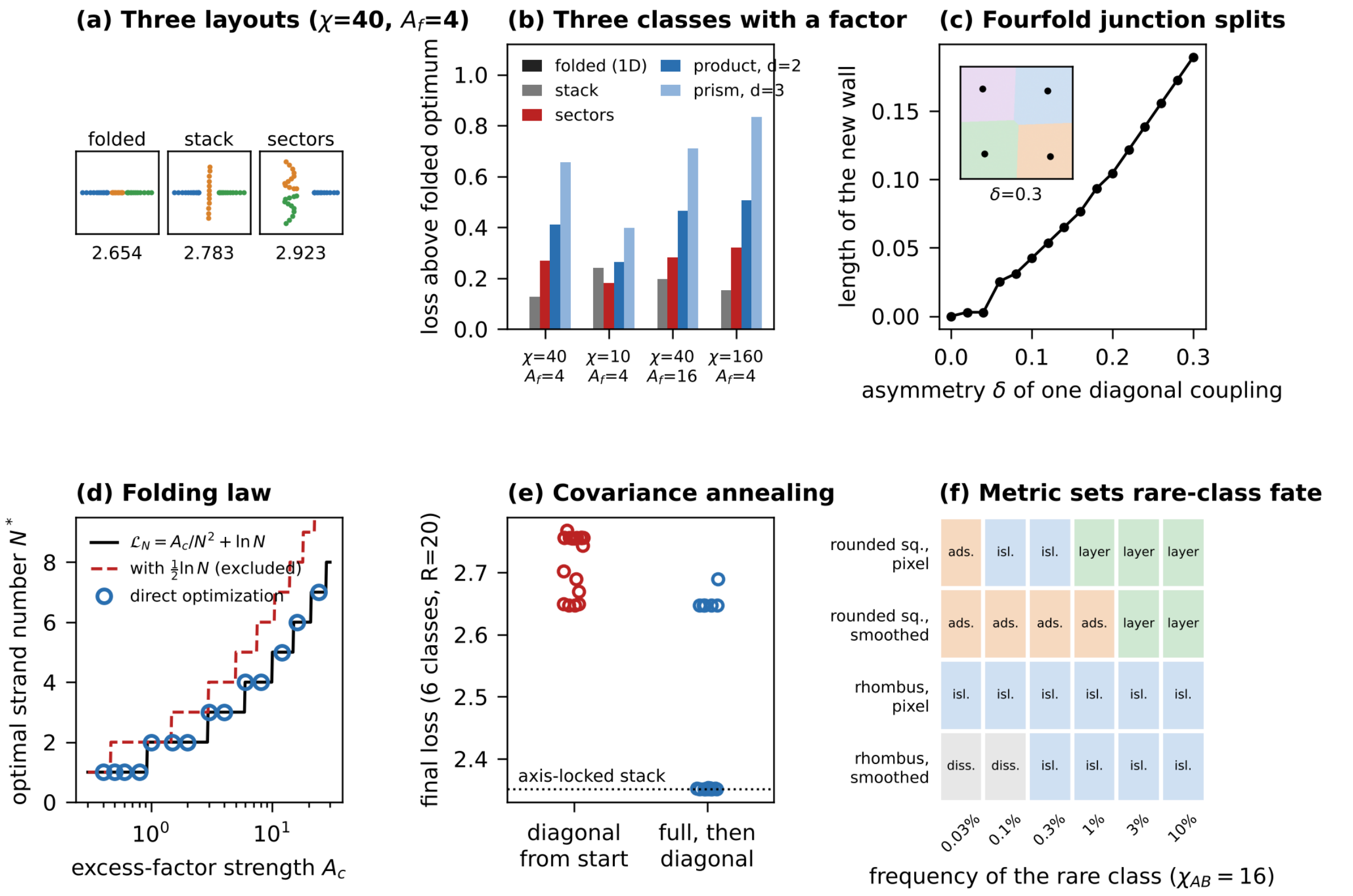

**Figure S2.** Numerical tests of the extended correspondence. (a) Encoded data of three classes carrying a continuous factor ($\chi = 40$, $A_f = 4$) in three layouts, with their losses: the class label folded into the factor axis, a stack, and 120° sectors with a triple junction. (b) Losses of five layouts of the same system relative to the optimum, in which the class label is folded into the factor axis; disentangled layouts, including the prism available only in three dimensions, are highest. (c) Length of the wall that replaces a fourfold junction when the coupling between one pair of opposite classes is reduced by $\delta$; inset, partition at $\delta = 0.3$. (d) Optimal strand number from direct optimization against the staircases of the folding law with $\ln N$ and with $1/2 \ln N$. (e) Final losses of six equal classes ($R = 20$) from eight random starts, with a diagonal posterior throughout or after a full-covariance stage. (f) Fate of two rare classes on the circle–square wall under the pixel metric and a smoothing metric.

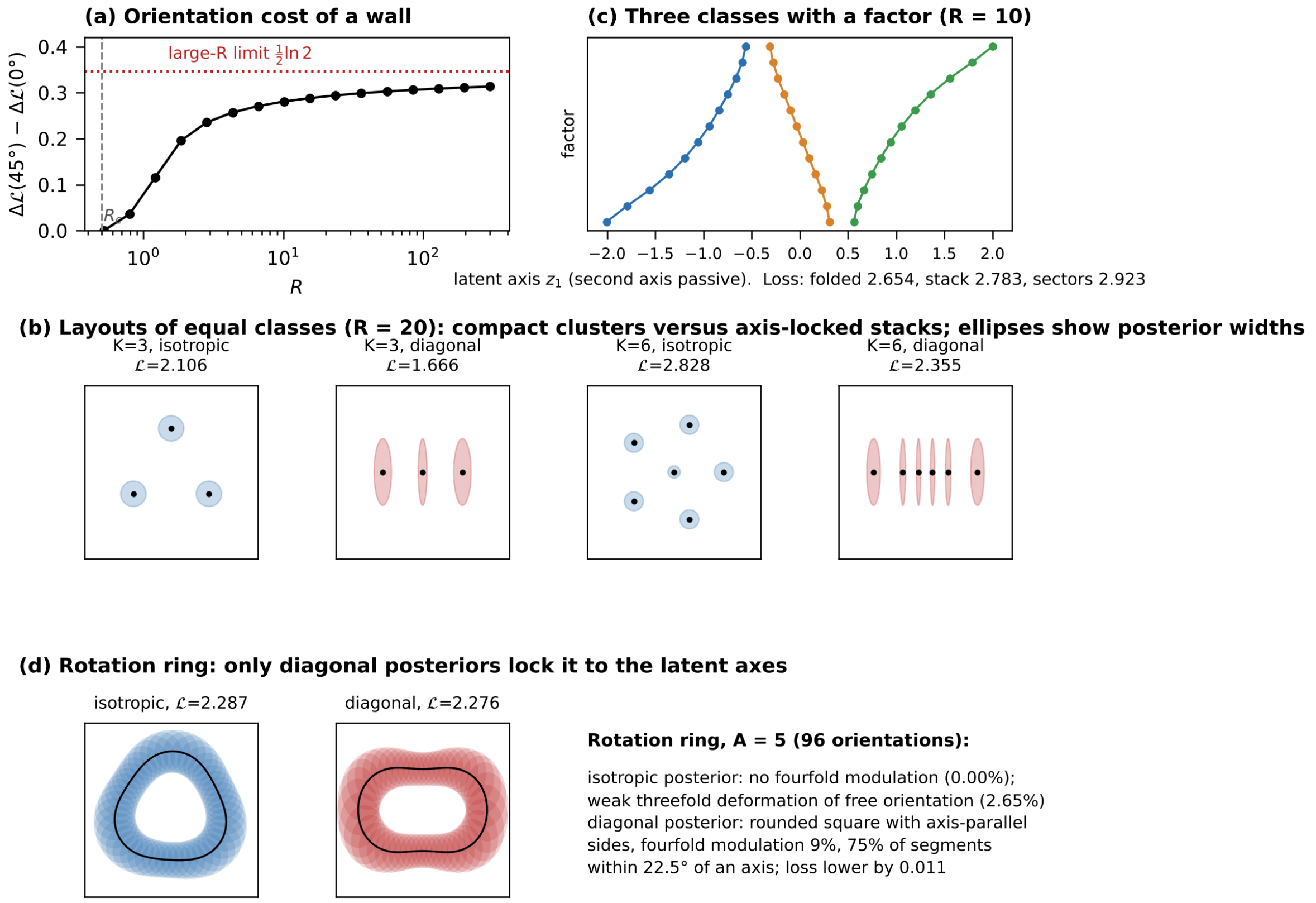

**Figure S3.** Posterior covariance and latent anisotropy. (a) Extra loss of a two-class wall at 45° to the latent axes, relative to an axis-parallel wall, for a diagonal posterior; the dotted line is the large-$R$ limit ½ ln 2. (b) Optimal layouts of three and six equal classes at $R = 20$ with isotropic and diagonal posteriors; ellipses show posterior widths. (c) Encoding of three classes carrying a continuous factor at $R = 10$ in the folded layout, in which the class label is folded into the factor axis. (d) Rotation ring at $A = 5$ with isotropic and diagonal per-point posteriors; the diagonal posterior squares the ring onto the latent axes.

### S3.3 Rotation, quotient orbits, cuts and the Cards comparison

**Ring model.** For an asymmetric rotating pattern the full angle is identifiable and the isotropic ring ansatz gives the circular solution. For an exactly n-fold symmetric pattern, however, the encoder can depend only on the quotient angle nθ modulo 2π. Rewriting the ring problem in that coordinate reduces it to the one-fold case: the ring forms continuously at the same threshold for every n, a degree-one quotient encoding winds n times as the physical angle traverses 2π, and the core keeps the same asymptotic scaling with the quotient coordinate. Explicit angle latents and weakly broken symmetry are treated below. Symmetry alone imposes the shorter periodicity. A noncollapsed loop avoiding its winding center may have any integer-multiple winding; the circular ansatz selects one traversal of the quotient orbit.

**Explicit circular angle.** Take a uniform prior on the circle and a von Mises posterior of concentration $\kappa$ centered on the true angle. With $A_m$ the variance of harmonic $m$ divided by $2\beta\sigma_x^2$, the KL term and the reconstruction cost for a rigidly rotated template decoder are

$$\mathrm{KL} = \kappa\frac{I_1(\kappa)}{I_0(\kappa)} - \ln I_0(\kappa) \approx \frac{1}{2}\ln 2\pi\kappa - \frac{1}{2} \quad \text{(S12a)}$$

$$\mathcal{L}_{\mathrm{rec}} = \sum_m 2\,A_m\left(1 - \frac{I_m(\kappa)}{I_0(\kappa)}\right) \approx \frac{C}{\kappa} \quad \text{(S12b)}$$

with $C = \sum_m A_m\, m^2$, so that $\kappa^* = 2C$. The angular variance $1/(2C)$ is half that of the plain-VAE ring, the angle uses one latent dimension instead of two, and because the aggregate angle posterior is uniform it equals the prior: every prior angle decodes to a valid pose and there is no core. The content latent sees the de-rotated pattern and reduces to the rotation-free problem. For an unconstrained conditional-mean decoder the exact finite-concentration cost instead uses one minus the squared harmonic attenuation; both costs have the same leading large-concentration term, but their finite-concentration losses must not be interchanged.

**Gaussian prior on a periodic angle.** With a prior $\mathcal{N}(0, \sigma_\theta^2)$ on a real-valued angle and a $2\pi$-periodic decoder, every prior sample still decodes to a valid pose, so there is no latent wall. Generated orientations follow the wrapped normal, whose harmonic $m$ has amplitude $e^{-m^2\sigma_\theta^2/2}$, and encodings fill one fundamental interval, which adds $\pi^2/(6\sigma_\theta^2)$ to the KL term. A real-valued head attempting faithful angle recovery cannot give a globally continuous inverse of the orbit [72, 73]; its angle error $e(\theta)$ obeys $e(\theta + 2\pi) = e(\theta) - 2\pi$ and must sweep through $2\pi$ somewhere: an encoder cut, a wall in orientation space rather than in latent space. Modeling encoder capacity by the number $K$ of angular Fourier modes it resolves, numerical optimization gives

$$\Delta\mathcal{L}_{\text{cut}} \approx \frac{1.3\,A}{K} \tag{S13a}$$

$$f_{\text{cut}} \approx \frac{0.27}{K} \tag{S13b}$$

for a single harmonic, where $f_{\text{cut}}$ is the fraction of orientations decoded more than a quarter period wrong. The cut narrows only with capacity, not with $\beta$.

**Symmetric patterns.** For an $n$-fold symmetric pattern the angle head must satisfy $n\varphi \equiv \psi$ modulo $2\pi$, where $\psi = n\theta$ is the quotient angle. A circular head of degree $d$ gives an error of degree $nd - 1$, so the error sweeps through $2\pi$ at least

$$N_{\text{cut}} = \min_{d\in\mathbb{Z}}|nd - 1| = 1 \qquad (n \geq 2) \tag{S14}$$

times per orbit, and a real-valued head ($d = 0$) has one cut. Only an angle valued in $SO(2)/C_n$, decoded by rotating through $\psi/n$, is cut-free.

**Weak symmetry breaking in a plain VAE.** Add a weak harmonic $A_\varepsilon$ at $m = 1$ to a strong harmonic $A_s$ at $m = n$. In a two-dimensional latent the encodings trace a closed curve whose winding number $w$ about the core labels a topological sector: $w = 1$ resolves the full orientation and $w = n$ only the quotient angle. For ideal rings,

$$\mathcal{L}_{w=1} \approx 1 + \ln\frac{n^2A_s + A_\varepsilon}{2} \tag{S15a}$$

$$\mathcal{L}_{w=n} \approx 1 + \ln\frac{A_s}{2} + A_\varepsilon \tag{S15b}$$

so full resolution wins only above $A_\varepsilon^* \approx \ln(n^2 + A_\varepsilon^*/A_s) \approx 2\ln n$; below it the weak feature is averaged away. A regular closed plane curve with only transverse double points and winding number $w$ about a point has at least $|w| - 1$ self-intersections (by induction on transverse double points: splitting at one gives two closed curves whose winding numbers add). In the quotient sector, partial resolution of the weak feature therefore forces at least $n - 1$ crossings at which different branches coincide in their means. The decoder must average them when their full posteriors coincide; variable covariances can retain information despite coincident means. The winding number changes only when the curve passes through the core, so the sectors are separated by a barrier.

**Explicit angle locked to the quotient.** An angle head that reads only the strong harmonic has degree 0 in $\psi$, a homotopy class that continuous training cannot leave. It returns one fundamental domain, so inputs from orientation sector $k$ reach the content encoder as $n$ equally frequent classes with couplings

$$R_{kl} = \frac{A_\varepsilon}{2}\left[1 - \cos\frac{2\pi(k-l)}{n}\right] \tag{S16}$$

For classes with small displacements $\delta_i$ from the centroid, expanding T1 gives $\mathbb{E}_{\bar{q}}[r_ir_j] \approx \pi_i^0\pi_j^0(1 + \delta_i \cdot \delta_j)$, and with $\text{KL} \approx \frac{1}{2}\sum_i \pi_i^0\,|\delta_i|^2$ the Landau quadratic form is

$$\mathcal{L}^{(2)} = \frac{1}{2}\sum_i \pi_i^0 \, |\delta_i|^2 + 4\sum_{i<j} R_{ij} \; \pi_i^0 \pi_j^0 \; \delta_i \cdot \delta_j \tag{S17}$$

For $n$ equal classes the collapsed state is unstable when the smallest eigenvalue of $R$ on $\sum_i \delta_i = 0$ falls below $-n/4$. The circulant matrix of equation (S16) has eigenvalues $\lambda_p = \frac{nA_\varepsilon}{2}\left[\delta_{p0} - \frac{1}{2}\left(\delta_{p1} + \delta_{p,n-1}\right)\right]$, so the onset is at $A_\varepsilon = 1/2$ for $n = 2$ and $A_\varepsilon = 1$ for $n \geq 3$. For $n = 3$ these are the couplings of the three-state Potts model, whose cubic invariant makes the onset weakly first-order [94]. The unstable mode, $\delta_k \propto e^{2\pi i k/n}$, places the $n$ domains on a ring in cyclic order: the content latent rebuilds the orbit that the angle head failed to resolve.

Cards comparison (Figure 7). Four rotation/shear conditions are used: Cards 1 (12°/1°), Cards 3 (120°/1°), Cards 2 (12°/20°) and Cards 4 (120°/20°). Figure 7 pairs ordinary-VAE decoded grids with suit-colored encoded distributions. In the corresponding rotationally invariant VAE, the content distributions are more compact than the ordinary-VAE trajectories at low shear but remain extended or branched at high shear, so removing orientation does not remove independent deformation. The angular encodings follow the imposed angle with suit-dependent offsets and a larger spread under high shear. These observations are qualitative.

**S3.4 Dimension counting and label folding**

A region in which $m$ continuous factors vary and $s$ classes meet needs $m$ directions for the factors and $s - 1$ for the composition, because the posteriors sum to one. A locally disentangled representation, in which each latent direction changes either the composition or one factor, therefore requires

$$d \geq m + s - 1 \tag{S18}$$

the latent counterpart of the Gibbs phase rule [95]: $s$ phases meet generically on a set of dimension $d - s + 1$, which can carry the $m$ factors only if $d - s + 1 \geq m$. When $d < M = m + s - 1$, the $M$ unit directions must overlap; the Gram matrix has rank at most $d$ and trace $M$, which gives the Welch bound [96]

$$\sum_{i \neq j} \left|v_i^\top v_j\right|^2 \geq \frac{M(M-d)}{d} \tag{S19}$$

A two-class wall carrying one factor in two dimensions is compatible with disentanglement ($M = d = 2$); a triple junction carrying one factor in two dimensions is not ($M = 3 > d$); and a fourfold junction in two dimensions violates the count even without factors, which is why it exists only at symmetry (section S2.4).

To test what the VAE does with this constraint, three equal classes each carry the same uniform continuous factor (12 levels), so $m = 1$ and $s = 3$. Five layouts were compared (figure S2(a, b)): the classes as successive strands along one axis with the factor running back and forth (folded); a two-dimensional stack in which the middle class runs its factor along the second axis; 120° sectors with the factor along the radius and a triple junction at the center; the disentangled product, with classes on one axis and the factor on the other ($d = 2$); and the disentangled prism, a triangle of classes times the factor axis ($d = 3$), the only layout that satisfies equation (S18) at the junction. The losses of the disentangled layouts follow exactly from separability: when classes and factor occupy different axes and are independent in the data, the posterior over both factorizes, the optimal decoder is the sum of the two conditional means, and the loss is the sum of the class-only and factor-only losses.

The tested folded layouts have lower loss than the disentangled product and prism constructions. Their leading cost is the discrete-label information rate added to the factor cost, $\ln 3 + 1/2 + 1/2 \ln 2A_f$. Encoding the label along an already precise axis avoids purchasing posterior precision a second time. Figure S2(b) compares the candidate losses.

### S3.5 Under-dimensional representations: fold law and competing branches

**Folding of excess factors.** Let one latent axis carry a factor $f$ and let a factor $c$ be added, both uniform, with strengths $A_f$ and $A_c$. In the serpentine fold, $c$ is quantized into $N$ equal levels, leaving residual variance $w_c^2/N^2$, and each level occupies a strand of length $L_z/N$ along which $f$ runs in alternating directions. The factor $f$ is then continuous across strand boundaries, and $c$ jumps by $\Delta = \sqrt{12}\, w_c/N$. Posterior noise $s$ along a strand produces an error in $f$ amplified by the compression of the strand, and T2 gives $L_z^2 = 12(1 - s^2)$. Per data point,

$$\mathcal{L}_N(s) = \frac{A_c}{N^2} + \frac{A_f N^2 s^2}{1-s^2} - \frac{1}{2}\ln s^2 + F_N \tag{S20}$$

Minimizing over $s$ for $s \ll 1$ gives $s^2 = 1/\left(2A_f N^2\right)$ and the reduced loss $\mathcal{L}_N = A_c/N^2 + \ln N + \frac{1}{2} + \frac{1}{2}\ln 2A_f$. Comparing $N$ and $N+1$ strands gives the transitions

$$A_c^{(N\to N+1)} = \ln\frac{N+1}{N}\,\frac{N^2(N+1)^2}{2N+1} \approx \frac{N(N+1)}{2} \tag{S21}$$

**Fold walls.** A strand boundary separates adjacent levels of $c$, with coupling $R_{\mathrm{fold}} = \Delta^2/(8\beta\sigma_x^2) = 3A_c/N^2$, equal to 3/2 at $N = N^*$. With uniform encoding density $1/L_z$, the posterior overlap across one boundary is $\int \Phi(1-\Phi)\, dz = s/\sqrt{\pi}$, so

$$F_N \approx \frac{(N-1)\, A_c}{N^3}\sqrt{\frac{6}{\pi A_f}} \tag{S22}$$

This is an upper estimate: optimized encoders come in 0.01–0.15 below the reduced loss even without $F_N$, because they also shape the strand density. For Gaussian rather than uniform factors, two-level quantization leaves $(1 - 2/\pi)\, w_c^2$, and the first fold appears at $A_c = \frac{\pi}{2}\ln 2 \approx 1.09$; this value was not checked numerically.

**Instability of the unfolded state.** Let the unfolded encoder map $f$ linearly onto the latent coordinate $y$ with posterior width $s$, and perturb its means by $\varepsilon\,(c/w_c)\cos ky$, which by the Cauchy–Schwarz inequality is the optimal dependence on $c$ for a given spatial profile. The optimal decoder acquires $\mathbb{E}[c \mid z] = \varepsilon\sqrt{A_c}\, k\sin(kz)\, e^{-k^2 s^2/2}$ in loss units, which lowers the Bayes risk of $c$ by $\frac{1}{2}\varepsilon^2 A_c k^2 e^{-k^2 s^2}$. The same perturbation adds latent noise $\varepsilon^2\cos^2 ky$ independent of $f$, costing $\frac{1}{2}\varepsilon^2 A_f$, and raises the second moment, costing $\frac{1}{4}\varepsilon^2$ in the KL term after rescaling. Hence

$$\delta\mathcal{L} = \frac{\varepsilon^2}{2}\left[A_f + \frac{1}{2} - A_c k^2 e^{-k^2 s^2}\right] \tag{S23}$$

The bracket is smallest at $k = 1/s$, where $k^2 e^{-k^2 s^2} = 1/(e s^2)$; with $s^2 = 1/\left(2A_f\right)$ this gives the instability threshold $A_c^{\mathrm{sp}} = \frac{e}{2}\left(1 + \frac{1}{2A_f}\right)$, at wavelength $2\pi s$. Two further unstable modes are localized at the ends of the range of a bounded factor, where points can move outward to encode $c$ without overlapping other encodings; they were unstable at every tested strength ($A_c \geq 0.8$).

**Three factors in two dimensions.** With $A_1 \geq A_2 \geq A_3$ and $d = 2$, the linear optimum drops factor 3. Folding it into the axis of factor 2 produces stripes, and the loss approaches

$$\mathcal{L} \approx \sum_{j=1}^{3}\left(\frac{1}{2} + \frac{1}{2}\ln 2A_j\right) \tag{S24}$$

For the bounded uniform factors, a modulated branch appears before the ordered serpentine becomes favorable. Its modulation begins near the ends of the data range, while the developed serpentine requires a global rearrangement. Figure S4 compares the branch losses and representative maps.

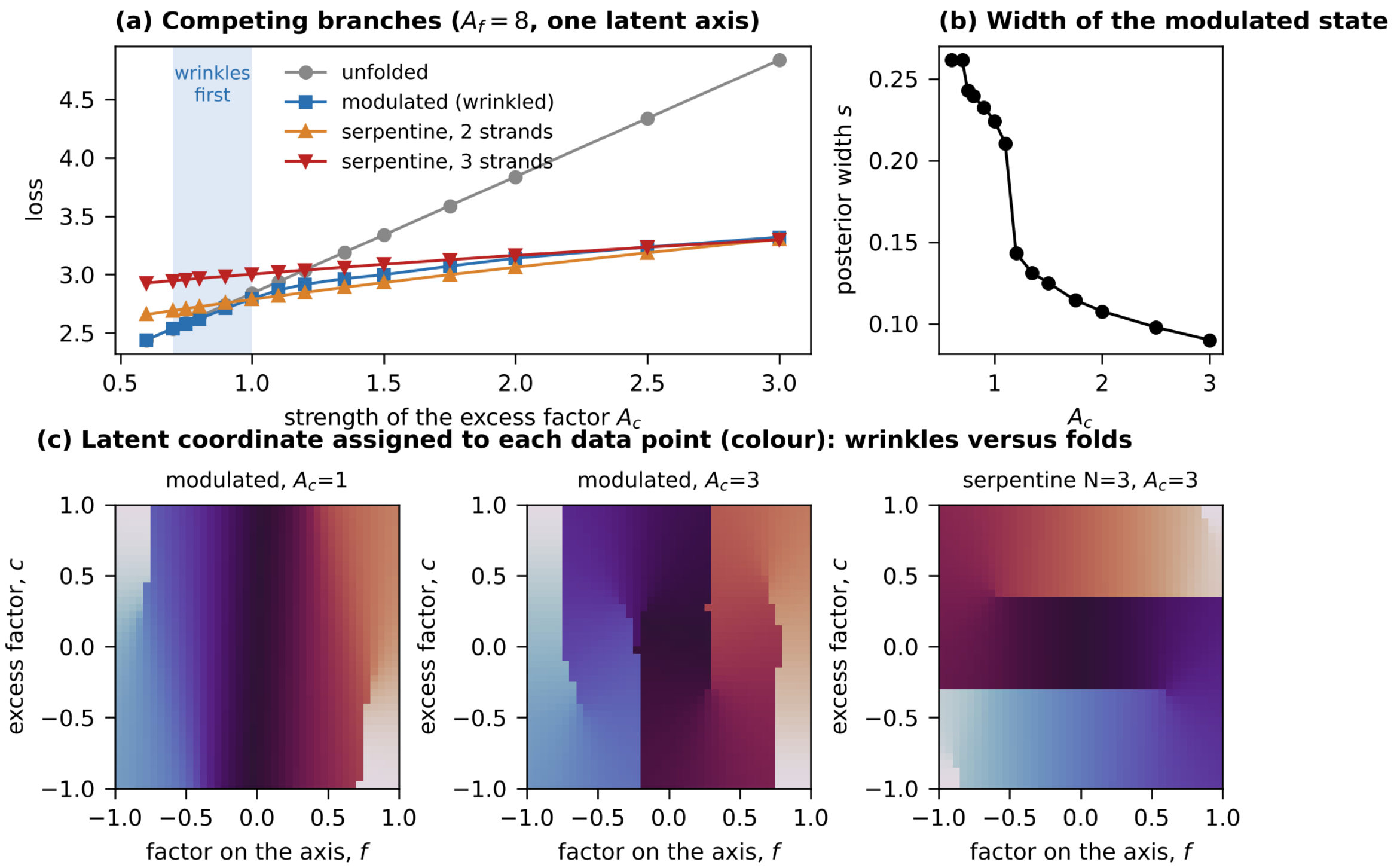

**Figure S4.** Dimensional frustration. (a) Losses of the unfolded, modulated and serpentine branches against the strength of the excess factor; the modulated (wrinkled) state appears first. (b) Posterior width of the modulated state. (c) Latent coordinate assigned to each data point: modulated states at $A_c = 1$ and 3 and the three-strand serpentine at $A_c = 3$.

**Strand exponent.** If each strand carries, besides its information cost, a fixed cost $\gamma$, for example from finite decoder capacity, minimizing $A_c/N^2 + \ln N + \gamma N$ gives

$$N^{*2}(1 + \gamma N^*) = 2A_c \tag{S25a}$$

$$\frac{d\ln N^*}{d\ln A_c} = \left(2 + \frac{\gamma N^*}{1+\gamma N^*}\right)^{-1} \tag{S25b}$$

so the exponent falls from $1/2$ to $1/3$ as the per-strand cost takes over, and a gradient or nonlocal cost would give $1/4$. A measured exponent in a trained network would help discriminate these cost models only after the relevant scaling regime and folded state were established. Direct optimization of the idealized model (figure S2(d)) gives $N^* = 1, 3, 4$ and 6 at $A_c = 0.6$, 4, 8 and 16, as equation (S21) predicts; a reduced loss with $1/2 \ln N$ in place of $\ln N$, which would place the first fold at $A_c \approx 0.46$ and give $N^* = 2, 4, 6$ and 8, is excluded. For several excess factors the losses add, $N_j^{*2} = 2A_j$, and the number of local branches is $\prod_j \sqrt{2A_j}$.

Trained under-dimensional networks were used as a deliberately stricter test of the folding prediction. Synthetic data contained three independent linear factors embedded in 64 dimensions and were fitted by two-dimensional VAEs from random initialization. The networks consistently organized two factors, but the third factor was not robustly converted into the ordered strand staircase of the idealized optimum; depending on regularization and initialization it was largely suppressed or weakly entangled. Figure S5 shows representative runs. These calculations therefore test optimization accessibility rather than the existence of folded minima: the latter is supported by the branch comparisons and direct optimization above, whereas spontaneous convergence to the equilibrium fold remains open.

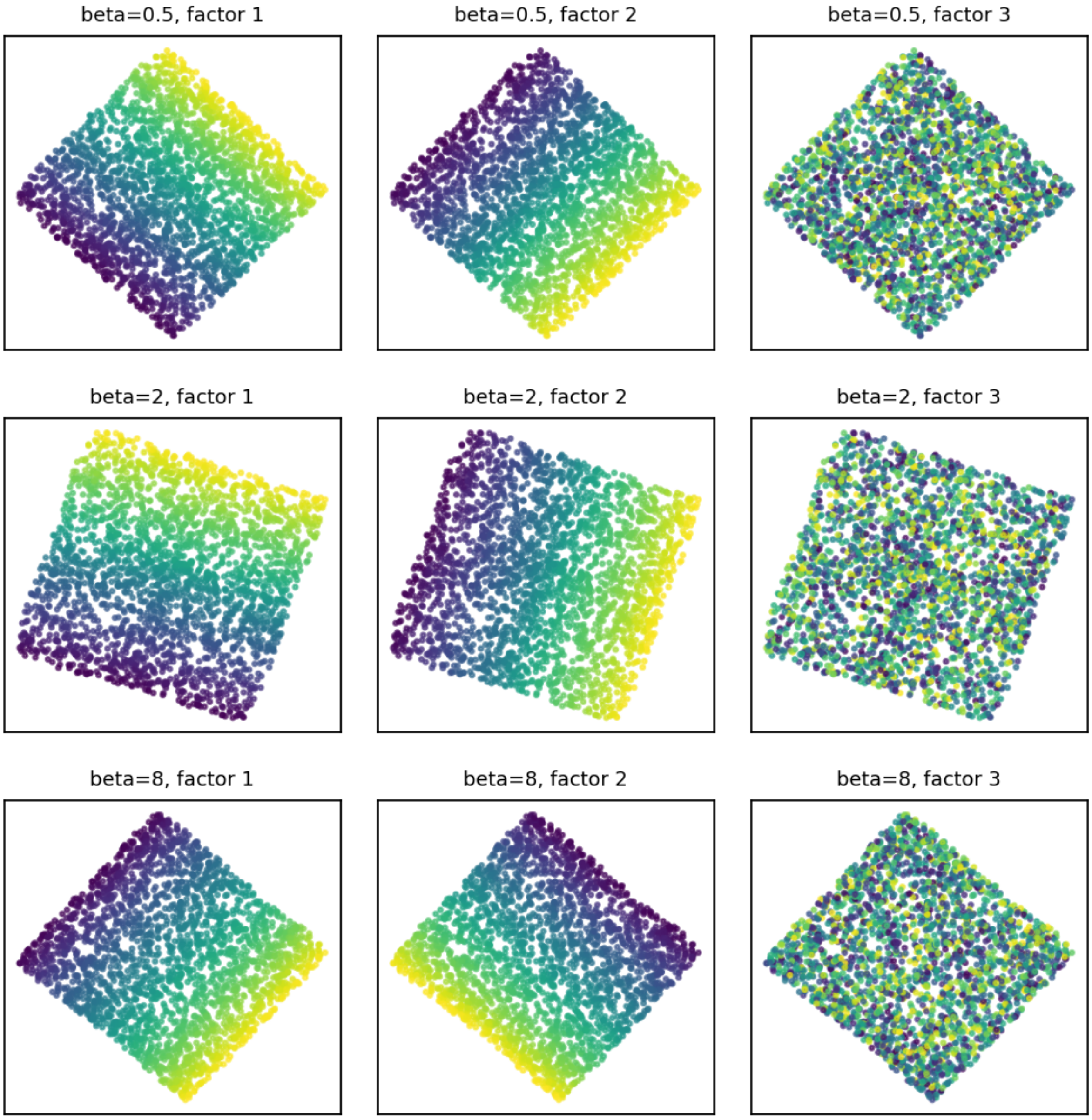

**Figure S5.** Randomly initialized trained two-dimensional VAEs on synthetic data with three continuous factors, colored by each factor. The first two factors are consistently represented, whereas the third factor is not robustly organized into the equilibrium folded-strand structure predicted by the idealized optimization model. These runs therefore provide a negative control: equilibrium folding is an available low-dimensional solution, but ordinary gradient training need not find it.

The neural and nonparametric results are not interchangeable. A nonparametric encoder assigns a separately optimized posterior to each data point; the trained network must learn an amortized mapping. Competing minima in the reduced model provide a plausible accessibility mechanism (Section S3.6), but no neural loss barrier has been measured. The folding-search protocol is given in Section S5.4 and the nonparametric implementation in Section S7.2.

**S3.6 Training dynamics, annealing and negative temporal tests**
Figure 2(d) tests the reduced-model escape, relaxation, and annealing rates. A separate trained-digit VAE trajectory exhibits nonmonotonic wall-width evolution, so the reduced-model time scales do not quantify neural-network training.
The two-class order-parameter gradient flow has escape rate 2(R − Rc) and an ordered-state relaxation rate approaching 4(R − Rc). A linear ramp R(t) = Rc + t/τQ gives the delay in main equation (24). These rates concern the specified reduced dynamics. They do not predict wall-clock time or training-step counts for a neural network.

**S4 Alternative generative and target-shaped objectives**
**S4.1 Continuous generators, flows and prescribed or learned priors**

Existing isoperimetric results for continuous generators [51, 52] make this idea precise. If separated data components are generated from one connected Gaussian latent distribution, their preimages cannot occupy the latent space without a separating boundary region. In the notation used here, the bound becomes a statement about walls. Let $T: \mathbb{R}^d \to \mathbb{R}^D$ be continuous and $L$-Lipschitz, and let $\nu$ be the push-forward of the standard Gaussian $\gamma$. Let $C_1, \dots, C_K$ be data components at pairwise distance at least $\Delta$, with open $\delta$-neighborhoods $U_i$, $\delta < \Delta/2$. Suppose $\nu$ places mass at most $\eta$ outside $\cup_i U_i$, and let $A_i = T^{-1}(U_i)$ with $p = \gamma(A_1)$. Then the $A_i$ are disjoint open cells; the wall set $W = \mathbb{R}^d \setminus \cup_i A_i$ is non-empty for $K \geq 2$, separates them and satisfies $\gamma(W) \leq \eta$. The resulting bound is main equation (25).
Main equation (25) applies to the stated continuous L-Lipschitz generator and separated component neighborhoods. Here Φ and φ are the Gaussian cumulative distribution and density. The inequality is a regularity-dependent lower bound, not a statement that every limiting transport to discrete modes has a finite endpoint Lipschitz constant.

**Contrastive flow matching.** For a negative-pair weight in [0, 1), minimizing $\mathbb{E}\|v - (x_1 - x_0)\|^2 - \lambda\mathbb{E}\|v - (\tilde{x}_1 - x_0)\|^2$ with an independent negative $\tilde{x}_1$ gives $v^* = (\mathbb{E}[x_1 \mid x_t] - \lambda\bar{x})/(1-\lambda) - \mathbb{E}[x_0 \mid x_t]$, where $\bar{x}$ is the data mean.

The prior $\frac{1}{2}\mathcal{N}(\delta, 1-\delta^2) + \frac{1}{2}\mathcal{N}(-\delta, 1-\delta^2)$ retains unit total variance. It becomes bimodal for $\delta > 1/\sqrt{2}$, reducing probability at the retained decoder transition. If the prior is learned jointly with the class aggregates, the separated limit has $\delta \to 1$, $m = \delta$ and $s^2 = 1 - \delta^2$. The aggregate then matches the prior and the loss approaches the separated-branch class-entropy limit ln2. With fewer prior components than encoded classes, some transitions must remain within populated modes. This learned-prior limit is distinct from the fixed-decoder prescribed-prior calculation in Figure 8(c).

Figure 8(a–c) calculation. The VAE decoder is $D(z) = \tanh(mz/s^2)$ with $m = 1.5$ and $s = 0.45$; the ambiguity interval has half-width $(s^2/m)\text{atanh}(0.8)$. For the flow interpolant $x_t = (1-t)x_0 + tx_1$, with standard-normal $x_0$ and equally probable $x_1 = \pm 1$, the conditional mean is $\mathbb{E}[x_1 \mid x_t] = \tanh[tx_t/(1-t)^2]$. The full ambiguous width is $2(1-t)^2\text{atanh}(0.8)/t$. Panel (c) evaluates $1/2\,\mathcal{N}(-a, 1-a^2) + 1/2\,\mathcal{N}(a, 1-a^2)$ at $a = 0, 0.5, 0.75, 0.9, 0.97$, with the decoder fixed. Wall probabilities are integrated by the Gaussian cumulative distribution. None of these curves requires neural training.

**S4.2 DKL objective, class separation and feature relevance**

To determine how target information sets latent geometry, we consider two classes of $n$ points each whose targets differ by $\delta$, and embed each class at a single point. Let $c = \exp(-\Delta^2/2\lambda^2)$ be the kernel coupling of the two classes at latent separation $\Delta$, and $A = n\sigma_f^2$. The kernel matrix has a symmetric class mode with eigenvalue $A(1+c) + \sigma_n^2$, an antisymmetric class mode with eigenvalue $v = A(1-c) + \sigma_n^2$, and within-class modes with eigenvalue $\sigma_n^2$. The target difference lies entirely in the antisymmetric mode, so up to constants

$$\mathcal{L}(c) = \frac{n\delta^2}{4v} + \frac{1}{2}\ln v + \frac{1}{2}\ln[A(1+c) + \sigma_n^2] \tag{S26}$$

Neglecting variation of the symmetric-mode term, the first two terms are minimized at $v = n\delta^2/2$. For noise small relative to the class signal, matching this value and expanding the kernel gives the small-separation approximation:

$$\frac{\Delta}{\lambda} \simeq \frac{\sqrt{[\delta^2 - 2\sigma_n^2/n]_+}}{\sigma_f}, \qquad \Delta/\lambda \ll 1 \tag{S27}$$

Here the positive part is max(u, 0). The approximation describes noise-limited merging and approximately target-proportional separation on its small-distance branch. Detachment requires comparison with the decoupled solution using the full evidence (S26), not extrapolation of (S27). Direct optimization of a three-cluster model reproduces the middle regime to within 2% for $\delta \le 0.5\,\sigma_f$.

Within the respective separated-VAE and small-distance DKL regimes, the contrast is as follows. In the VAE, classes separate by $2s\sqrt{k_{ij}}$ with $k_{ij} \approx 2\ln R_{ij}$ (section 3), a distance set by how different the classes look. In DKL they separate by $\lambda\delta/\sigma_f$, a distance set by how different their targets are. The VAE orders its domains by structure; DKL orders them by target.

To derive equations (S26) and (S27), order the $2n$ points so that the first $n$ belong to class 1 and the rest to class 2, with targets $+\delta/2$ and $-\delta/2$ after centering. With all points of a class at one latent position, the kernel matrix is

$$K + \sigma_n^2 I = \frac{A}{n}\begin{pmatrix} J & cJ \\ cJ & J \end{pmatrix} + \sigma_n^2 I \tag{S28}$$

where $J$ is the $n \times n$ matrix of ones. Its eigenvectors are the class-symmetric vector, with eigenvalue $A(1+c) + \sigma_n^2$; the class-antisymmetric vector, with eigenvalue $v = A(1-c) + \sigma_n^2$; and $2(n-2)$ zero-mean within-class vectors, with eigenvalue $\sigma_n^2$. The target vector is antisymmetric with squared norm $n\delta^2/2$, which gives equation (S26). Setting $\partial\mathcal{L}/\partial v = 0$ for the first two terms gives $v^* = n\delta^2/2$. Within this approximation, because $v \ge \sigma_n^2$, the classes merge ($c = 1$) when $n\delta^2/2 \le \sigma_n^2$. Otherwise $1 - c = (n\delta^2/2 - \sigma_n^2)/A$, and with $1 - c \approx \Delta^2/2\lambda^2$ for small separations this gives equation (S27). When the separation is no longer small, the symmetric-mode term changes the optimum and the full expression (S26) must be minimized. The signal-amplitude scale estimates the breakdown of the approximation; it is not a detachment threshold. In a two-class problem the signal variance $A$ can shrink to zero and the optimum becomes degenerate; in practice other data hold $A$ at a finite value, which we model with a third class at a distant target value. Figure 8(d) instead fixes the signal variance and directly minimizes the two-cluster evidence; its search-bound solutions and the three-cluster comparison are distinct calculations.

If a class varies along a factor $\xi$ with target slope $g$, so that $y = y_j + g\xi + \varepsilon$, and DKL embeds it as $z = \mu_j + \alpha\xi$, then within the class, where $\alpha|\xi| \ll \lambda$, the GP reduces to Bayesian linear regression on $\xi$ with prior slope variance $s = \sigma_f^2\alpha^2/\lambda^2$. The evidence is maximized at

$$s^* = \max\left(0, \frac{(\xi^\top y)^2}{|\xi|^4} - \frac{\sigma_n^2}{|\xi|^2}\right) \tag{S29}$$

the classical automatic-relevance result [97]. A factor whose explained target variance is below the noise is collapsed, $\alpha = 0$; a factor that predicts the target is laid out along a direction in proportion to its effect. In this restricted calculation, DKL keeps target-relevant variation and arranges it monotonically in $y$. This is the mechanism behind the feature collapse documented for unconstrained DKL [37] and its tendency to overfit through the marginal likelihood [36].

To derive equation (S29), expand the kernel within a class for $\alpha|\xi| \ll \lambda$: $k(z_a, z_b) \approx \sigma_f^2(1 - \alpha^2(\xi_a - \xi_b)^2/2\lambda^2)$. Up to a constant absorbed in the class mean, this is a linear kernel $s\,\xi_a\xi_b$ with $s = \sigma_f^2\alpha^2/\lambda^2$. For Bayesian linear regression with one feature, the negative log evidence is $\frac{1}{2}\mathrm{y}^\top(s\xi\xi^\top + \sigma_n^2 I)^{-1}\mathrm{y} + \frac{1}{2}\ln(1 + s|\xi|^2/\sigma_n^2)$ plus constants, whose minimum over $s \geq 0$ is equation (S29).

Merging also limits structural resolution. Classes whose targets differ by less than $\delta_c = (2\sigma_n^2/n)^{1/2}$ merge (equation (S27)), so the number of structural classes that DKL can keep apart is bounded by the number of resolvable target levels,

$$K_{\mathrm{eff}} \lesssim 1 + \frac{y_{\max} - y_{\min}}{\delta_c} \tag{S30}$$

regardless of how many distinct structures the data contain.

Figure 8(d) calculation. The panel minimizes equation (S26) with fixed signal and noise variances, using $v_\pm = n\sigma_f^2(1 \pm c) + \sigma_n^2$ and $c = \exp[-d^2/2]$, with $d = \Delta/\lambda$. It uses $n = 20$, $\sigma_f^2 = 2.5$ and $\sigma_n^2 = 0.01$. A logarithmic distance search, including zero, is followed by bounded refinement. The upper search bound is $d = 8$; boundary solutions are labeled as such, not reported as a uniquely determined detached separation. The small-distance comparison is $d \simeq \sqrt{\max(\delta^2 - 2\sigma_n^2/n, 0)}/\sigma_f$, restricted to its small-separation range. The separate three-cluster calculation instead uses the three-cluster model with optimized hyperparameters.

### S4.3 Hybrid VAE–DKL geometry

The hybrid objective per data point is

$$\mathcal{L}_{\mathrm{hyb}} = \mathcal{L}_{\mathrm{VAE}} + \frac{\eta}{N}\,\mathcal{L}_{\mathrm{DKL}}(\mu_1, \dots, \mu_N) \tag{S31}$$

where $\mathcal{L}_{\mathrm{VAE}}$ is the per-point loss of section 2, $\mathcal{L}_{\mathrm{DKL}}$ is equation (26) evaluated on the latent means $\mu_a$ of $N$ labeled points, and $\eta$ sets the relative weight. Four results follow.

For classes with different targets, the two objectives need not compete. Because the kernel length scale is learned, the DKL term depends on latent distances only through $\Delta/\lambda$. For two classes with $\delta^2 > 2\sigma_n^2/n$, the GP reaches its optimum at any VAE separation $\Delta$ by choosing

$$\lambda^* = \frac{\sigma_f\,\Delta}{\sqrt{\delta^2 - 2\sigma_n^2/n}} \tag{S32}$$

so the DKL term exerts no force on the VAE layout. Walls, the wall-width law and, for isotropic rescaling, the moment identity T2 are unchanged. With one length scale shared by many classes, this freedom is lost; the remaining force acts on the arrangement, not on the scale, as described below.

For classes with statistically indistinguishable targets, a finite predictive coupling can instead drive merging. DKL prefers to merge two such classes (Section S4.2), while the VAE pays the full Bayes risk of confusing them. Comparing the DKL gain from merging with the VAE cost gives the weight at which merging sets in:

$$\eta_{\mathrm{merge}} \approx \frac{4n_b R_{ij}}{\ln\left(n_b\sigma_f^2/2\sigma_n^2\right)} \tag{S33}$$

where $n_b$ is the number of points of each class in the batch on which the GP is evaluated and $R_{ij}$ is the structural coupling of section 2. Classes that look different (large $R_{ij}$) resist merging; classes that look alike merge at moderate weight. The hybrid therefore keeps structural distinctions that pure DKL erases, but only for structurally dissimilar classes and only below $\eta_{\mathrm{merge}}$.

Within each class, the predictive relevance of a factor determines whether it survives. For a within-class factor that does not predict the target, the DKL term penalizes latent spread through the log-determinant. Adding this penalty to the VAE cost of section 4.1 raises the collapse threshold of the factor from $A_j = 1/2$ to

$$A_{j,c} = \frac{1}{2} + \frac{\eta\kappa}{2N} \tag{S34a}$$

$$\kappa = \frac{\sigma_f^2 s_j^2}{\lambda^2 \sigma_n^2} \tag{S34b}$$

so increasing $\eta$ removes structural detail that does not predict the target, and reconstruction degrades. For a factor that does predict the target, the DKL gain from resolving it does not depend on its latent extent, because $\lambda$ adapts. When the GP reads latent means, such a factor is retained even if the VAE alone would collapse it, possibly at an extent below the posterior width; it is then predictive but not decodable. When the GP reads latent samples, the factor must exceed the posterior width to be resolved, and its survival again depends on $\eta$. Together these two effects produce the trade-off between prediction and reconstruction reported for QM9, where longer training improved reconstruction and worsened prediction, and a large DKL weight suppressed latent variability [39].

**Merging weight.** For two classes with equal targets, the DKL loss is lower by approximately $G = \frac{1}{2}\ln\left(n_b \sigma_f^2 / 2\sigma_n^2\right)$ when they merge ($c = 1$) than when they are separated (section S4.2). Merging costs the VAE the Bayes risk of confusing the classes, about $R_{ij}$ per point for each of $2n_b$ points in the batch, less the small KL saving. Setting $\eta G$ equal to this cost, with the DKL term normalized per batch, gives equation (S33).

**Collapse threshold for target-irrelevant factors.** For a factor of within-class spread $c_j$ and posterior width $s_j$, the VAE cost per point is $A_j/\left(1+\rho_j\right) + \frac{1}{2}\ln\left(1+\rho_j\right)$ with $\rho_j = c_j^2/s_j^2$ (section 4.1, equations (19a) and (19b)). Spreading $n$ points with equal targets over a latent extent small compared with $\lambda$ raises the within-class eigenvalues of the kernel matrix from $\sigma_n^2$ by about $\sigma_f^2 c_j^2 \xi^2/\lambda^2$, so the log-determinant rises by about $n\sigma_f^2 c_j^2/(\lambda^2\sigma_n^2)$ per unit factor variance. Per point this adds $\frac{1}{2}(\eta/N)\kappa\rho_j$ to the cost, with $\kappa$ as in equations (S34a) and (S34b). With $t = 1 + \rho_j$, stationarity gives

$$\frac{\eta\kappa}{2N} t^2 + \frac{t}{2} - A_j = 0 \tag{S35}$$

whose root satisfies $t \leq 1$, that is collapse, when $A_j \leq \frac{1}{2} + \eta\kappa/2N$.

The normalization of the GP contribution is part of the coupling definition. The formulas above use a per-batch GP evidence and distinguish a GP acting on posterior means from one acting on samples; the raw predictive weights of the MNIST sweep in Section S4.4 follow a different normalization and are not directly comparable with η.

**S4.4 MNIST comparison and the structure–prediction trade-off**

Figure 9 compares ordinary VAEs, DKL, and hybrid VAE–DKL on MNIST. The predictive targets are ink density and a scalar assigned to digit identity. Results use five training seeds and held-out measurements; Figure 9(e,f) displays means and sample standard deviations. A zero-predictive-weight hybrid is the matched-budget reconstructive control.

The result is task-dependent. The scalar class target explicitly distinguishes digit identities, whereas ink density does not. These targets test different information requirements and should not be interpreted as a universal ordering of learning methods.

## S5 Sampling, interpolation and latent-space search

### S5.1 Wall probability, density ratios and paths

In main equation (27), $\mathrm{Per}_\gamma$ is the Gaussian perimeter of the partition: $1/\sqrt{2\pi}$ for two classes and $K/(2\sqrt{2\pi})$ for $K$ sectors in two dimensions. The formula matches numerical optima to 0.3% in one dimension and to 1–3% for three classes in two dimensions. For $R = 10$ to $10^4$ it predicts that 4–11% of prior samples fall in walls, decaying only as $1/\ln(1/\beta)$.

**Wall probability relative to the data.** At the two-class optimum of section 3 the aggregate posterior at the wall is suppressed relative to the prior by the factor $e^{-k^*/2}/s$, whereas integration over the finite responsibility band gives

$$\frac{P_{\mathrm{prior}}(W)}{P_{\mathrm{post}}(W)} \simeq \frac{b}{\sinh b}\, s\, e^{k^*/2} \simeq 1.033\, R \tag{S36}$$

with b = atanh(0.8) = ln 3 for the 0.9 responsibility threshold, using the strong-separation relations in main equations (11a) and (11b).

### S5.2 Conditional surrogate and information-gain estimates

**Resolution-count heuristic.** For a squared-exponential kernel with length scale $\lambda$ on a latent region of size $D$, the number of regions at the kernel resolution gives the heuristic scaling below. The information-gain symbol denotes only a proxy in equations (S37a) and (S37b), not a proved bound:

$$\gamma_T \propto \left(\frac{D}{\lambda}\right)^d \tag{S37a}$$

$$\gamma_T \propto (2D\ln R)^d \quad \text{for } \lambda \approx \ell \tag{S37b}$$

This is a scaling estimate for a squared-exponential surrogate when the target actually varies on the wall scale and the fitted length scale is comparable to that width. It is not an information-gain theorem or a sample-complexity bound; an actual bound must retain observation-budget and noise dependence. Dependence on domain size, noise, kernel assumptions and logarithmic factors is suppressed in the proportionality. Equal-property structural classes, alternative kernels and domain-aware models need not incur this particular penalty.

### S5.3 Weighted retraining and rare-state retention

**Weighted retraining.** When the entire per-example objective is multiplied by class weights $w_j$, the rare-class result of section 3 becomes, to leading order,

$$\frac{s_i}{s_j} \to \frac{w_i \pi_i^0}{w_j \pi_j^0} \tag{S38}$$

so up-weighted classes expand and the rest compress into islands. The effective frequencies are proportional to weight times class frequency and normalized to sum to one. Weighting only reconstruction or only prediction does not give this substitution.

### S5.4 Folded-versus-unfolded search and its protocol

**Folded latents.** Each strand of a folded latent carries one level of the folded factor and a full sweep of the factor running along it, so the specified separable quadratic test can have a local optimum on each strand. This is not a statement about every possible objective on folded data. With several folded factors,

$$N_{\mathrm{loc}} \approx \textstyle\prod_j N_j^* = \prod_j \sqrt{2A_j} \tag{S39a}$$

$$\gamma_T^{\mathrm{fold}} \propto N_c \left(\frac{D}{\lambda}\right)^d \tag{S39b}$$

$$\gamma_T^{\mathrm{unfold}} \propto \left(\frac{D}{\lambda}\right)^{d+1} \tag{S39c}$$

## S6 Scope of the analytical results

The conditional-mean decoder, admissible-affine stationarity and regular-solution decomposition are identities under their stated assumptions. Wall, rare-class, covariance, topology and folding laws follow from the restricted models developed in Sections S1–S5. Numerical minimization tests these models; trained-network experiments test whether related patterns are reached under the specified architecture and training procedure.

## S7 Methods

### S7.1 Analytic optimization and class-aggregate layouts

Analytical curves were evaluated by numerical quadrature and direct optimization of the stated objectives. Numerical tests compare the analytical predictions with finite-parameter minima and separately examine covariance restrictions, class frequencies, and initialization dependence.

**Two-class quantities.** The sharpness $k^*$ minimizes equations (10a) and (10b), with $g$ evaluated by a mixed quadrature scheme (256-point Gauss–Hermite evaluation for smaller sharpness and boundary-centered Gauss–Legendre integration for sharper walls, checked at 512 points) and a dense logarithmic grid followed by bounded refinement. Binodals compare the separated optimum (multistart minimization over means and log widths, continued along the separated branch, with the crossing located by root finding to $10^{-4}$; Section S2.5) with the mixed loss $\chi\pi(1-\pi)$.

**Class-aggregate layouts.** Domain centers and posterior widths (isotropic, diagonal per class, or full covariance per class through a Cholesky factor) were optimized with L-BFGS-B on a $181 \times 181$ grid over $[-4,4]^2$ ($121 \times 121$ for covariance annealing), from several random and structured starts. Three-dimensional layouts used a $51^3$ grid over $[-3.2,3.2]^3$, and junction splitting used triple points located on an $801 \times 801$ label map.

### S7.2 Nonparametric encoders, synthetic factors and rare-class images

**Nonparametric encoder.** Each data point has its own Gaussian posterior (free mean and per-dimension log width, bounded below by 2.5 grid spacings), and the decoder is the exact posterior-weighted mean on a latent grid of 2400–6000 points in one dimension or 161–301 points per axis in two dimensions. Gradients are analytic and were checked by finite differences. This model was used for rare classes, rings, label folding and folding. For label folding the class prototypes lie on an equilateral triangle with $|a_i - a_j|^2 = \chi$.

**Folding model.** Two uniform factors on a $30 \times 30$ grid share one latent axis ($A_f = 8$, 1000 latent grid points). Unfolded, perturbed (modulated) and serpentine starts with prescribed strand numbers were relaxed; the wrinkle number is the number of sign changes of the slope of the latent coordinate with respect to the excess factor along the shared factor.

**Rare-class shapes.** Prototypes are $40 \times 40$ pixel images of a circle, a square, a superellipse (rounded square) and an elongated rhombus, lightly smoothed; the circle–square squared distance sets $\chi_{AB}$. For the smoothing metric the images were convolved with a Gaussian of 4 pixels before distances were computed.

### S7.3 Trained benchmarks: data, estimators and metric definitions

Two-class digits. Figure 3 uses the scikit-learn digits 0/1 benchmark and the estimators defined in Section S2.2. The separate training-dynamics comparison is discussed in Section S3.6.

MNIST. Figures 1 and 9 use ten-class digit representations. Figure 1 compares encoded means, decoded classes, classifier confidence, and encoded-data density; Figure 9 compares reconstruction and predictive objectives. Encoded-mean density estimates are distinct from the full aggregate posterior.
Density estimators. Figure 1's encoded-mean density is a kernel-density estimate, whereas Figure 3 uses one regularized Gaussian per class fitted to encoded means. These are not interchangeable estimators and neither is the full mixture of per-observation posteriors. Figure 3's variance-inclusive control adds $\mathrm{diag}(\mathbb{E}[\exp(\log\sigma^2)])$ to the within-class covariance of means. The moment diagnostic uses mean squared encoded means plus mean encoder variance. The fitted-density construction must accompany each reported wall probability or width.
Predictive targets and metrics. Ink density is the mean normalized MNIST pixel intensity; the class-valued target assigns one scalar to each digit. Targets are standardized on training data. Digit discrimination is measured by five-nearest-neighbor accuracy; ink smoothness compares target differences among latent neighbors with random pairs. Reconstruction cost is mean pixelwise binary cross-entropy.
Rotated digits. Figure 6 uses randomly rotated scikit-learn digits with a two-dimensional Bernoulli VAE. Winding is determined from encoded trajectories over a full rotation. The base-image identities in the displayed sweeps were also used for training, so the result is a representation diagnostic rather than an identity-disjoint test.
Cards. Figure 7 compares ordinary VAE decoded grids and latent distributions for four rotation/shear conditions. The rotationally invariant representations and their angle diagnostics provide the comparison discussed in Section S3.3.
Synthetic folding. The reduced nonparametric calculations and the trained-network test employ different encoder families. Figure S5 illustrates the absence of robust equilibrium folding in the trained examples; Section S3.6 discusses training dynamics.

## S8 Extended scientific and methodological context

### S8.1 Scientific representations

Prior studies use latent coordinates for imaging, spectroscopy, microstructures, molecular systems, and complementary measurements [25–31, 88, 98–121]. They motivate the question of physical interpretability but do not independently validate the domain-wall laws derived here.

### S8.2 Latent-space optimization

Molecular, materials and process-design studies use learned coordinates for search [27–29, 32, 33, 38, 81, 87–91, 122–133]. Valid decoding, representation coverage, and property-measurement coverage remain distinct requirements. The mechanisms in this paper provide possible explanations, not additional experimental findings about those studies.

### S8.3 Predictive and hybrid architectures

Reconstruction-based and target-shaped representations retain different information [34–39, 81–86, 134]. The principal geometric consequences are derived in Sections S4.2–S4.4.

**Supplementary references**

References cited in the Supplementary Information and not in the main text are listed below.

[92] Helfand E and Tagami Y 1972 Theory of the interface between immiscible polymers. II *J. Chem. Phys.* **56** 3592–601
[93] Horan D, Richardson E and Weiss Y 2021 When is unsupervised disentanglement possible? *Advances in Neural Information Processing Systems* **34**
[94] Wu F Y 1982 The Potts model *Rev. Mod. Phys.* **54** 235–68
[95] Gibbs J W 1878 On the equilibrium of heterogeneous substances *Trans. Conn. Acad. Arts Sci.* **3** 108–248 and 343–524
[96] Welch L R 1974 Lower bounds on the maximum cross correlation of signals *IEEE Trans. Inf. Theory* **20** 397–9
[97] Tipping M E 2001 Sparse Bayesian learning and the relevance vector machine *J. Mach. Learn. Res.* **1** 211–44
[98] Banko L *et al* 2021 Deep learning for visualization and novelty detection in large x-ray diffraction datasets *npj Comput. Mater.* **7** 104
[99] Grossutti M *et al* 2022 Deep learning and infrared spectroscopy: representation learning with a β-variational autoencoder *J. Phys. Chem. Lett.* **13** 5787–93
[100] Ziatdinov M A, Yaman M Y, Liu Y, Ginger D and Kalinin S V 2024 Semi-supervised learning of images with strong rotational disorder: assembling nanoparticle libraries *Digit. Discov.* **3** 1213–20
[101] Haghparast S *et al* 2026 Detecting continuous structural heterogeneity in single molecule localization microscopy data with a point cloud variational auto-encoder *Sci. Rep.* **16** 1379
[102] Ayyubi R A W *et al* 2026 Unsupervised defect clustering in atomic-resolution microscopy using a convolutional variational autoencoder *npj Comput. Mater.* **12** 166
[103] Shang M *et al* 2022 Investigating carboxysome morphology dynamics with a rotationally invariant variational autoencoder *J. Phys. Chem. A* **126** 5021–30
[104] Ziatdinov M, Ghosh A, Wong C Y and Kalinin S V 2022 AtomAI framework for deep learning analysis of image and spectroscopy data in electron and scanning probe microscopy *Nat. Mach. Intell.* **4** 1101–12
[105] Kalinin S V *et al* 2022 Machine learning in scanning transmission electron microscopy *Nat. Rev. Methods Primers* **2** 11
[106] Zhong E D, Bepler T, Berger B and Davis J H 2021 CryoDRGN: reconstruction of heterogeneous cryo-EM structures using neural networks *Nat. Methods* **18** 176–85
[107] Lopez R, Regier J, Cole M B, Jordan M I and Yosef N 2018 Deep generative modeling for single-cell transcriptomics *Nat. Methods* **15** 1053–8
[108] White M D *et al* 2024 Exploring descriptors for titanium microstructure via digital fingerprints from variational autoencoders *Comput. Mater. Sci.* **240** 112992
[109] Templeton W F *et al* 2024 Expediting structure–property analyses using variational autoencoders with regression *Comput. Mater. Sci.* **242** 113056
[110] Jones R E *et al* 2024 Multiscale simulation of spatially correlated microstructure via a latent space representation *Int. J. Solids Struct.* **301** 112966
[111] Wang L, Chan Y-C, Ahmed F, Liu Z, Zhu P and Chen W 2020 Deep generative modeling for mechanistic-based learning and design of metamaterial systems *Comput. Methods Appl. Mech. Eng.* **372** 113377
[112] Pan Z, Wang Y, Cao Y and Gui W 2024 VAE-based interpretable latent variable model for process monitoring *IEEE Trans. Neural Netw. Learn. Syst.* **35** 6075–88
[113] Solera-Rico A *et al* 2024 β-variational autoencoders and transformers for reduced-order modelling of fluid flows *Nat. Commun.* **15** 1361

[114] Sultan M M, Wayment-Steele H K and Pande V S 2018 Transferable neural networks for enhanced sampling of protein dynamics *J. Chem. Theory Comput.* **14** 1887–94
[115] Chen W and Ferguson A L 2018 Molecular enhanced sampling with autoencoders: on-the-fly collective variable discovery and accelerated free energy landscape exploration *J. Comput. Chem.* **39** 2079–102
[116] Jin W, Barzilay R and Jaakkola T 2018 Junction tree variational autoencoder for molecular graph generation *Proc. 35th Int. Conf. on Machine Learning (PMLR vol 80)* pp 2323–32
[117] Simonovsky M and Komodakis N 2018 GraphVAE: towards generation of small graphs using variational autoencoders *Artificial Neural Networks and Machine Learning – ICANN 2018* pp 412–22
[118] Samanta B *et al* 2020 NeVAE: a deep generative model for molecular graphs *J. Mach. Learn. Res.* **21**
[119] Ilnicka A and Schneider G 2023 Designing molecules with autoencoder networks *Nat. Comput. Sci.* **3** 922–33
[120] Chun S *et al* 2026 Inverse design of metamaterials using progressively trained variational autoencoders *Eng. Res. Express* **8** 195353
[121] Tran T V, Nanthakumar S S and Zhuang X 2025 Deep learning-based framework for the on-demand inverse design of metamaterials with arbitrary target band gap *npj Artif. Intell.* **1** 2
[122] Lee S, Chu J, Kim S, Ko J and Kim H J 2023 Advancing Bayesian optimization via learning correlated latent space *Advances in Neural Information Processing Systems* **36**
[123] Deshwal A and Doppa J R 2021 Combining latent space and structured kernels for Bayesian optimization over combinatorial spaces *Advances in Neural Information Processing Systems* **34**
[124] Chu J, Park J, Lee S and Kim H J 2024 Inversion-based latent Bayesian optimization *Advances in Neural Information Processing Systems* **37**
[125] Noh J, Kim J, Stein H S, Sanchez-Lengeling B, Gregoire J M, Aspuru-Guzik A and Jung Y 2019 Inverse design of solid-state materials via a continuous representation *Matter* **1** 1370–84
[126] Yao Z *et al* 2021 Inverse design of nanoporous crystalline reticular materials with deep generative models *Nat. Mach. Intell.* **3** 76–86
[127] Batra R, Dai H, Huan T D, Chen L, Kim C, Gutekunst W R, Song L and Ramprasad R 2020 Polymers for extreme conditions designed using syntax-directed variational autoencoders *Chem. Mater.* (https://doi.org/10.1021/acs.chemmater.0c03332)
[128] Liu Z, Raju L, Zhu D and Cai W 2019 A hybrid strategy for the discovery and design of photonic nanostructures arXiv:1902.02293
[129] Biswas A, Vasudevan R, Ziatdinov M and Kalinin S V 2023 Optimizing training trajectories in variational autoencoders via latent Bayesian optimization approach *Mach. Learn.: Sci. Technol.* **4** 015011
[130] Antonova R, Rai A, Li T and Kragic D 2020 Bayesian optimization in variational latent spaces with dynamic compression *Proc. Conf. on Robot Learning (PMLR vol 100)* pp 456–65
[131] Chandra R, Horne R I and Vendruscolo M 2024 Bayesian optimization in the latent space of a variational autoencoder for the generation of selective FLT3 inhibitors *J. Chem. Theory Comput.* **20** 469–76
[132] Slautin B N, Liu Y, Funakubo H and Kalinin S V 2024 Unraveling the impact of initial choices and in-loop interventions on learning dynamics in autonomous scanning probe microscopy *J. Appl. Phys.* **135** 154901

[133] Slautin B N, Barakati K, Funakubo H, Ziatdinov M A, Shvartsman V V, Lupascu D C and Kalinin S V 2025 Reward driven discovery of the optimal microstructure representations with invariant variational autoencoders arXiv:2510.00243
[134] Moss H B, Ober S W and Diethe T 2025 Return of the latent space COWBOYS: re-thinking the use of VAEs for Bayesian optimisation of structured spaces *Proc. 42nd Int. Conf. on Machine Learning (ICML)*